%% file: main_arxiv.tex
\documentclass{article}
\usepackage[a4paper]{geometry}
\usepackage[colorlinks=true, allcolors=blue]{hyperref}
\usepackage[table]{xcolor} 
\usepackage{graphicx}
\usepackage{multirow}
\usepackage{amsmath,amssymb,amsfonts}
\usepackage{amsthm}
\usepackage{mathrsfs}
\usepackage[title]{appendix}
\usepackage{textcomp}
\usepackage{manyfoot}
\usepackage{booktabs}
\usepackage{fancyvrb}
\usepackage{subcaption}
\usepackage{algorithm}
\usepackage{algorithmicx}
\usepackage{algpseudocode}
\usepackage{listings}
\usepackage[normalem]{ulem}
\usepackage{longtable}
\usepackage{float}
\usepackage[T1]{fontenc}  
\begin{document}

\title{Pretrained Multilingual Transformers Reveal Quantitative Distance Between Human Languages}

\author{Yue Zhao$^{1}$ \and Jiatao Gu$^{2}$ \and Paloma Jereti\v{c}$^{3,\ast}$ \and Weijie Su$^{4,\ast}$}

\maketitle

\begingroup
\renewcommand{\thefootnote}{\textsuperscript{$\ast$}}
\footnotetext{Corresponding authors.}
\renewcommand{\thefootnote}{\arabic{footnote}}
\footnotetext[1]{Institute for Functional Intelligent Materials, National University of Singapore. Email: \texttt{yuezhao@nus.edu.sg}.}
\footnotetext[2]{Department of Computer and Information Science, University of Pennsylvania. Email: \texttt{jgu32@upenn.edu}.}
\footnotetext[3]{Department of Linguistics, University of Pennsylvania. Email: \texttt{pjeretic@sas.upenn.edu}.}
\footnotetext[4]{Department of Statistics and Data Science, Wharton School, University of Pennsylvania. Email: \texttt{suw@wharton.upenn.edu}.}
\endgroup
\setcounter{footnote}{0}

\begin{abstract}
Understanding the distance between human languages is central to linguistics, anthropology, and tracing human evolutionary history. Yet, while linguistics has long provided rich qualitative accounts of cross-linguistic variation, a unified and scalable quantitative approach to measuring language distance remains lacking. In this paper, we introduce a method that leverages pretrained multilingual language models as systematic instruments for linguistic measurement. Specifically, we show that the spontaneously emerged attention mechanisms of these models provide a robust, tokenization-agnostic measure of cross-linguistic distance, termed Attention Transport Distance (ATD). By treating attention matrices as probability distributions and measuring their geometric divergence via optimal transport, we quantify the representational distance between languages during translation. Applying ATD to a large and diverse set of languages, we demonstrate that the resulting distances recover established linguistic groupings with high fidelity and reveal patterns aligned with geographic and contact-induced relationships. Furthermore, incorporating ATD as a regularizer improves transfer performance in low-resource machine translation. Our results establish a principled foundation for testing linguistic hypotheses using artificial neural networks. This framework transforms multilingual models into powerful tools for quantitative linguistic discovery, facilitating more equitable multilingual AI.
\end{abstract}

\input{sections/main_intro}

\input{sections/main_method}
\input{sections/main_tree_new}

\input{sections/main_transfer}

\input{sections/main_discussion}

\section*{Acknowledgments}
We thank Marlyse Baptista and Don Ringe for their feedback on this work, which was informed by their expertise in contact linguistics and historical linguistics. This work was supported in part by the National University of Singapore and the AI for Science Institute, Beijing, through the AISI--NUS Joint Research Initiative Fund 2025 Award, a Meta Faculty Research Award, and Wharton AI for Business.

\section*{Data and Code Availability}
The code is available at \url{https://github.com/yzhao98/ATD-Linguistics}.

\bibliographystyle{plain}
\bibliography{main_arxiv}

\begin{appendices}
	
\input{sections/appendix_method}

\input{sections/appendix_setting}
\input{sections/appendix_control_stat}
\input{sections/appendix_m2m_100}

\input{sections/appendix_llama_new}
\input{sections/appendix_transfer}

\end{appendices}

\end{document}

%% file: sections/main_intro.tex
\section{Introduction}\label{sec1}

The similarities and differences among human languages are central questions in linguistics and anthropology, offering crucial insights into human history, migration, and cultural exchange~\cite{campbell2013historical}. Traditional historical linguistics relies on identifying cognates and systematic sound correspondences to establish genealogical relationships and precisely reconstruct language families~\cite{campbell2013historical}. Simultaneously, typological databases such as WALS~\cite{dryer2022wals} and Glottolog~\cite{swj-glottocodes} provide structured resources for large-scale comparisons independent of direct historical relatedness. While these approaches provide a basis for establishing similarities, their reliance on manually curated features makes it difficult to scale to the world’s over 7,000 languages or to integrate them with modern computational workflows.

In recent years, these long-standing questions have gained renewed relevance as pretrained multilingual language models are increasingly deployed across a wide range of languages~\cite{openai2023gpt4, dubey2024llama}. This shift has amplified the need for principled, quantitative notions of language distance that are both scalable and deeply coupled with the internal representations of these models. However, bridging the gap between large-scale artificial intelligence and formal linguistic measurement remains a significant challenge.

Computational approaches have sought to address these challenges. Early work used dictionaries or string-edit distance to approximate lexical similarity~\cite{serva2008indo, holman2008explorations}, while more recent studies exploited cross-lingual embeddings and alignment to quantify language distances~\cite{conneau2017word,artetxe2019massively, feng2022language}. In machine translation, the zero-shot performance of multilingual models has been proposed as an indirect measure of relatedness~\cite{johnson2017google}. Bayesian phylogenetic models have also been used to reconstruct Indo-European trees~\cite{gray2003language, bouckaert2012mapping}. These methods have advanced scalable comparison, but they share well-known limitations: dependence on tokenization and vocabulary design, sensitivity to alignment and data biases, and instability in low-resource languages. Importantly, their relationship to transfer performance in modern pretrained models has been shown to vary across tasks and languages~\cite{pires2019multilingual, wu2020all, lauscher2020zero}. A tokenization-agnostic, model-coupled, and data-driven quantitative distance is still lacking.

Here we introduce Attention Transport Distance (ATD), a new measure of cross-linguistic distance derived from pretrained language models. The key insight is that the attention mechanism in transformer models naturally encodes “who attends to whom” during translation, representing a probabilistic alignment~\cite{vaswani2017attention, fan2021beyond}. By comparing attention distributions using optimal transport~\cite{cuturi2013sinkhorn,peyre2019computational}, we obtain a tokenization-agnostic quantitative distance. Unlike dictionary- or embedding-based approaches, ATD arises directly from the pretrained model’s internal structure and is related to the attention patterns learned inside the model rather than the tokenizer, which may introduce human-specific biases. In this way, ATD turns the problem of cross-linguistic comparison into one of distance matrices, enabling systematic analysis through trees, networks, and clustering.

Applying ATD to large-scale pretrained multilingual language models, we construct a pairwise distance matrix over languages with reliable translation quality and analyze its structure using classical distance-based methods. Language groupings obtained via the Neighbor-Joining (NJ) algorithm~\cite{saitou1987neighbor}, a standard method for reconstructing phylogenetic trees from distance matrices, exhibit strong consistency with established linguistic classifications, with Indo-European languages examined as a detailed case study. Complementary geographic visualizations further reveal patterns aligned with known distributions of languages and historical contact, providing external validation of the inferred distances. Finally, incorporating ATD as a regularizer in transfer learning yields improved translation performance for low-resource languages, suggesting that language distances derived from pretrained models can inform cross-lingual generalization and downstream model adaptation.

Overall, this study reframes pretrained multilingual language models as quantitative instruments for linguistic analysis, providing a scalable framework for cross-linguistic comparison in the context of established typological resources \cite{dryer2022wals, swj-glottocodes}, while revealing cross-linguistic structures implicitly encoded in pretrained models that can inform cross-lingual transfer and low-resource settings \cite{lauscher2020zero}. More broadly, our work illustrates how large-scale AI models can serve not only as engineering systems but also as tools for scientific inquiry, fostering deeper integration between language science and artificial intelligence.

%% file: sections/main_method.tex
\section{Results}

We organize our results in four steps. We first define the proposed ATD and show how it yields a tokenization-agnostic distance matrix across languages. Based on this distance matrix, we then apply classical unsupervised analyses to test whether ATD recovers known relationships between languages, including geneological tree structure and  network-like contact patterns. Finally, we evaluate ATD as a regularizer in transfer learning, showing improvements for low-resource translation. Together, these results establish ATD both as a scientifically interpretable measure of language relations and as a practically useful signal for multilingual AI.

\subsection{Attention Transport Distance: A Tokenization-Agnostic Attention-Based Measure of Cross-Linguistic Similarity}

We begin by introducing ATD, a tokenization-agnostic measure of cross-linguistic similarity derived from pretrained multilingual models. The core idea of ATD is to transform model-internal attention patterns into a quantitative object that captures how languages relate to one another in the representational space learned through multilingual translation. By interpreting cross-lingual attention as a probabilistic alignment and comparing these distributions using optimal transport, ATD yields a principled distance measure that is independent of target-language tokenization and vocabulary design.

\begin{figure}[ht]
	\centering
	\includegraphics[width=\textwidth]{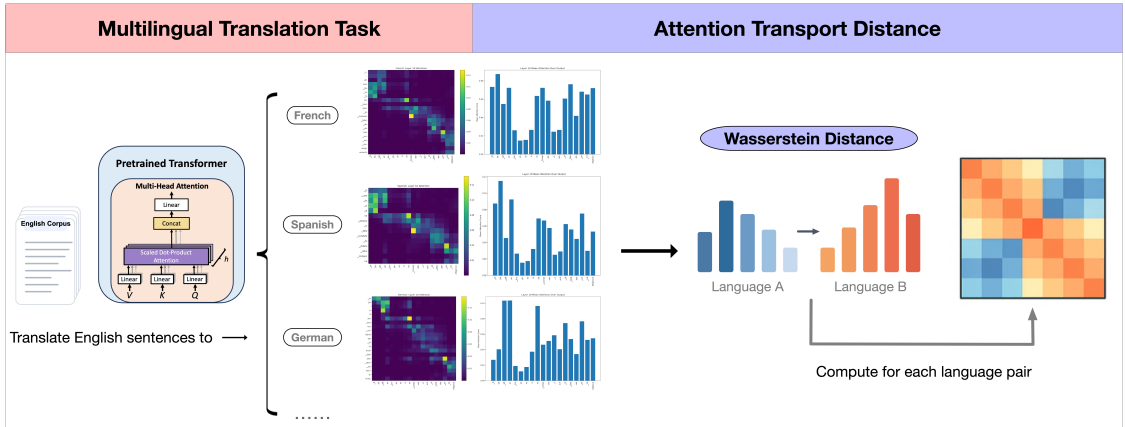}
	\caption{Flowchart of the Attention Transport Distance method.}
	\label{fig:method_flowchart_simplified}
\end{figure}

\begin{figure}[h]
	\centering
	\includegraphics[width=\linewidth]{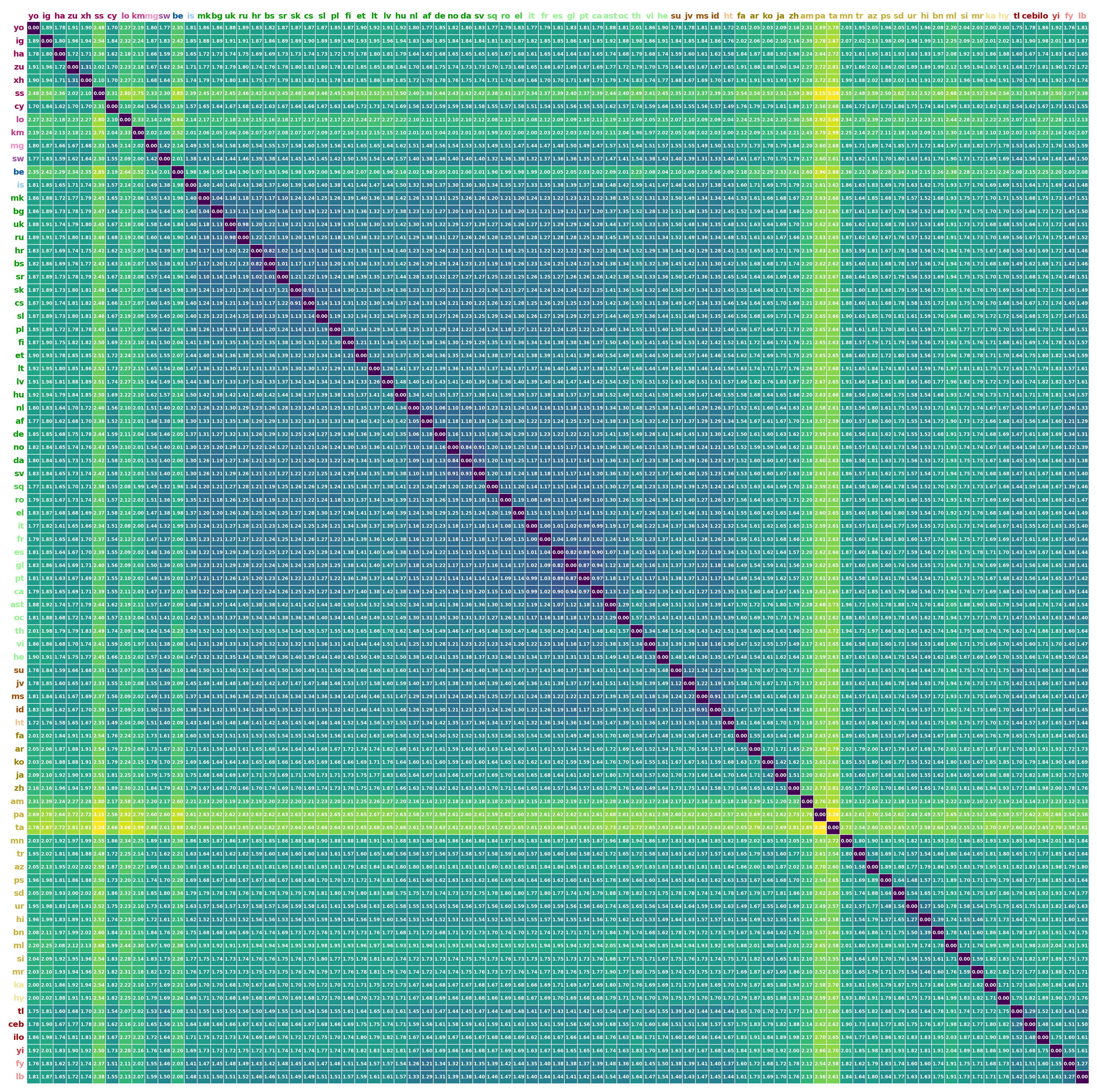}
	\caption{\textbf{ATD heat-map for M2M-100.}
		Pair-wise ATD values between the 81 target languages are shown as a
		heat-map. } 
	\label{fig:heatmap_atd}
\end{figure}

Fig.~\ref{fig:method_flowchart_simplified} provides an overview of the ATD pipeline. We consider a multilingual translation setting in which a fixed set of English source sentences is translated into multiple target languages using a pretrained multilingual model. From each translation, we extract cross-attention matrices that describe how source tokens attend to target-side representations. These attention matrices are averaged across heads and marginalized over target tokens to obtain a language-specific attention distribution over the shared source tokens. The ATD between a pair of languages is then defined as the Wasserstein-2 distance between their corresponding attention-derived distributions (see Appendix~\ref{sec:methods} for the formal mathematical derivation).
By averaging ATD values across sentences in the corpus, we obtain a dense, symmetric pairwise distance matrix over languages. This matrix constitutes the primary empirical object studied in the remainder of this work. Fig.~\ref{fig:heatmap_atd} visualizes the ATD distance matrix for the M2M-100 model~\cite{fan2021beyond} over 81 target languages with reliable translation quality. Additional results on the full set of 100 M2M-100 languages and on the Llama-3 model~\cite{dubey2024llama} are reported in the Appendix~\ref{app:results_m2m_100} and Appendix~\ref{app:results_llama}.

Even without any linguistic supervision, the ATD matrix exhibits pronounced large-scale structure. Languages organize into contiguous blocks with systematically smaller within-block distances and larger between-block distances, indicating that pretrained multilingual models encode coherent cross-linguistic organization in their attention patterns. This block structure is visible at multiple scales, suggesting the presence of both coarse-grained and fine-grained relationships among languages.
For ease of visualization, languages in Fig.~\ref{fig:heatmap_atd} are ordered and annotated according to the hierarchical clustering derived in the following section.

%% file: sections/main_tree_new.tex
\subsection{Phylogenetic structure recovered from ATD via Neighbor-Joining}
\label{sec:results_nj}

\begin{figure}[!htbp]
    \centering
    \includegraphics[width=0.9\textwidth]{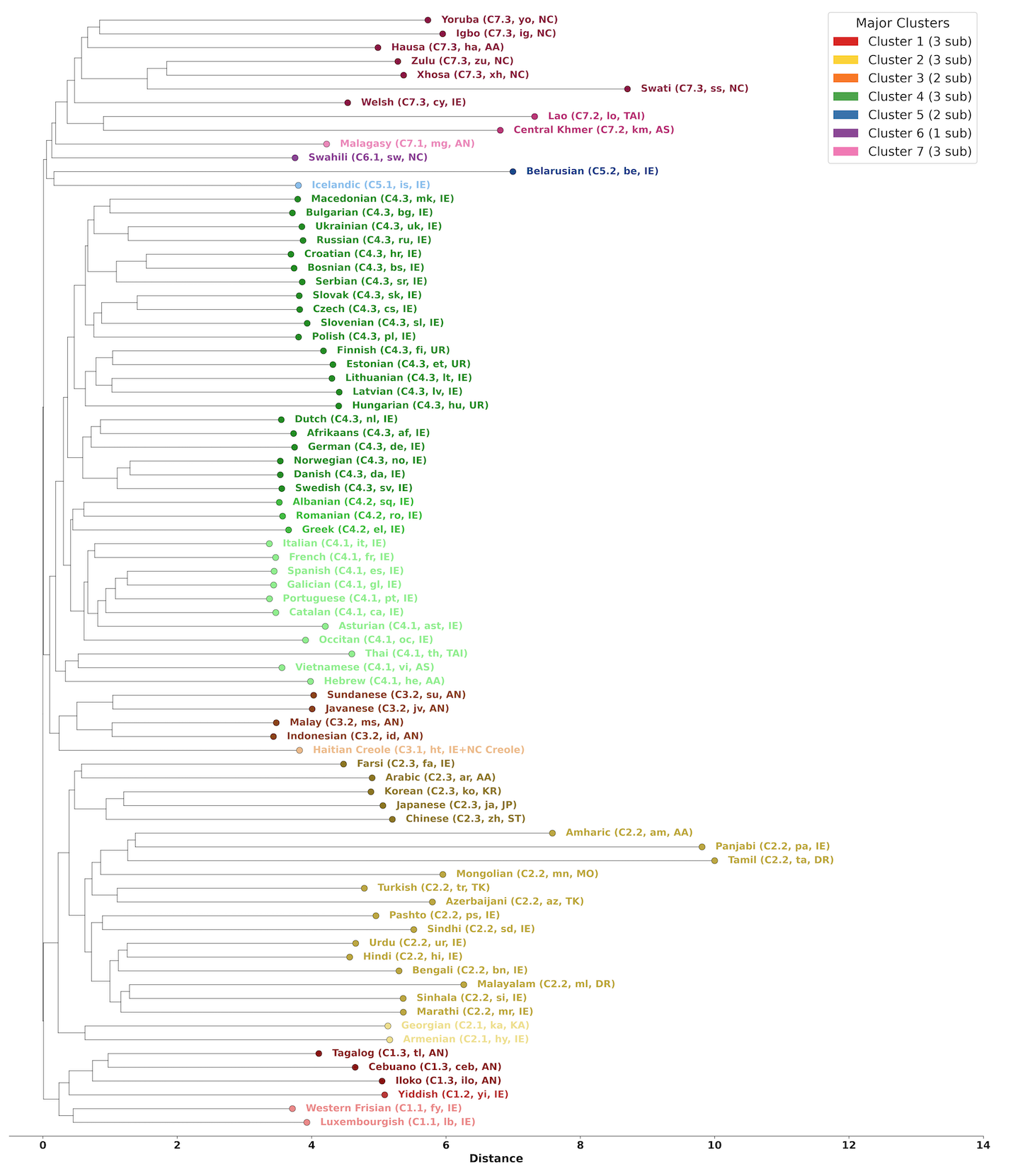}
    \caption{Neighbor-Joining tree inferred from ATD distances in the M2M-100 model.}
    \label{fig:nj_tree_all}
\end{figure}

To convert the ATD distance matrix into an interpretable hierarchical representation, we apply the classical Neighbor-Joining algorithm~\cite{saitou1987neighbor} to reconstruct an unrooted, distance-based tree over languages (Fig.~\ref{fig:nj_tree_all}). Based on the resulting topology, we further derive seven major clusters via depth-based cutting of the NJ tree (see Appendix \ref{app:nj} for implementation and clustering details), and use these cluster assignments consistently across figures for visualization and analysis.

At a global level, the inferred NJ structure exhibits strong correspondence with established linguistic organization. A broad Eurasian macro-structure emerges, within which Indo-European languages form a coherent subtree. Major subfamilies appear as internally compact regions of the tree, while more distant branches separate with progressively longer patristic distances. Importantly, the resulting topology reflects graded similarity rather than enforcing a binary notion of relatedness, suggesting that ATD captures continuous variation in cross-linguistic representational similarity.

Before proceeding to fine-grained linguistic interpretation, we assess the fidelity of the NJ reconstruction. The cophenetic correlation between the original ATD distances and the patristic distances induced by the NJ tree is very high (Pearson $r=0.9881$, Spearman $\rho=0.9846$, both $p<0.001$). This result indicates that the NJ tree provides an accurate low-dimensional structural summary of the ATD geometry, supporting its use as a reliable basis for subsequent linguistic analysis.

With this global structure and quantitative fidelity established, we now turn to a detailed linguistic interpretation of the ATD-derived clusters. Several well-known genealogical patterns are recovered with high coherence. 

All of the Niger-Congo languages are found in cluster 7.3. 
Austronesian has 4 languages in cluster 3.2, grouping languages of the Indo-Malayan region 
and 3 languages in cluster 1.3, grouping the Philippine languages
(no other languages are found in those clusters). While these are two separate clusters, they are not far from each other in the ATD geometry. For example, we find Indonesian and Tagalog to be 1.44 apart despite being in different clusters, while Iloko and Cebuano at 1.48, which are in the same cluster.

With a few exceptions, the Indo-European languages are grouped together in cluster 4.
Romance languages form a compact substructure, with Iberian Romance varieties (Spanish, Portuguese, Galician, Catalan, Asturian) tightly grouped and Gallo- and Occitano-Romance varieties located nearby. Slavic languages likewise cluster closely, with a clear separation between West Slavic (Czech, Slovak, Polish), East Slavic (Russian, Ukrainian, Belarusian), and South Slavic (Bosnian, Croatian, Serbian, Bulgarian, Macedonian, Slovenian), consistent with known genealogical grouping. A more fine-grained look at the Indo-European subclusterings can be found in the following section.

\subsection{Indo-European case study: agreement and representational divergence}
\label{sec:results_ie}

\begin{figure}[!htbp]
	\centering
    \includegraphics[width=0.9\textwidth]{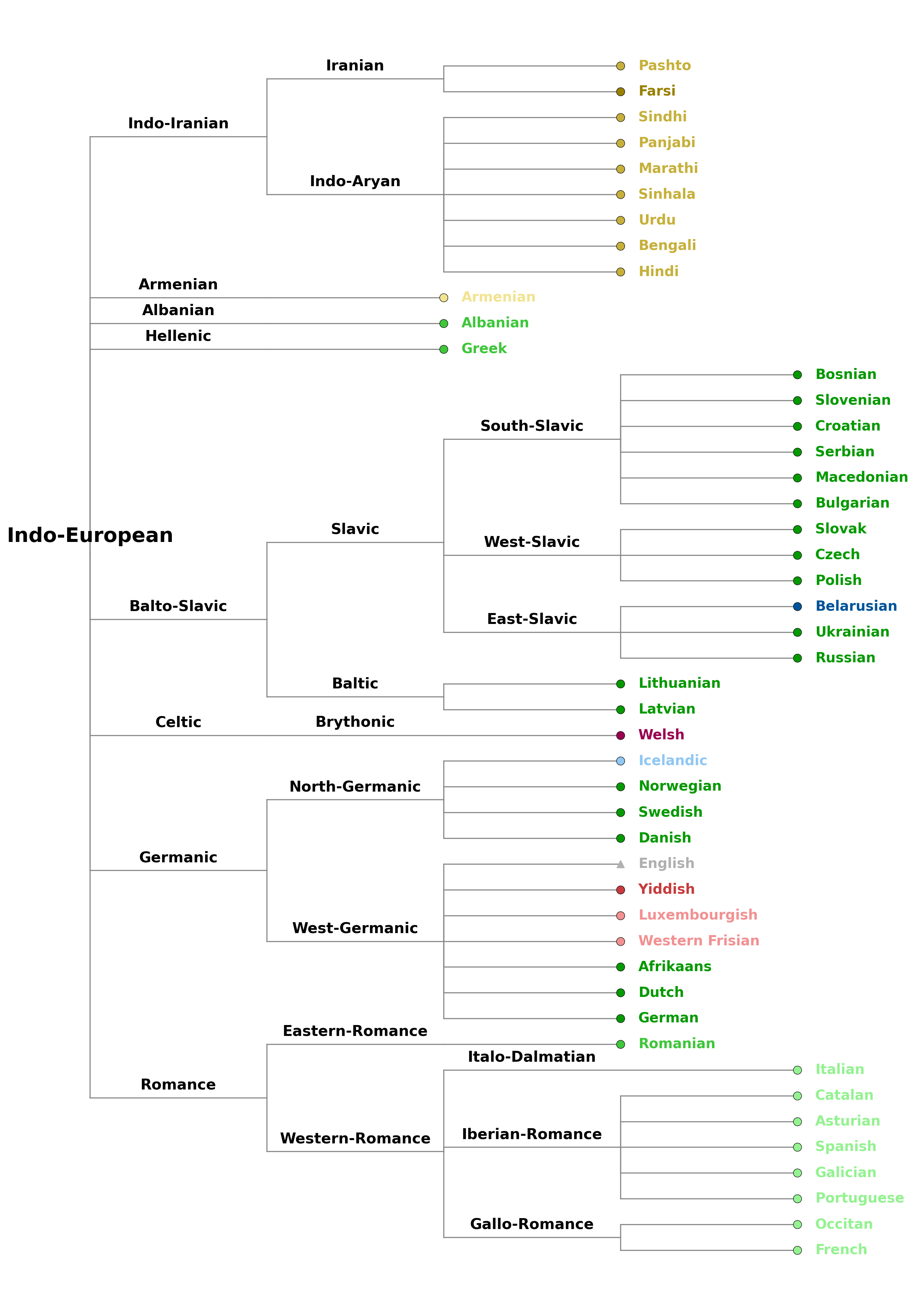}
    \caption{Established Indo-European language tree annotated with ATD-based cluster assignments.}
    \label{fig:nj_tree_indo}

\end{figure}

To examine correspondence between ATD-derived structure and fine-grained genealogical organization, we project our ATD-based cluster assignments onto an established reference tree of the Indo-European family (Fig.~\ref{fig:nj_tree_indo}). Each language is colored by its ATD-derived cluster label. To support multiscale interpretation, internal nodes of the reference tree are colored by a weighted aggregation of their descendant languages. This visualization strategy allows coarse structure to be identified through dominant internal-node colors, while preserving finer-grained variation among individual languages.

At the coarsest level, the Indo-European tree exhibits a pronounced separation into two major ATD macro-clusters. One macro-cluster is dominated by European Indo-European branches, including Romance, Germanic, and Slavic languages, while the other is dominated by Indo-Iranian languages together with geographically adjacent varieties. 

Within these macro-clusters, ATD further reveals meaningful internal structure that largely aligns with canonical subgrouping. Romance languages share consistent assignments, and Slavic languages remain internally coherent. At the same time, several languages form distinct fine-grained subclusters while remaining embedded within a broader genealogical group. For example, Luxembourgish and West Frisian appear as a coherent subcluster clearly differentiated from other West Germanic languages, yet still merge with the broader Germanic macro-cluster at higher internal nodes. Icelandic exhibits a similar pattern: although genealogically North Germanic, it forms a distinct subcluster separated from Mainland Scandinavian languages, reflecting its conservative morphology, relative isolation, and distinctive orthographic conventions. These cases illustrate how ATD captures graded representational similarity rather than enforcing hard categorical boundaries.

We also observe informative local divergences relative to a purely genealogical partition. A particularly salient case is Belarusian, which is genealogically classified as East Slavic but emerges as a distinct ATD subcluster with larger distances from Russian and Ukrainian. This separation likely reflects a combination of sociolinguistic factors, language contact, orthographic conventions, and uneven representation in translation corpora.

Overall, the Indo-European projection demonstrates that ATD captures a multiscale and continuous notion of language relatedness. It preserves broad genealogical structure at the macro level, while simultaneously revealing finer substructures and representational divergences compatible with known typological and contact-driven effects.

\subsection{Geographic consistency of ATD-derived structure}
\label{sec:results_geo}

Beyond genealogical families, ATD also reflects areal convergence, which for the most part correspond to sharing of vocabulary (and some surface-level structural features) between languages spoken in proximity of each other. As a result, we can observe ATD proximity between
non-genealogically related languages. This is the case, for example, of the Uralic languages Finnish, Estonian, and Hungarian clustered together (as seen back in Fig. \ref{fig:nj_tree_all}) with their neighboring Baltic languages (Lithuanian and Latvian) and the Slavic branch, all from the Indo-European language family. Furthermore, Finnish and Estonian (Uralic) are closer to neighboring Lithuanian and Latvian (Indo-European) than to Hungarian, which is geographically separated from it. 
The relative ATD-closeness of unrelated neighboring languages relative to related ones suggests that ATD is particularly sensitive to lexical similarities, which ultimately can have a stronger impact than the features inherited from genealogical relatedness.

\begin{figure}[!htbp]
	\centering
	\includegraphics[width=\linewidth]{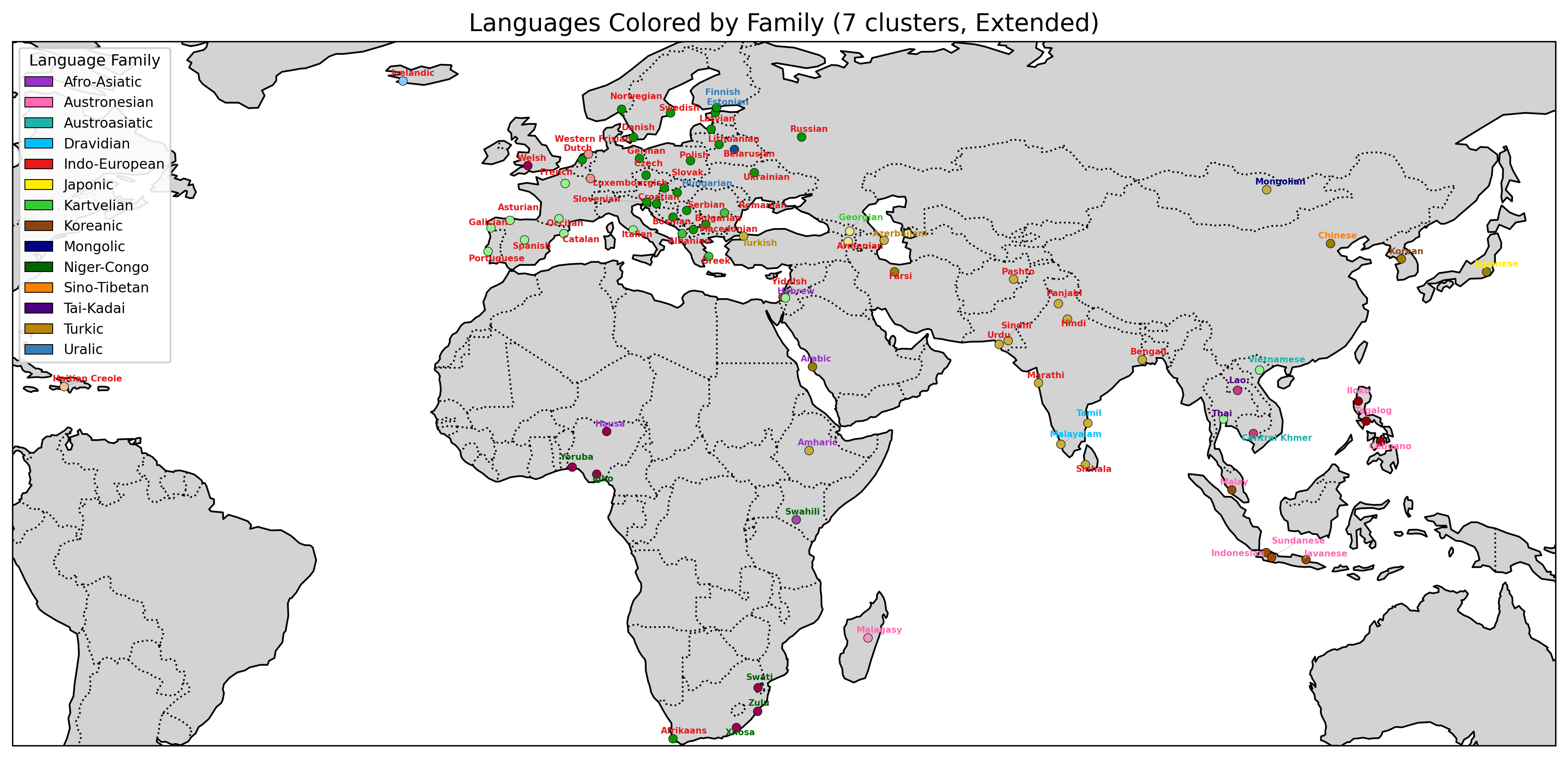}
	\caption{\textbf{Geographic distribution of ATD-derived language clusters.}
	Languages are shown at their approximate geographic locations, with markers colored according to ATD-based cluster assignments obtained from the M2M-100 model, using the same cluster definitions as in Fig.~\ref{fig:nj_tree_all}. Text labels indicate established linguistic family classifications. This visualization provides an external reference for assessing the alignment between model-derived language structure and known geographic and genealogical patterns.} 
	\label{fig:geo_map}
\end{figure}

To visualize language contact effects in ATD-derived language similarity, we project the resulting clusters onto a geographic map using the approximate locations of the corresponding languages (Fig.~\ref{fig:geo_map}). Languages are shown as nodes colored by their ATD-based cluster assignments, while text labels indicate established genealogical family classifications. This visualization provides an independent reference for evaluating whether model-derived language relationships align with known geographic and historical patterns.

We observe that ATD-derived clusters exhibit a strong degree of geographic coherence. For example, languages belonging to major Indo-European subgroups cluster tightly within Europe,
where Slavic and Romance languages appear both geographically and structurally proximate, consistent with well-established patterns of regional relatedness.
Similarly, two major representative subgroups of the Malayo-Polynesian family are found in geographically consistent ATD clusters, corresponding to Indonesian and Philipino languages. 

We can also find many examples of unrelated but geographically proximal languages being grouped together. Dravidian languages (Tamil and Malayalam) clustering with the Indo-Aryan branch of the Indo-European family, presumably due to shared vocabulary through language contact in India. 
Other examples are Georgian and Armenian; Chinese, Japanese and Korean; Lao and Khmer; Vietnamese and Thai; Hausa and Niger-Congo.

Finally, it also seems like geographic isolation plays a role in ATD: Icelandic, Malagasy and Welsh are not clustered with their genealogical neighbors, perhaps due to language divergence from their most closely related languages.

\subsection{Orthographic effects and unexpected ATD values}

It is difficult to determine whether shared script affects ATD similarity between languages, since it usually correlates with influence of a language over another.
However, we can note that language relatedness is a stronger indicator of similarity: Urdu and Hindi are essentially the same language but are written in very different scripts, with Urdu using the Arabic script and Hindi uses Devanagari. ATD places Urdu closer to Hindi ($d = 1.27$) than to Farsi ($d = 1.49$), also written in the Arabic script, which is an indicator that language relatedness has stronger weight than script similarity. Similarly, Serbian and Croatian are considered the same language, but written in Cyrillic and Latin scripts, respectively, and are $d=1.02$ apart.
In comparison, Serbian is further from Russian ($d=1.19$), even though they share a script. However, Serbian is expected to be closer to Slovenian than Russian, but ATD with Slovenian is also at $d=1.19$. This suggests an effect of script, since Slovenian is written in the Latin alphabet.

Despite these general patterns, several ATD values deviate from linguistic expectations in systematic ways that reveal specific model biases. One involves languages whose isolation can be linked to identifiable model properties. Tamil and Amharic stand out in the heatmap, with uniformly dark rows indicating high distances to virtually all other languages. Tamil's nearest neighbor is still at $d = 2.53$, and Amharic's distance to Arabic reaches $d = 2.29$. Both languages have highly distinctive scripts and are among the lower-resource languages in M2M-100. This is further supported by Fig.~\ref{fig:translation_scores}, which shows that their translation quality scores are among the lowest across the selected languages, consistent with the model having failed to learn stable cross-lingual representations for these languages.

Although the NJ tree places Yiddish (cluster 1.2) close to the Austronesian branch (cluster 1.3), the raw ATD matrix tells a different story: Yiddish is closer to Western Frisian ($d = 1.55$) and Luxembourgish ($d = 1.61$) than to any Austronesian language ($d \geq 1.63$), though the margin is small. This narrow gap, combined with the fact that Yiddish and the Austronesian outliers share similarly high average distances to all other languages, allows the NJ's $Q$-criterion to override the pairwise signal and group them together. This highlights the necessity of validating tree topology against the underlying pairwise distance matrix.

Finally, some anomalies remain difficult to explain through simple data or script factors. Hebrew and Vietnamese ($d = 1.33$) are genealogically unrelated, yet both sit anomalously close to high-resource European languages. Their mutual proximity may be a byproduct of this shared positioning, possibly driven by script overlap in the case of Vietnamese and training data composition in the case of Hebrew, though we cannot confirm either explanation from the distance matrix alone. Welsh shows a similar diffuse profile, sitting at comparable distances from French, Tagalog, and Swahili ($d \approx 1.54$-$1.55$). Belarusian presents a different kind of unexplained anomaly: its distance to Russian ($d = 1.90$) is nearly twice the Czech--Slovak distance ($d = 0.91$), and its entire row is uniformly elevated compared to other Slavic languages. The precise cause is unclear, though it is worth noting that Belarusian does not rank among the higher-quality translation directions in M2M-100.

\subsection{Controlled Analysis of Word Order Effect in ATD}

\begin{figure}[h]
    \centering
    \includegraphics[width=1.0\linewidth]{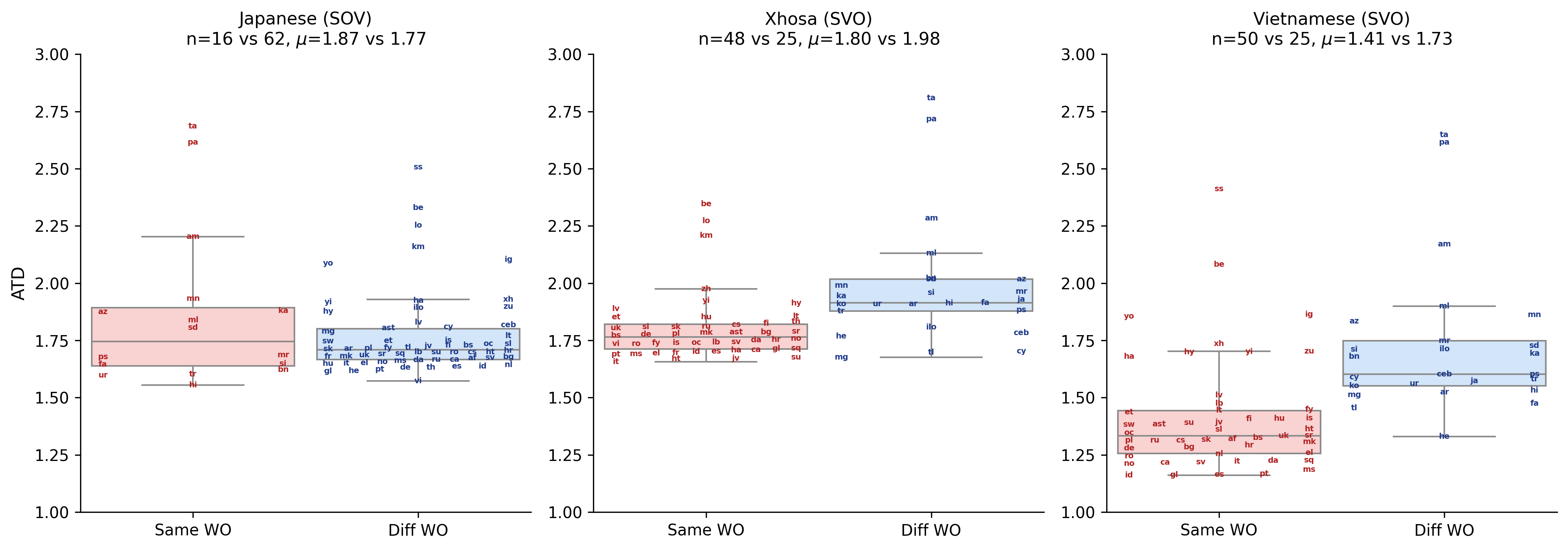}
    \caption{ATD comparison between languages with the same versus different word order relative to three source languages. Box plots show ATD distributions when translating from Japanese (SOV), Xhosa (SVO), and Vietnamese (SVO) to target languages grouped by word order congruence. Red boxes indicate target languages sharing the same dominant word order as the source; blue boxes indicate different word orders. Individual language codes are displayed as scatter points. Horizontal lines within boxes represent medians; box boundaries indicate interquartile ranges. Group sizes (n) and mean ATD values ($\mu$) are reported for each comparison.}
    \label{fig:atd_compare_box}
\end{figure}

Finally, we might observe non-genealogically related structural similarities, based on basic word order of the subject, object and verb, as shown in Fig.~\ref{fig:atd_compare_box}. This order is observed to correlate with a number of structural features in languages. For instance, SVO languages will typically have prepositions and adjectives before nouns, while SOV languages will typically have postpositions and adjectives after nouns. Guided by this knowledge we hypothesize that a language with a given word order will exhibit lower ATD with languages with its same word order, and higher ATD with those that don't. 

For instance, Turkish is SOV. We compare it to unrelated languages  Korean and Japanese, which are SOV, and Mandarin, Javanese, Swahili, which are SVO, and Malagasy, which is VOS.
It is 1.53 and 1.60 away from Korean and Japanese, while it is 1.77, 1.62, and 1.62 away from Mandarin, Swahili and Javanese. And it is 1.71 away from Malagasy.

We investigated whether typological similarity in word order influences attention patterns during neural machine translation by comparing ATD between source-target language pairs with the same versus different dominant word orders, after controlling for genetic relatedness and areal contact (see Appendix Table~\ref{tab:language_groupings_full} for complete language groupings). We used the Mann-Whitney U test, a non-parametric test suitable for comparing distributions without assuming normality, and report Cohen's $d$ as a standardized measure of effect size. The results reveal an asymmetric pattern depending on the source language's word order type. For Japanese (SOV), translations to other SOV languages ($n$=16, $\mu$=1.87) showed no significant difference compared to non-SOV languages ($n$=62, $\mu$=1.77; $p$=0.946, $d$=0.43). However, Xhosa (SVO) showed lower ATD when translating to SVO targets ($n$=48, $\mu$=1.80) than to non-SVO targets ($n$=25, $\mu$=1.98; $p<$0.001, $d$=$-$0.93), and this pattern was even more pronounced for Vietnamese (SVO: $n$=50, $\mu$=1.41; non-SVO: $n$=25, $\mu$=1.73; $p<$0.001, $d$=$-$1.15). 

The results for Xhosa and Vietnamese suggest ATD can be sensitive to structural features. What about the observation that SVO languages seem more similar to each other than SOV languages? It is premature to make any conclusions based on this small amount of data. However, the expectation from a linguist's point of view is that SOV languages are more similar to each other than SVO languages (whose more fine-grained word order directionality is more mixed than for SOV languages), which goes in the opposite direction to these results.

%% file: sections/main_transfer.tex
\subsection{Transfer Learning with ATD Regularization}
\label{sec:transfer}

Enhancing the performance of low-resource languages remains a persistent challenge in multilingual NLP. The severe data imbalance inherent in global corpora often prevents machine learning models from achieving equitable performance across the linguistic landscape. However, our preceding analyses establish that attention structures in multilingual models encode robust cross-linguistic regularities. A natural question follows: can we leverage this discovered structure to address the data scarcity bottleneck in low-resource translation? We hypothesize that by regularizing the attention distributions of a low-resource language to align with those of a linguistically related high-resource language, we can provide an effective inductive bias that improves translation quality without requiring external linguistic annotation.

Prior work on low-resource transfer learning has sought to regularize model behavior through different forms of structural constraints. At the parameter level, \cite{zoph2016transfer} pioneered transfer learning by initializing low-resource models with parameters from high-resource siblings, while~\cite{johnson2017google} and~\cite{arivazhagan2019massively} extended this to massively multilingual models where parameter sharing is implicit across languages. More explicitly, Universal Lexical Representation (ULR) combined with Mixture of Language Experts (MoLE) \cite{gu2018universal} learns family-specific lexical representations, operating on the premise that related languages should share learned parameters. At the alignment level, supervised attention methods \cite{liu2016neural} regularize model attention to match word alignments from external aligners, assuming attention should respect linguistically meaningful correspondences. However, these approaches either rely on implicit parameter sharing without leveraging discovered linguistic structure or depend on external alignment tools. In contrast, our ATD-based regularization directly operationalizes cross-linguistic structure encoded in attention patterns, constraining the attention geometry of low-resource languages to align with linguistically related high-resource languages based on discovered typological relationships.

To test this hypothesis, we focus on three representative low-resource languages: Pashto, Marathi, and Icelandic. Compared to high-resource counterparts like Farsi, these languages typically have parallel corpora that are several orders of magnitude smaller. As illustrated in Fig. \ref{fig:translation_scores}, this data scarcity leads directly to degraded performance, providing a principled testbed for our ATD-based regularization approach. Each of the target languages is paired with a linguistically related high-resource reference, namely Farsi (\texttt{fa}), Hindi (\texttt{hi}), and Norwegian (\texttt{no}), respectively. These pairings are linguistically motivated. Farsi and Pashto both belong to the Indo-Iranian branch and share core morphosyntactic properties such as SOV order, rich inflectional morphology, and postpositional phrase structure. Marathi and Hindi, both Indo-Aryan languages, exhibit high lexical and grammatical similarity. Icelandic and Norwegian, as North Germanic languages, share strong typological and lexical overlap while differing primarily in orthographic and phonological conventions. This configuration provides a typologically diverse framework to evaluate whether ATD regularization can effectively transfer structural priors across related languages.

\begin{figure}[h]
	\centering
	\includegraphics[width=\linewidth]{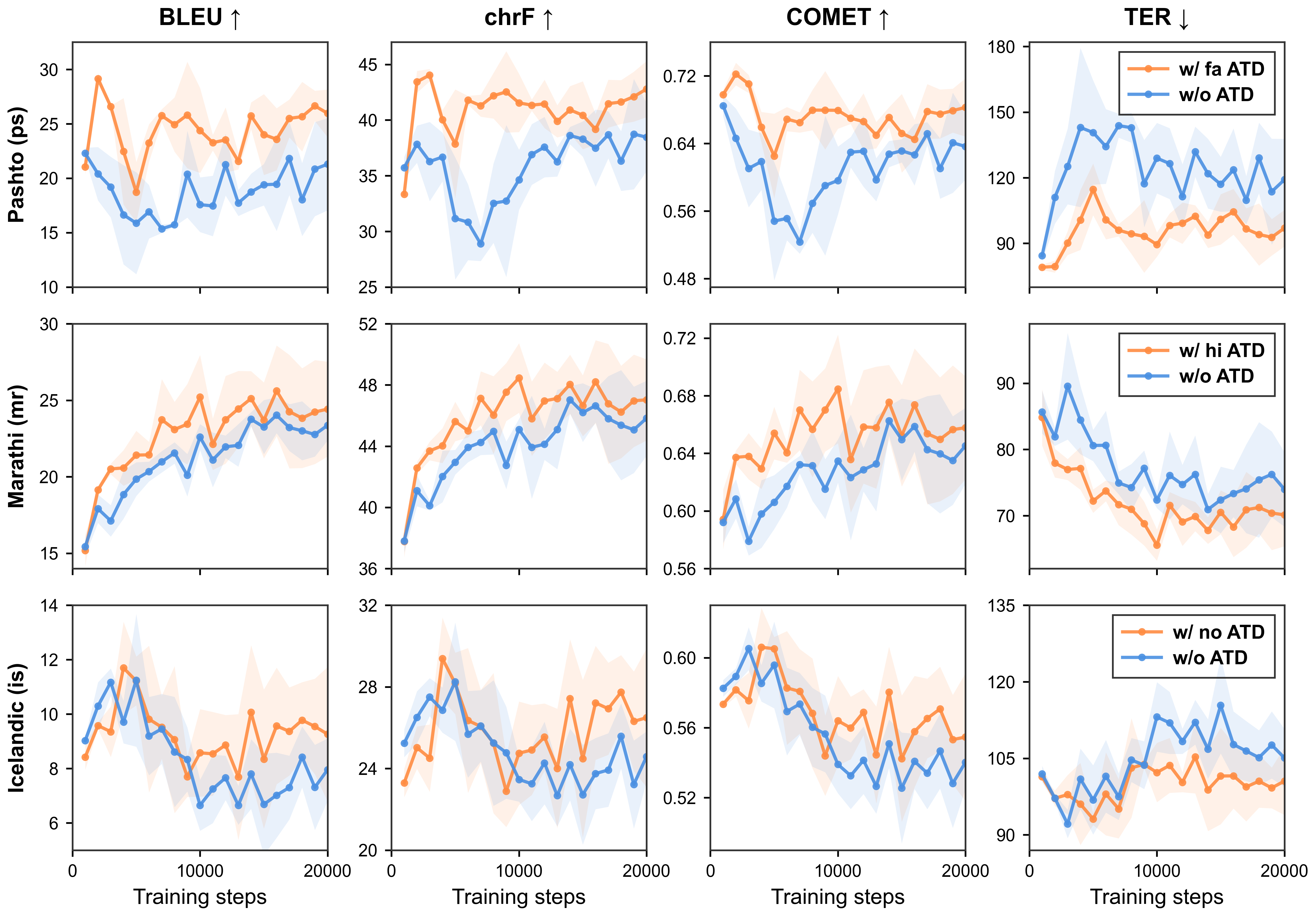}
    \caption{Translation quality during fine-tuning on three low-resource languages, comparing models trained with (w/) and without (w/o) ATD regularization toward a related high-resource reference. Shaded regions show standard deviation across three random seeds. Arrows ($\uparrow/\downarrow$) indicate the direction of improvement.}
	\label{fig:transfer}
\end{figure}

We initialize from the pretrained M2M-100 model~\cite{fan2021beyond} and fine-tune each low-resource language on its corresponding English-parallel subset of OPUS-100, with or without ATD regularization. The training objective is:
\begin{equation}
	\mathcal{L} = \mathcal{L}_{\text{CE}} + \lambda \, \mathcal{L}_{\text{ATD}},
\end{equation}
where $\mathcal{L}_{\text{CE}}$ is the standard cross-entropy loss, and $\mathcal{L}_{\text{ATD}}$ penalizes discrepancies between the low-resource and reference language attention distributions. We use identical hyperparameters for both models, employing uniform layer weighting and no additional tuning to isolate the effect of the regularization mechanism itself. All models are trained for 20k steps on the available training pairs (up to 50k) and evaluated on 2k held-out test pairs. Implementation details and additional experiments are provided in Appendix~\ref{app:transfer_appendix}.

Fig.~\ref{fig:transfer} shows that throughout the entire training trajectory, the ATD-regularized model consistently outperforms the baseline across all evaluation metrics: BLEU and chrF measure translation accuracy at the form level, TER measures error rate, and COMET measures semantic fidelity. The consistent advantage across training steps, not merely at convergence, indicates stable regularization from early training onward.
For clarity, the caption annotates each metric with upward ($\uparrow$) or downward ($\downarrow$) arrows to indicate the direction of improvement.
Overall, these results demonstrate that ATD regularization effectively enhances transfer learning for low-resource languages by aligning their attention geometry with related high-resource counterparts. This simple yet principled mechanism validates our central hypothesis and suggests promising directions for future work, including adaptive regularization and multi-language supervision.

%% file: sections/main_discussion.tex
\section{Discussion}\label{sec13}

This study provides evidence that pretrained multilingual language models implicitly encode nontrivial linguistic structure across languages. Rather than treating such structure as an abstract or anecdotal property of large models, we show that it can be systematically exposed and analyzed through their internal attention patterns. To this end, we introduce ATD, a tokenization-agnostic, attention-based quantitative framework that transforms latent cross-linguistic relations learned by pretrained models into interpretable distance matrices. ATD thus offers a principled lens for probing how linguistic structure is organized within model representations.

Distances derived from ATD reveal structured and reusable cross-linguistic relationships encoded in the attention patterns of pretrained multilingual models. These distances recover hierarchical groupings that closely align with established typological and genealogical classifications, while also highlighting deviations that likely reflect language contact, areal effects, or representational biases introduced during training. The resulting tree and distance structures should not be interpreted as reconstructions of historical causality, but rather as representations of functional similarity as encoded by the model. 
The alignment with geneological structure means that ATD is highly sensitive to 
the presence of cognates, which are words from different languages with a common ancestor. Cognate sets between languages are not always identified by shared sounds/characters, but are instead 
characterized by systematic sound correspondences, which must be information ATD picks up on. Similarly, ATD must be sensitive to shared vocabulary through language contact, which also exhibits a high degree of sound correspondences. This is especially notable when the languages use different scripts, such as the Dravidian and Indo-Aryan languages: in these cases, character correspondences must be established by ATD.

ATD thus provides a quantitative interface between traditional linguistic knowledge and model-internal representations, enabling systematic comparison across languages at multiple levels of granularity. As such, it is an exciting new tool that can contribute to linguistic theory. The results presented in this paper are an independent confirmation of findings in historical and contact linguistics, and show promise that ATD can be applied to answering current questions in these fields and raising new ones. For instance, ATD could be useful for gauging the relative contact influence of two related languages to an unrelated third one, in
more obscure contact situations where historical sources and linguistic data are insufficient.
On the other hand, the unexpected ATD values for some linguistic relationships bring up questions that went previously unnoticed by linguists. For instance, why is Welsh closer to Niger-Congo languages than Indo-European ones? Icelandic and Norwegian were the same language only a thousand years ago, yet their ATD is unexpectedly large. ATD could thus help determine the pace of language change when  a language is spoken in isolation of others, a question that doesn't have an obvious answer in linguistics. Further work applying ATD to these questions may reveal to be fruitful.

At the same time, the observation that ATD-based constraints can improve transfer performance in low-resource translation settings suggests that such attention-derived structure is not only interpretable, but also functionally relevant for cross-lingual generalization and model adaptation.

Several limitations should be noted. ATD captures model-internal functional distances rather than true historical, geographic, or sociolinguistic relationships, and its structure inevitably reflects biases and imbalances in the underlying training data. Moreover, the quality of attention-derived distances depends on the reliability of cross-lingual alignment, which varies across languages and model architectures. Our transfer learning experiments are intended as a demonstration that constraining or leveraging attention-derived structure can be beneficial, rather than as an optimized engineering solution. How such structural constraints can be most efficiently integrated into large-scale training or fine-tuning pipelines remains an open question for future work. Together, these considerations place important limits on the interpretation of ATD and motivate its use alongside external linguistic knowledge.

Future work may extend ATD to a broader range of model architectures and attention variants, including decoder-only large language models, for which we present preliminary results on Llama-3 in the Appendix. Beyond architectural differences, attention-derived distance measures could be examined across model scales and training stages to characterize how cross-linguistic structure emerges, stabilizes, or shifts during learning. More generally, ATD may be explored in multimodal settings such as speech, or leveraged to inform multilingual pretraining strategies and curriculum learning. Together, these directions may further clarify the relationship between attention mechanisms, representation learning, and linguistic structure.

Overall, this work shows that large-scale pretrained multilingual language models implicitly encode structured relationships among languages. By providing a quantitative framework to expose and analyze the linguistic structure embedded in attention, ATD highlights the potential of such models as scientific instruments. In doing so, it bridges modern AI systems and long-standing questions in language science, while offering practical insights for building more robust and inclusive multilingual technologies.

%% file: sections/appendix_method.tex
\section{Methodology}\label{sec:methods} \label{sec:appendix_method}

We divide our analysis into three stages: (i) task definition and attention extraction, (ii) tokenization-agnostic distance computation, and (iii) visualization and clustering.

\subsection{Task definition and attention extraction}
\label{ssec:extract}

Let
\begin{equation}\label{eq:source-set}
    \mathcal{M}=\{s_1,\dots,s_{S}\}, \qquad S=|\mathcal{M}|
\end{equation}
be a set of $S$ English sentences that spans a broad range of syntactic constructions and discourse functions.
Each \(s_m\) is drawn from the official test set of the \emph{WMT 2014 News Translation} task, which covers a broad spectrum of syntactic constructions and discourse functions.\footnote{The WMT14 English test set contains 3{,}003 sentences~\cite{bojar2014findings}.}

For each translation pair $(s_m, \ell_{m}^n)$, where $m = 1, \dots, S$ and $n = 1, \dots, N_L$ index the source English sentences and different target languages respectively, we record a four-dimensional cross-attention tensor
\begin{equation}
    A_{m,n} \in \mathbb{R}^{L \times H \times T_{\text{out}} \times T_{\text{in}}},
\end{equation}
where $L$ and $H$ denote the number of layers and attention heads, while $T_{\text{in}}$ and $T_{\text{out}}$ represent the number of source and target tokens, respectively. 
For the encoder–decoder model, M2M-100, we extract the \textit{cross-attention} block, which directly provides \(A_{m,n}\). For the decoder-only model, Llama-3-8B-Instruct, we cast translation as an instruction-following task: a fixed prompt requests the model to translate a given English sentence into the target language. During generation, we extract the self-attention weights that connect source tokens to their corresponding target tokens. The full prompt template and extraction protocol are documented in Appendix \ref{sec11}.

\subsection{Tokenization-agnostic attention distance}
\label{ssec:w2}

Sub-word tokenisers such as BPE~\cite{sennrich2016neural} or SentencePiece~\cite{kudo2018sentencepiece} introduce language-specific segmentation that can confound cross-lingual comparison.  To neutralize this factor we marginalize each tensor by averaging over \emph{target} tokens to obtain a source-side distribution:
\begin{equation}\label{eq:compress}
    \bar{a}_{l, h, m, n}(t) = \frac{1}{T_{\text{out}}} \sum_{t'=1}^{T_{\text{out}}} A_{l, h, m, n}(t', t), \quad t=1,\dots,T_{\text{in}}.
\end{equation}
The marginalized weights are normalized to form a valid discrete probability distribution $\tilde{a}_{l, h, m, n}$ such that $\sum_t \tilde{a}_{l, h, m, n}(t) = 1$.

To extract a universal linguistic signature from the model, we first compute a consensus attention distribution for each layer by averaging across all $H$ heads, and then aggregate the resulting distances across the entire corpus and all $L$ layers. The entry $D_{i,j}$ in the global symmetric distance matrix $\mathbf{D} \in \mathbb{R}^{N_L \times N_L}$ is defined as:
\begin{equation}\label{eq:aggregation}
    D_{i,j} = \frac{1}{S \cdot L} \sum_{m=1}^{S} \sum_{l=1}^{L}  W_2(\frac{1}{H} \sum_{h=1}^{H}\tilde{a}_{l, h, m, i}, \frac{1}{H} \sum_{h=1}^{H}\tilde{a}_{l, h, m, j}).
\end{equation}
This procedure filters out head-specific functional variations by calculating the distance between layer-wise expected distributions, while subsequent averaging over layers and sentences suppresses semantic noise to yield a robust measure of structural divergence.

We define the distance between these 1D distributions using the 2-Wasserstein metric. Under a squared Euclidean ground cost $C_{tt'} = (t-t')^2$, the distance $W_2$ is given by:
\begin{equation}\label{eq:w2}
    W_2(P, Q) = \left( \min_{\gamma \in \Pi(P, Q)} \sum_{t, t'} \gamma_{t, t'} (t-t')^2 \right)^{1/2},
\end{equation}
where $\Pi(P, Q)$ denotes the set of all valid couplings. Crucially, for distributions on a one-dimensional sequence, $W_2$ is computed efficiently and exactly via the $L_2$ difference of their cumulative distribution functions (CDFs).

Overall, $D_{i,j}$ behaves as a genuine metric: it is non-negative and vanishes only when the expected attention distributions coincide; it is symmetric ($D_{i,j} = D_{j,i}$); and it satisfies the triangle inequality. This metric property provides a rigorous foundation for the subsequent phylogenetic tree reconstruction and clustering analysis.

\subsection{Neighbor-Joining Tree and Hierarchical Clustering}
\label{app:nj}

To visualize the hierarchical organization of languages, we apply the Neighbor-Joining algorithm~\cite{saitou1987neighbor} to the ATD matrix. Let $\mathbf{D} \in \mathbb{R}^{N_L \times N_L}$ be the symmetric distance matrix where each entry $D_{i,j}$ represents the averaged ATD between languages $L_i$ and $L_j$. NJ is a bottom-up clustering method that iteratively transforms this distance matrix into a phylogenetic tree structure.

Starting with a set of $m$ active language nodes, we iteratively select the pair $(i, j)$ that minimizes the $Q$-criterion:
\begin{equation}
    Q(i, j) = (m-2)D_{i,j} - \sum_{k=1}^m D_{i,k} - \sum_{k=1}^m D_{j,k}
\end{equation}
This criterion balances the direct distance between two nodes against their average distance to all other nodes in the current set, effectively isolating ``neighbors" in the representation space.
Once the pair $(i, j)$ is selected, a new internal node $u$ is created. The branch lengths from $u$ to the merged nodes are calculated as:
\begin{equation}
    f_{i,u} = \frac{1}{2} D_{i,j} + \frac{1}{2(m-2)} \left( \sum_{k=1}^m D_{i,k} - \sum_{k=1}^m D_{j,k} \right)
\end{equation}
\begin{equation}
    f_{j,u} = D_{i,j} - f_{i,u}
\end{equation}
To ensure physical plausibility and account for numerical noise, we apply a non-negativity constraint $f = \max(0, f)$. The distance matrix is then updated by replacing $i$ and $j$ with the new node $u$, where the distance from $u$ to any remaining node $k$ is given by:
\begin{equation}
    D_{u,k} = \frac{1}{2}(D_{i,k} + D_{j,k} - D_{i,j})
\end{equation}
This process repeats until only two nodes remain, which are joined to form the final unrooted tree.

Rather than applying standard hierarchical clustering directly to the distance matrix, we perform clustering on the NJ tree topology to respect its derived branch lengths. We define a cluster as a subtree formed by cutting the tree at a specific cumulative depth threshold $d$ from the root. To determine the optimal threshold $d^*$, we implement a bisection search algorithm with a maximum of 50 iterations. The algorithm explores the search range between zero and the maximum tree depth; for each candidate $d$, it traverses the tree to identify nodes where the path length from the root first exceeds $d$, counts the resulting clusters $k$, and bisects the search interval according to our target $k_{\text{target}}=7$. This procedure converges on a threshold $d^*$ that yields the most stable partitioning of the language space within the prescribed iteration limit.

To capture finer linguistic nuances, our implementation supports a two-tier hierarchy. After identifying major clusters, we optionally partition them into sub-clusters (capped at a maximum of 3 per major cluster) using the same topological cutting logic. This ensures that while the macro-structure is preserved, highly diverse clusters can be inspected at a higher resolution.

%% file: sections/appendix_setting.tex
	\section{Experimental Setup}\label{sec11}

	\subsection{Benchmark Design}\label{ssec:benchmark}

	\textbf{Translation Task.} We use the publicly available WMT 2014 News Translation English test set (3{,}003 sentences) as our evaluation corpus for measuring translation behavior and extracting attention statistics~\cite{bojar2014findings}. We consider 99 target languages in the M2M-100 inventory \cite[Table~1]{fan2021beyond} excluding English. For each language, we generate translations from English and record the associated attention maps. To ensure reliable analysis, we score translation quality across the 100-language pool and retain only languages whose average score exceeds a preset threshold; the retained languages are listed in Table~\ref{tab:languages}.

\begingroup
\setlength{\LTpre}{0pt}\setlength{\LTpost}{0pt}
\renewcommand{\arraystretch}{1.1}
\begin{longtable}{@{}l >{\raggedright\arraybackslash}p{5.2cm} >{\raggedright\arraybackslash}p{5.5cm}@{}}
    \caption{Basic information for all languages reported in the results. Columns are: Code (ISO 639-1/3 identifier), Language Name (English exonym), and Family (Largest genealogical classification).}
    \label{tab:languages} \\
    \toprule
    \textbf{Code} & \textbf{Language Name} & \textbf{Family} \\
    \midrule
    \endfirsthead

    \multicolumn{3}{c}%
{{\bfseries \tablename\ \thetable{} -- continued from previous page}} \\
    \toprule
    \textbf{Code} & \textbf{Language Name} & \textbf{Family} \\
    \midrule
    \endhead

    \midrule \multicolumn{3}{r}{{Continued on next page}} \\
    \endfoot

    \bottomrule
    \endlastfoot

    af    & Afrikaans        & Indo-European (IE)   \\
    am    & Amharic          & Afroasiatic (AF)     \\
    ar    & Arabic           & Afroasiatic (AF)     \\
    ast   & Asturian         & Indo-European (IE)   \\
    az    & Azerbaijani      & Turkic (TR)          \\
    ba    & Bashkir          & Turkic (TR)          \\
    be    & Belarusian       & Indo-European (IE)   \\
    bg    & Bulgarian        & Indo-European (IE)   \\
    bn    & Bengali          & Indo-European (IE)   \\
    br    & Breton           & Indo-European (IE)   \\
    bs    & Bosnian          & Indo-European (IE)   \\
    ca    & Catalan          & Indo-European (IE)   \\
    ceb   & Cebuano          & Austronesian (AN)    \\
    cs    & Czech            & Indo-European (IE)   \\
    cy    & Welsh            & Indo-European (IE)   \\
    da    & Danish           & Indo-European (IE)   \\
    de    & German           & Indo-European (IE)   \\
    el    & Greek            & Indo-European (IE)   \\
    es    & Spanish          & Indo-European (IE)   \\
    et    & Estonian         & Uralic (UR)          \\
    fa    & Persian (Farsi)  & Indo-European (IE)   \\
    ff    & Fulah            & Niger-Congo (NC)     \\
    fi    & Finnish          & Uralic (UR)          \\
    fr    & French           & Indo-European (IE)   \\
    fy    & West Frisian     & Indo-European (IE)   \\
    ga    & Irish            & Indo-European (IE)   \\
    gd    & Scottish Gaelic  & Indo-European (IE)   \\
    gl    & Galician         & Indo-European (IE)   \\
    gu    & Gujarati         & Indo-European (IE)   \\
    ha    & Hausa            & Afroasiatic (AF)     \\
    he    & Hebrew           & Afroasiatic (AF)     \\
    hi    & Hindi            & Indo-European (IE)   \\
    hr    & Croatian         & Indo-European (IE)   \\
    ht    & Haitian Creole   & Creole/Pidgin (CP)   \\
    hu    & Hungarian        & Uralic (UR)          \\
    hy    & Armenian         & Indo-European (IE)   \\
    id    & Indonesian       & Austronesian (AN)    \\
    ig    & Igbo             & Niger-Congo (NC)     \\
    ilo   & Ilokano          & Austronesian (AN)    \\
    is    & Icelandic        & Indo-European (IE)   \\
    it    & Italian          & Indo-European (IE)   \\
    ja    & Japanese         & Japonic (JP)         \\
    jv    & Javanese         & Austronesian (AN)    \\
    ka    & Georgian         & Kartvelian (KK)      \\
    kk    & Kazakh           & Turkic (TR)          \\
    km    & Khmer            & Austroasiatic (AA)   \\
    kn    & Kannada          & Dravidian (DR)       \\
    ko    & Korean           & Koreanic (KO)        \\
    lb    & Luxembourgish    & Indo-European (IE)   \\
    lg    & Ganda            & Niger-Congo (NC)     \\
    ln    & Lingala          & Niger-Congo (NC)     \\
    lo    & Lao              & Tai-Kadai (TK)       \\
    lt    & Lithuanian       & Indo-European (IE)   \\
    lv    & Latvian          & Indo-European (IE)   \\
    mg    & Malagasy         & Austronesian (AN)    \\
    mk    & Macedonian       & Indo-European (IE)   \\
    ml    & Malayalam        & Dravidian (DR)       \\
    mn    & Mongolian        & Mongolic (MG)        \\
    mr    & Marathi          & Indo-European (IE)   \\
    ms    & Malay            & Austronesian (AN)    \\
    my    & Burmese          & Sino-Tibetan (ST)    \\
    ne    & Nepali           & Indo-European (IE)   \\
    nl    & Dutch            & Indo-European (IE)   \\
    no    & Norwegian        & Indo-European (IE)   \\
    ns    & Northern Sotho   & Niger-Congo (NC)     \\
    oc    & Occitan          & Indo-European (IE)   \\
    or    & Odia             & Indo-European (IE)   \\
    pa    & Punjabi          & Indo-European (IE)   \\
    pl    & Polish           & Indo-European (IE)   \\
    ps    & Pashto           & Indo-European (IE)   \\
    pt    & Portuguese       & Indo-European (IE)   \\
    ro    & Romanian         & Indo-European (IE)   \\
    ru    & Russian          & Indo-European (IE)   \\
    sd    & Sindhi           & Indo-European (IE)   \\
    si    & Sinhala          & Indo-European (IE)   \\
    sk    & Slovak           & Indo-European (IE)   \\
    sl    & Slovenian        & Indo-European (IE)   \\
    so    & Somali           & Afroasiatic (AF)     \\
    sq    & Albanian         & Indo-European (IE)   \\
    sr    & Serbian          & Indo-European (IE)   \\
    ss    & Swati            & Niger-Congo (NC)     \\
    su    & Sundanese        & Austronesian (AN)    \\
    sv    & Swedish          & Indo-European (IE)   \\
    sw    & Swahili          & Niger-Congo (NC)     \\
    ta    & Tamil            & Dravidian (DR)       \\
    th    & Thai             & Tai-Kadai (TK)       \\
    tl    & Tagalog          & Austronesian (AN)    \\
    tn    & Tswana           & Niger-Congo (NC)     \\
    tr    & Turkish          & Turkic (TR)          \\
    uk    & Ukrainian        & Indo-European (IE)   \\
    ur    & Urdu             & Indo-European (IE)   \\
    uz    & Uzbek            & Turkic (TR)          \\
    vi    & Vietnamese       & Austroasiatic (AA)   \\
    wo    & Wolof            & Niger-Congo (NC)     \\
    xh    & Xhosa            & Niger-Congo (NC)     \\
    yi    & Yiddish          & Indo-European (IE)   \\
    yo    & Yoruba           & Niger-Congo (NC)     \\
    zh    & Chinese          & Sino-Tibetan (ST)    \\
    zu    & Zulu             & Niger-Congo (NC)     \\
\end{longtable}
\endgroup
	\textbf{Translation Setup for M2M-100.}
	For the encoder-decoder model M2M-100, the translation task is straightforward. Given a source sentence in English and a specified target language, the model directly produces the translated sentence. Since M2M-100 is an encoder-decoder model trained with token-aligned multilingual corpora, we are also able to extract structured attention maps across all encoder and decoder layers directly from the model output.

	\textbf{Translation Prompting for Llama-3-8B-Instruct.}
	As Llama-3-8B-Instruct is a decoder-only general-purpose language model without explicit parallel translation supervision, we manually define a translation prompt to guide the model's generation. Specifically, we use the following template:

	\begin{verbatim}
Translate the following sentence from <src_lang> to <tgt_lang>. 
Only reply with the translated sentence, strictly using the format 
\'<START> translation <END>\'. 
Sentence to translate: <<original_text>> 
Here is the correct translation: <START>
\end{verbatim}

	We extract the model's response between the \texttt{<START>} and \texttt{<END>} tokens and retain it as the translation output. To study attention behavior, we further filter the attention maps to isolate the regions corresponding to the source and target spans. This allows us to analyze how Llama-3-8B-Instruct allocates attention in zero-shot translation scenarios, despite lacking explicit alignment training.

	\subsection{Attention Extraction}

\begin{figure}[!htbp]
		\centering

		\begin{subfigure}[b]{0.48\textwidth}
			\centering
			\includegraphics[width=\linewidth]{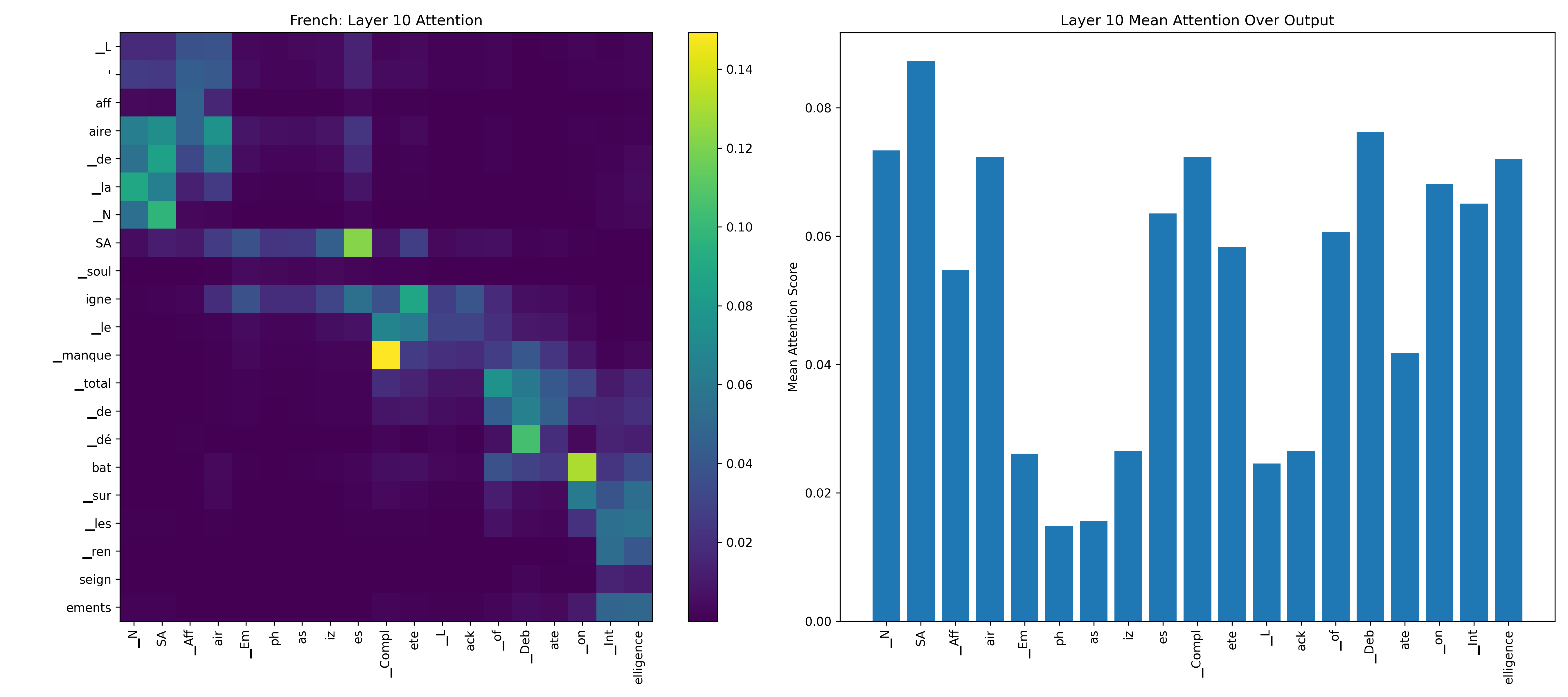}
			\caption{French}
		\end{subfigure}
		\hfill
		\begin{subfigure}[b]{0.48\textwidth}
			\centering
			\includegraphics[width=\linewidth]{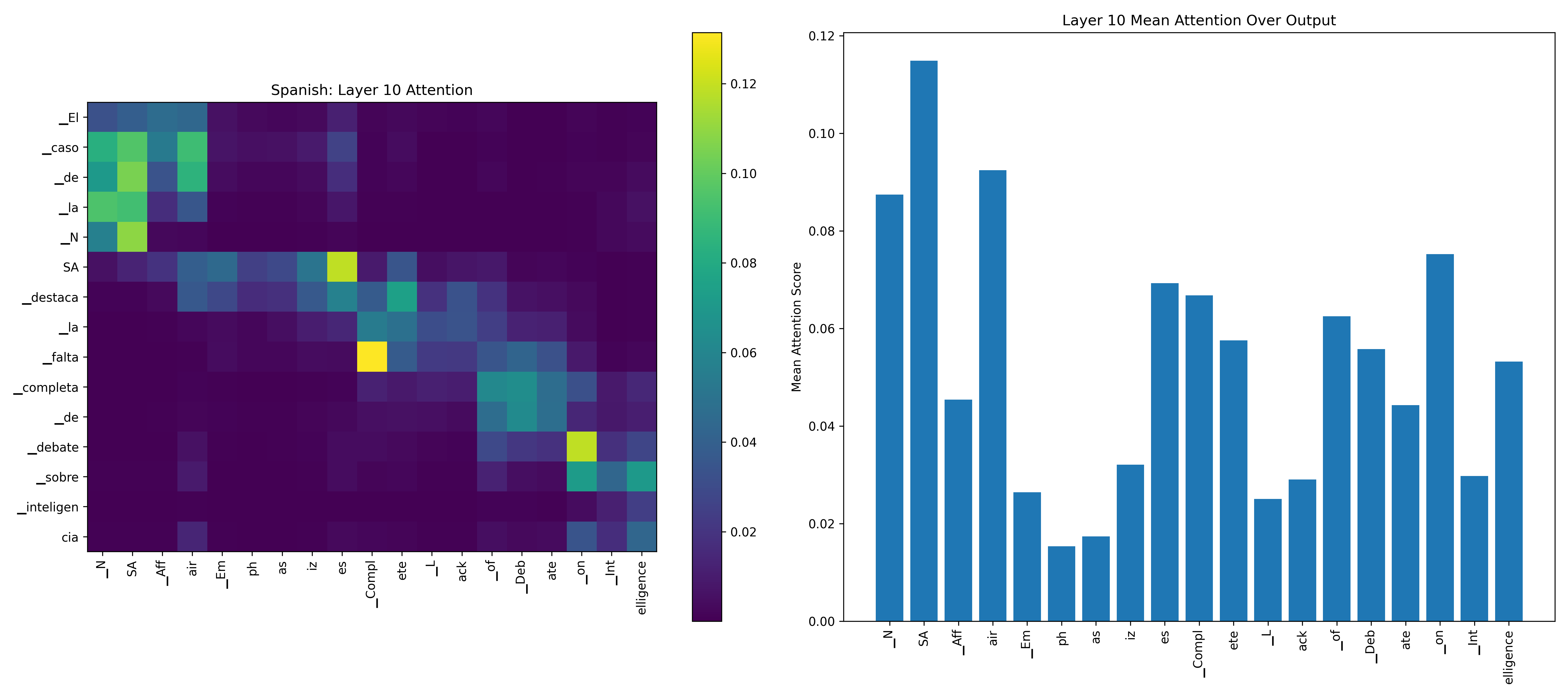}
			\caption{Spanish}
		\end{subfigure}

		\begin{subfigure}[b]{0.48\textwidth}
			\centering
			\includegraphics[width=\linewidth]{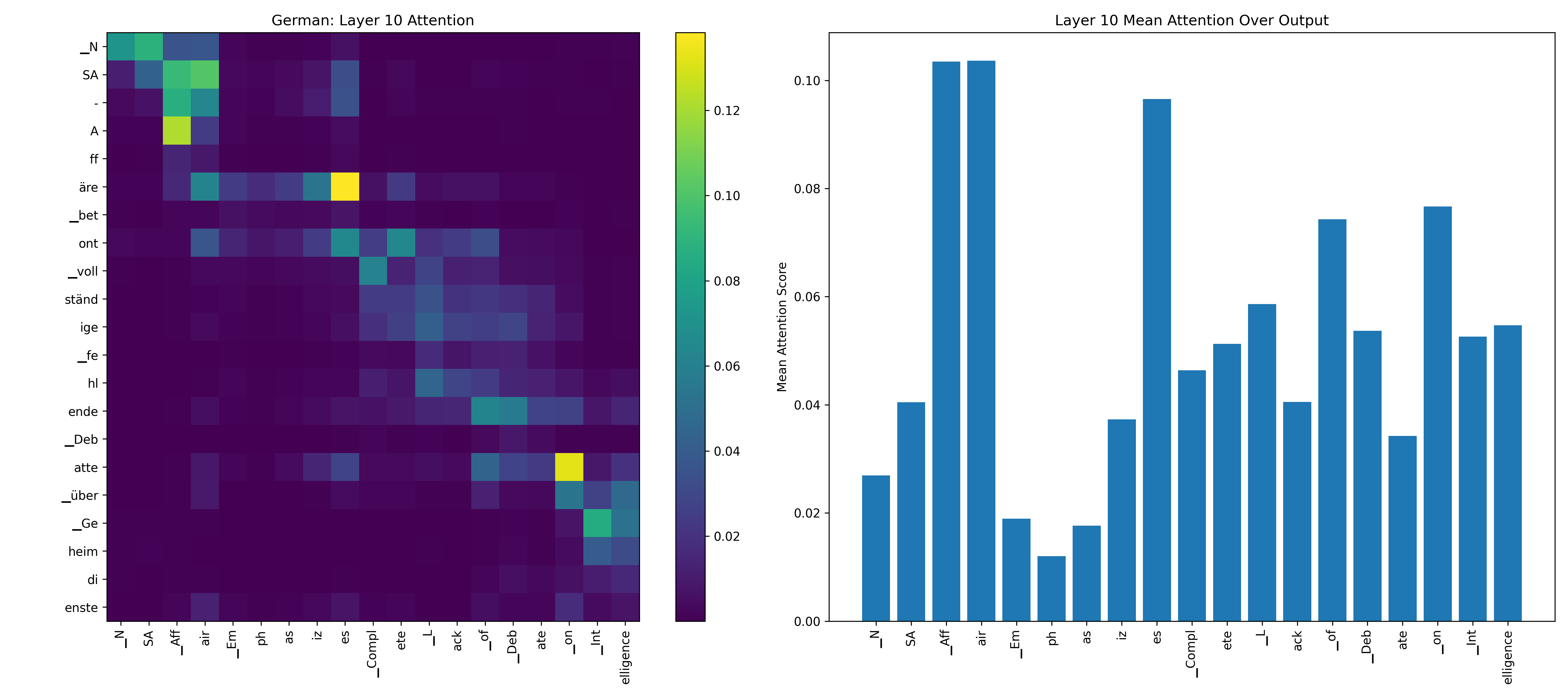}
			\caption{German}
		\end{subfigure}
		\hfill
		\begin{subfigure}[b]{0.48\textwidth}
			\centering
			\includegraphics[width=\linewidth]{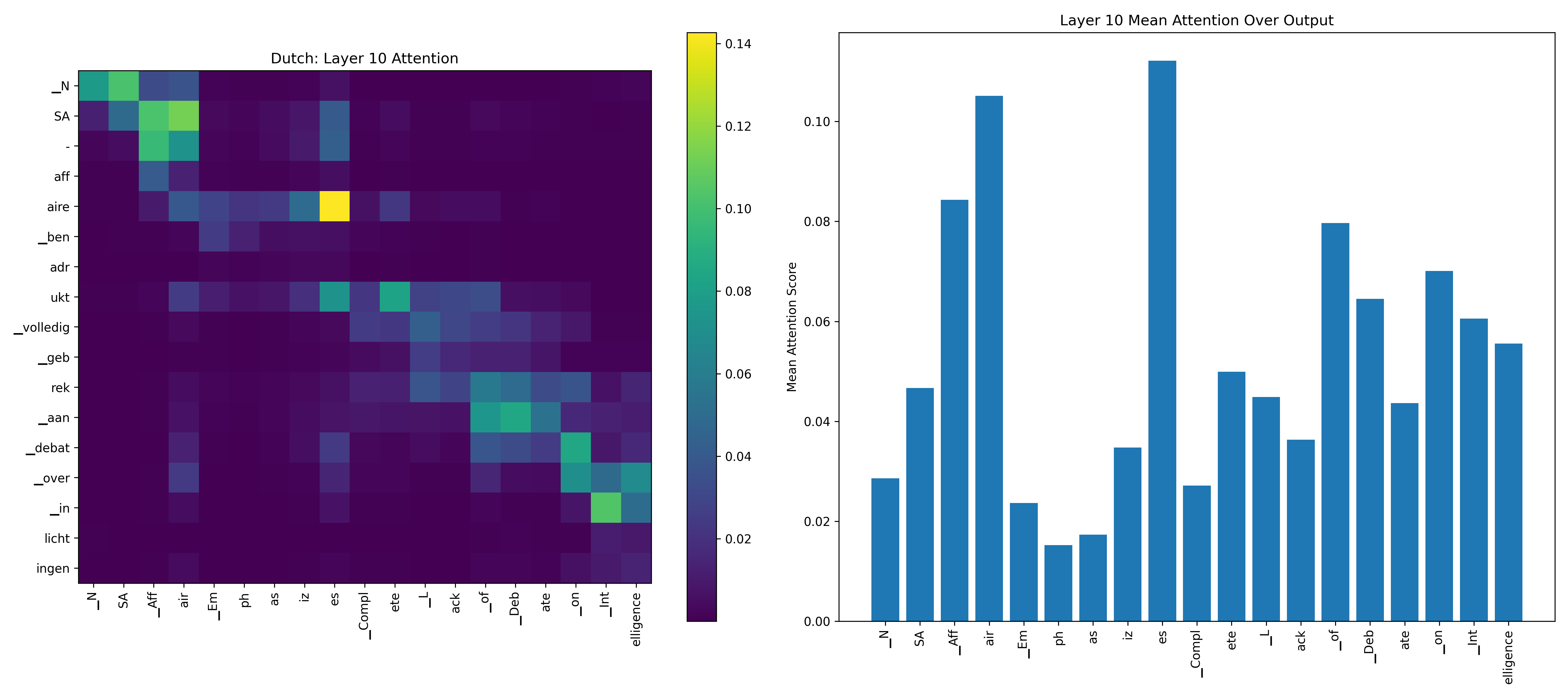}
			\caption{Dutch}
		\end{subfigure}

		\begin{subfigure}[b]{0.48\textwidth}
			\centering
			\includegraphics[width=\linewidth]{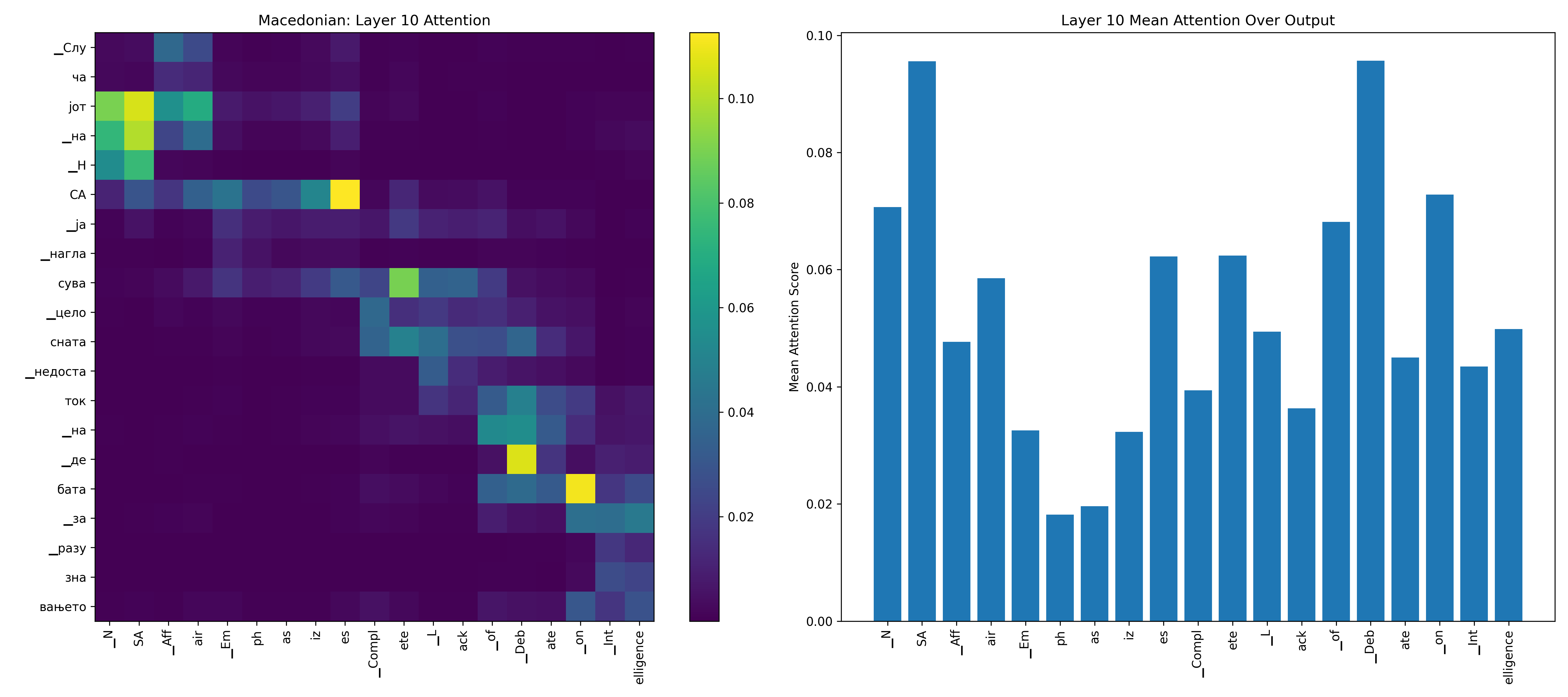}
			\caption{Macedonian}
		\end{subfigure}
		\hfill
		\begin{subfigure}[b]{0.48\textwidth}
			\centering
			\includegraphics[width=\linewidth]{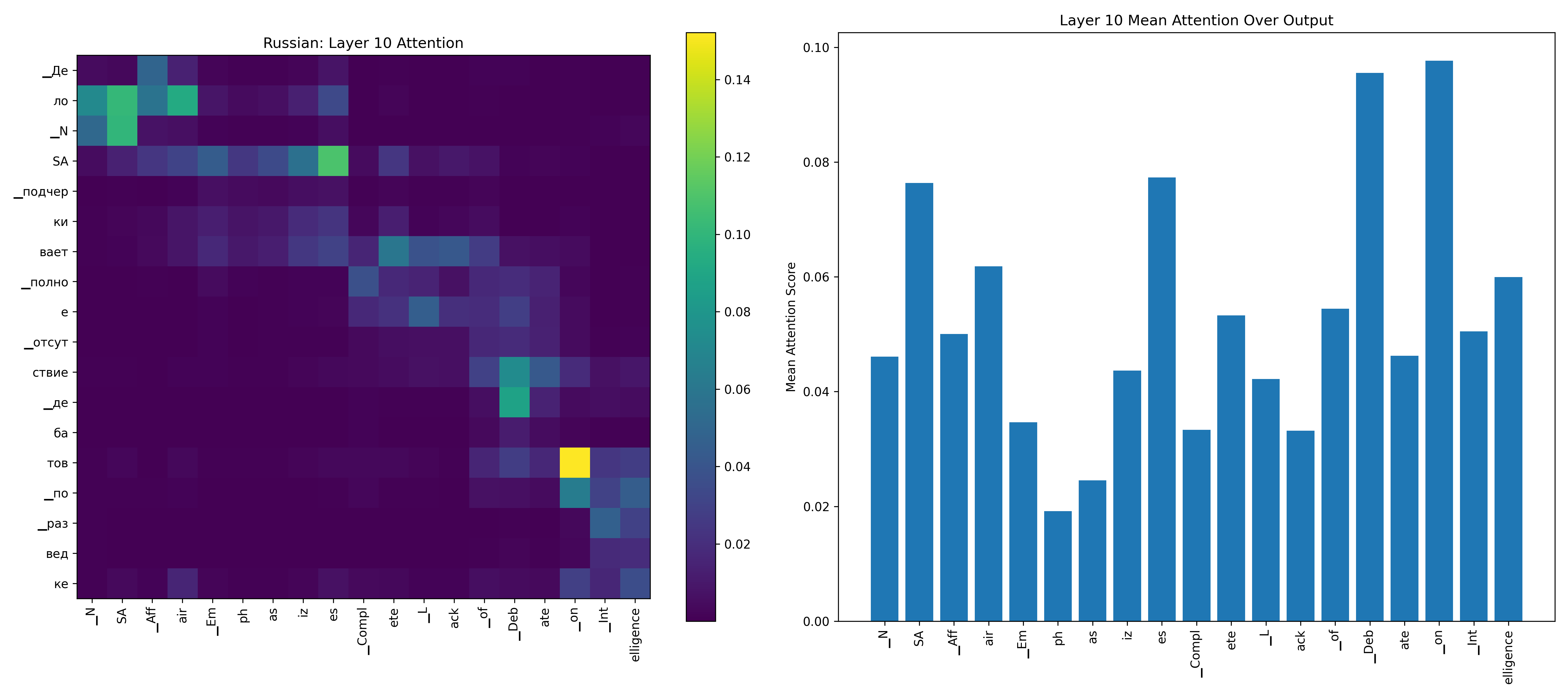}
			\caption{Russian}
		\end{subfigure}

		\begin{subfigure}[b]{0.48\textwidth}
			\centering
			\includegraphics[width=\linewidth]{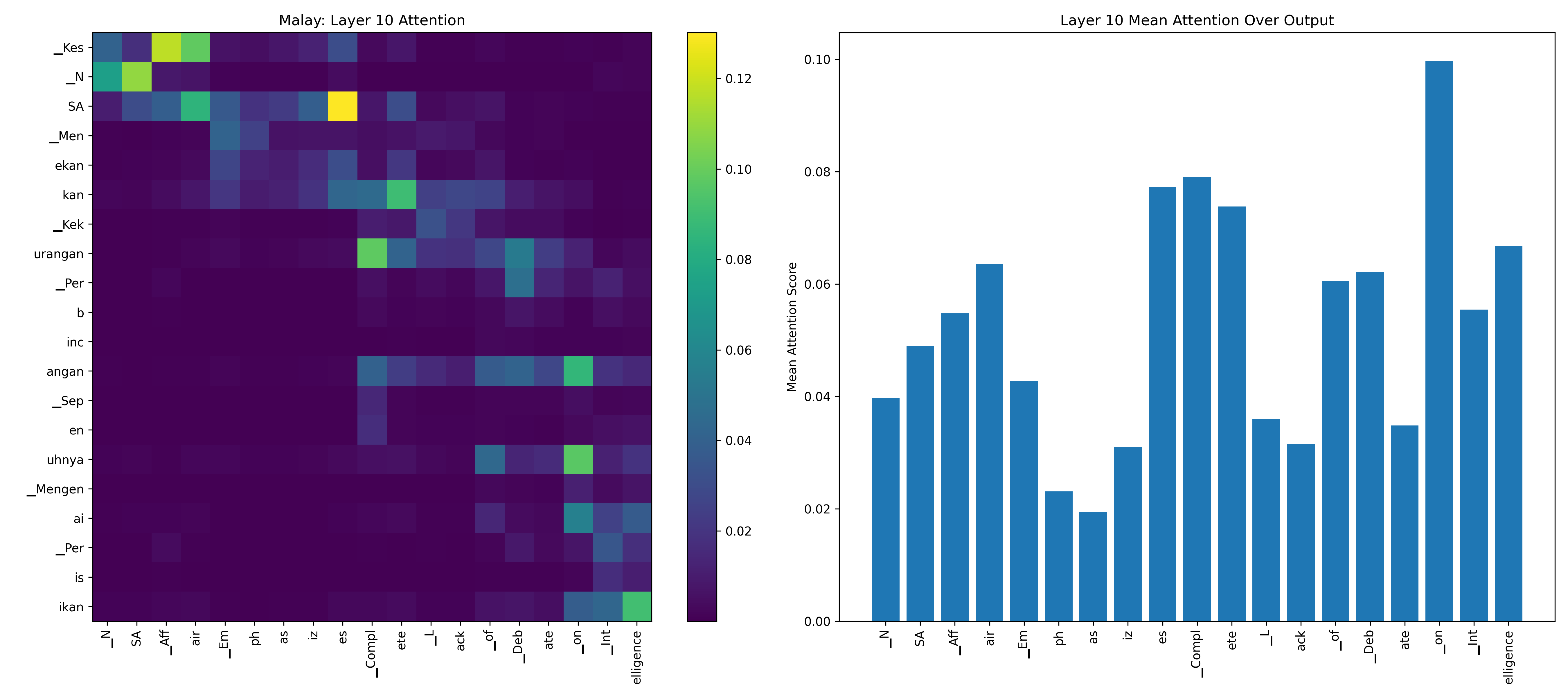}
			\caption{Malay}
		\end{subfigure}
		\hfill
		\begin{subfigure}[b]{0.48\textwidth}
			\centering
			\includegraphics[width=\linewidth]{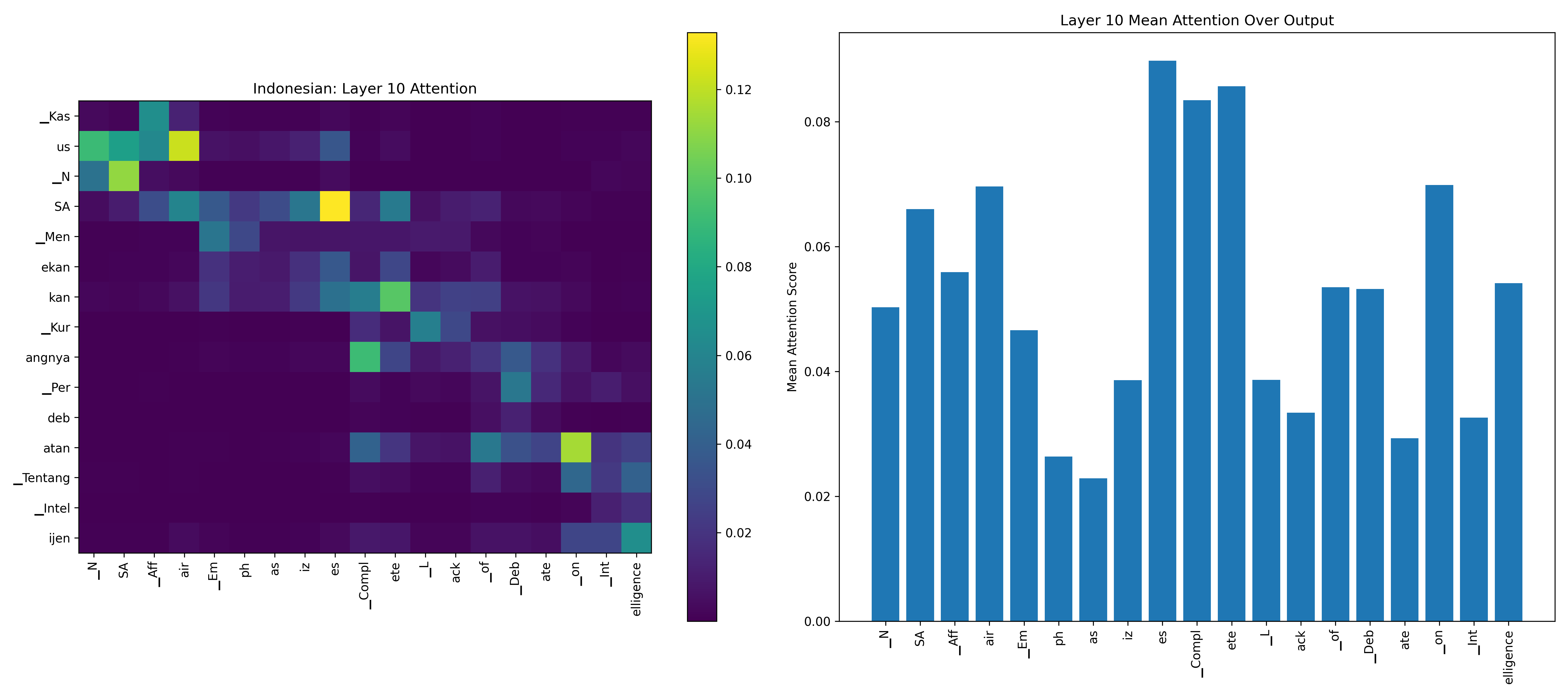}
			\caption{Indonesian}
		\end{subfigure}

		\begin{subfigure}[b]{0.48\textwidth}
			\centering
			\includegraphics[width=\linewidth]{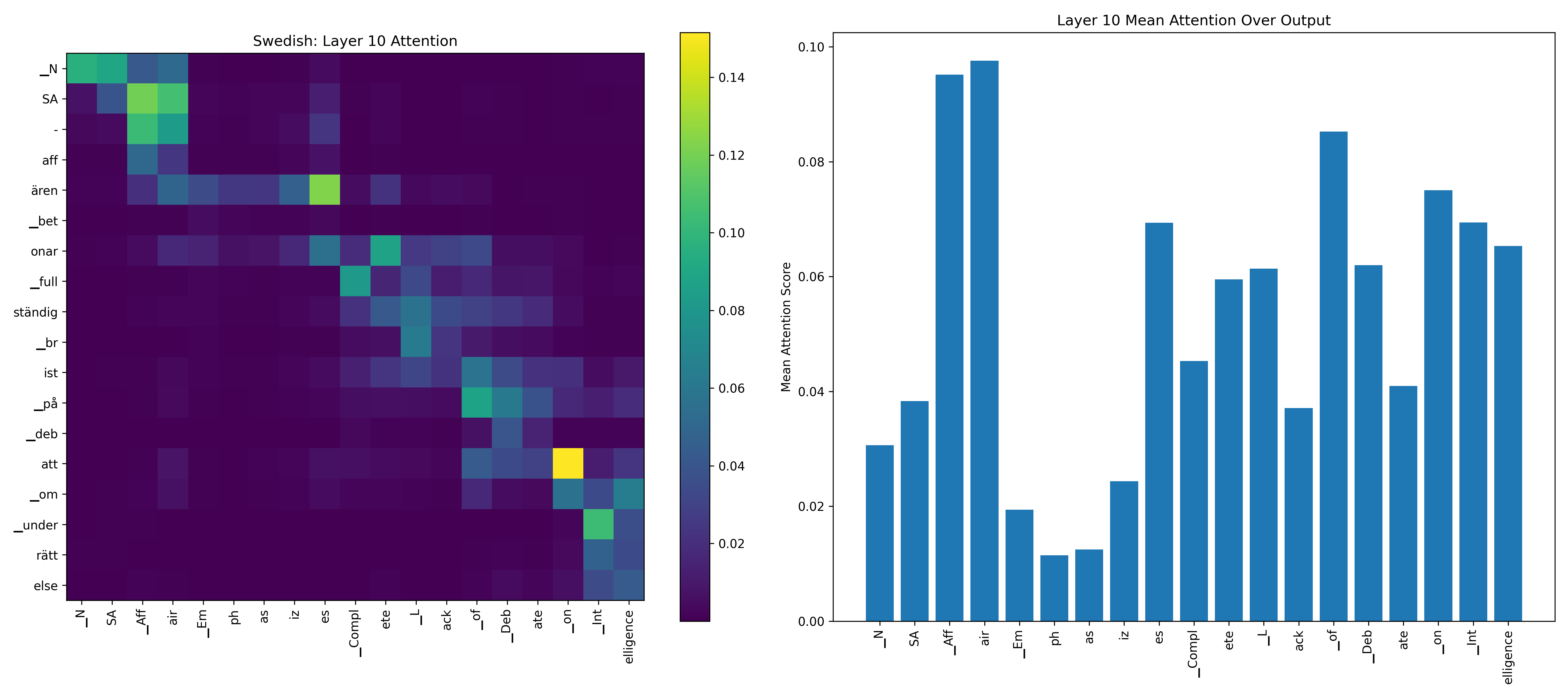}
			\caption{Swedish}
		\end{subfigure}
		\hfill
		\begin{subfigure}[b]{0.48\textwidth}
			\centering
			\includegraphics[width=\linewidth]{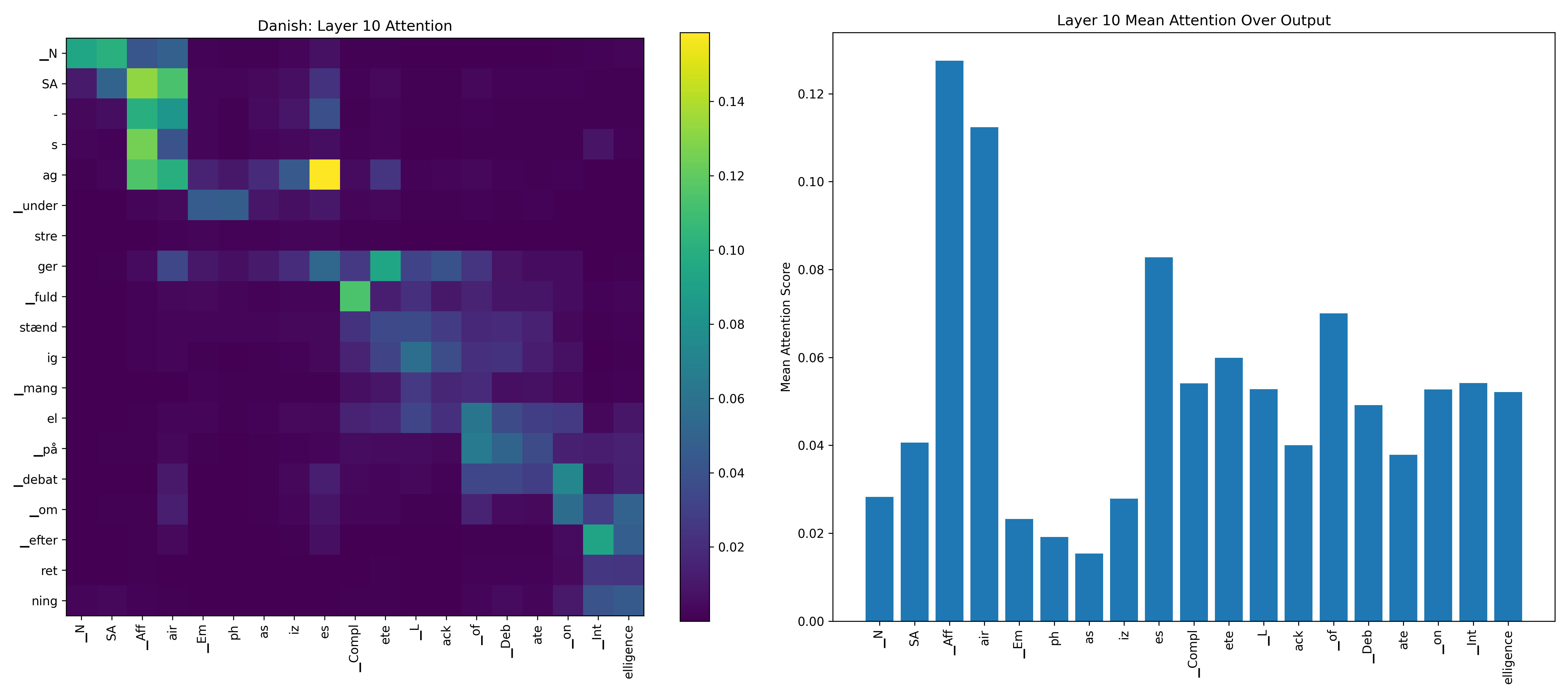}
			\caption{Danish}
		\end{subfigure}

		\caption{Attention maps for different languages in M2M-100}
		\label{fig:attention_m2m}
	\end{figure}

\begin{figure}[!htbp]
		\centering

		\begin{subfigure}[b]{0.48\textwidth}
			\centering
			\includegraphics[width=\linewidth]{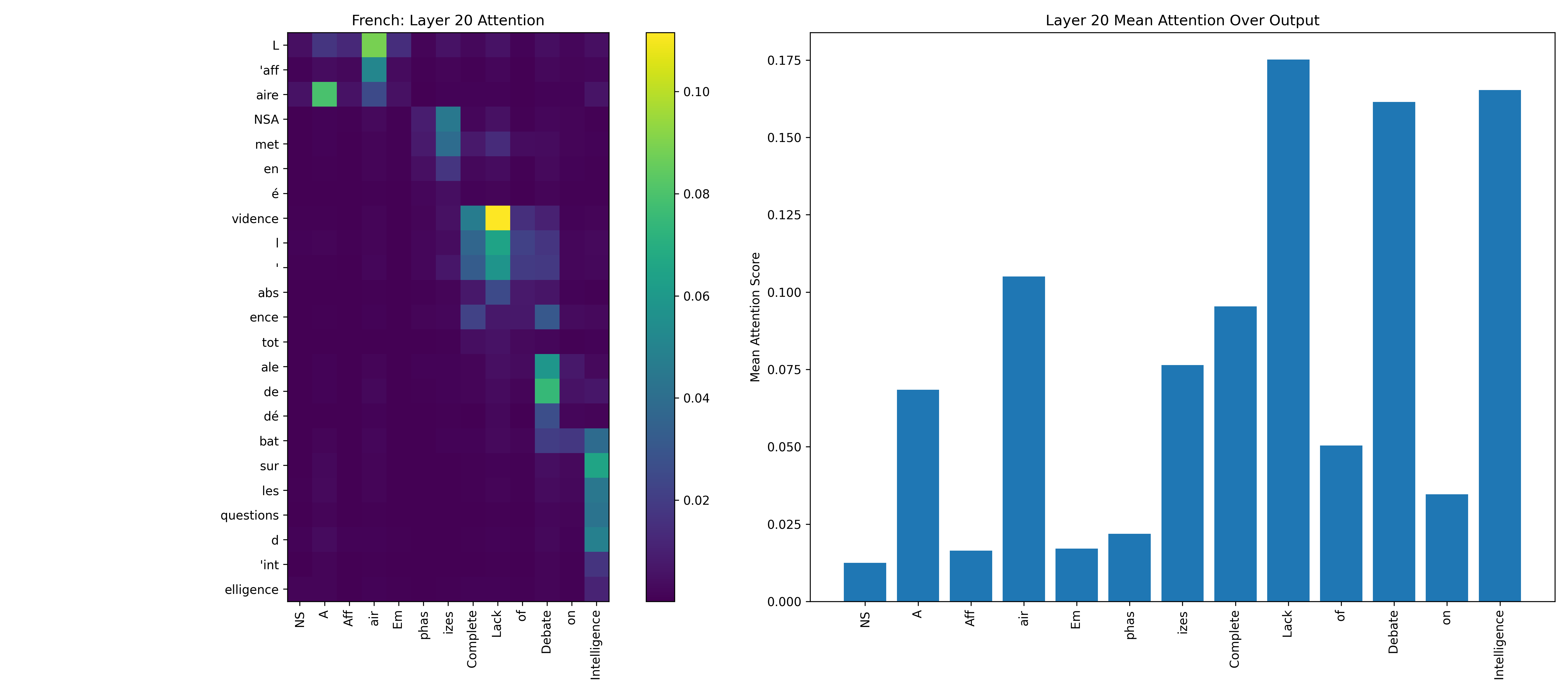}
			\caption{French}
		\end{subfigure}
		\hfill
		\begin{subfigure}[b]{0.48\textwidth}
			\centering
			\includegraphics[width=\linewidth]{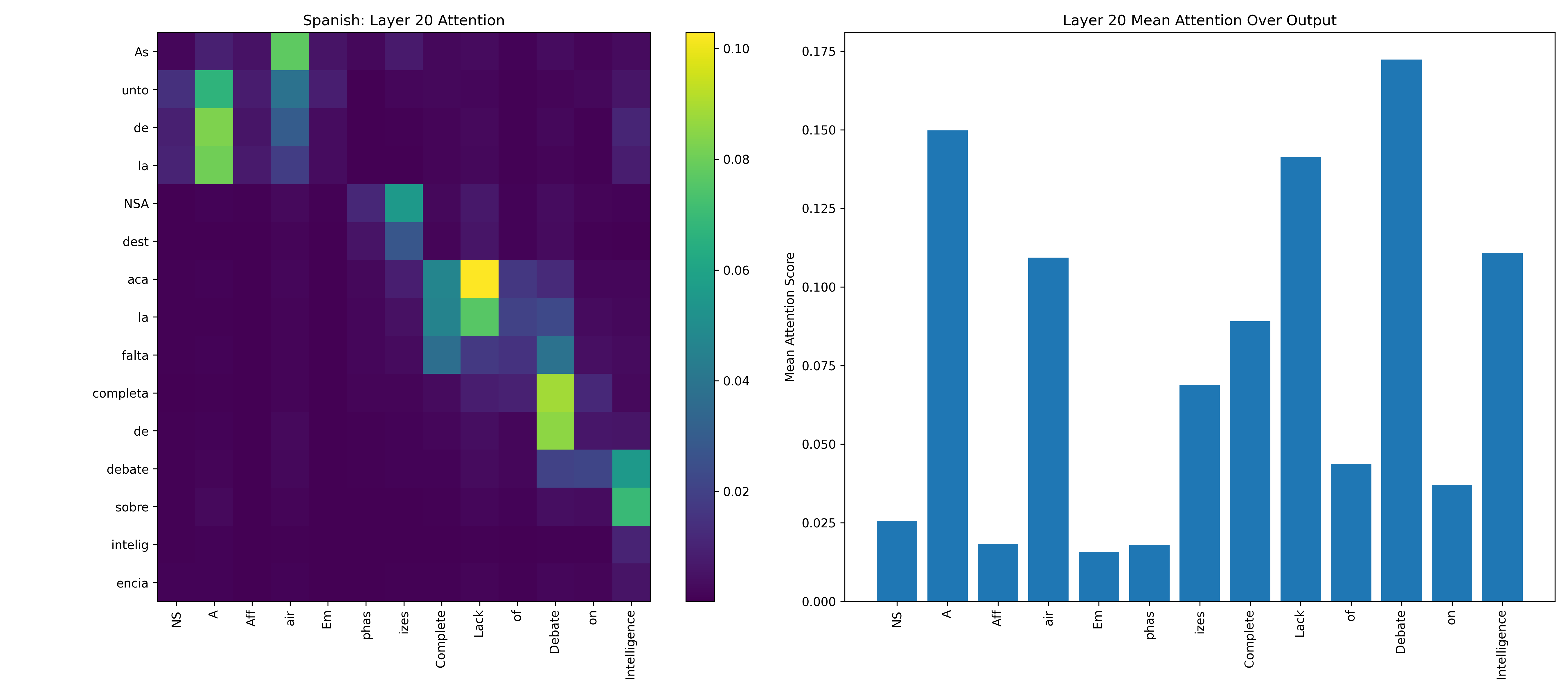}
			\caption{Spanish}
		\end{subfigure}

		\begin{subfigure}[b]{0.48\textwidth}
			\centering
			\includegraphics[width=\linewidth]{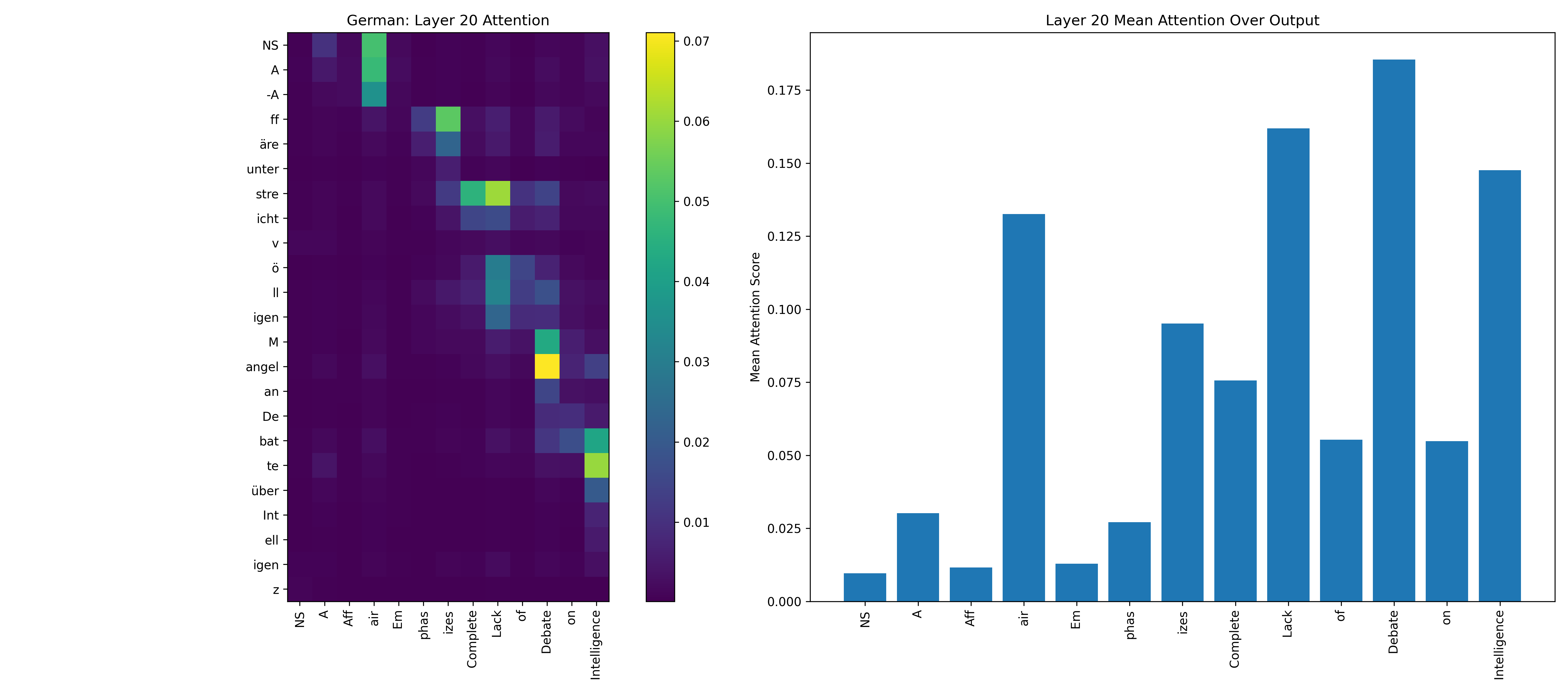}
			\caption{German}
		\end{subfigure}
		\hfill
		\begin{subfigure}[b]{0.48\textwidth}
			\centering
			\includegraphics[width=\linewidth]{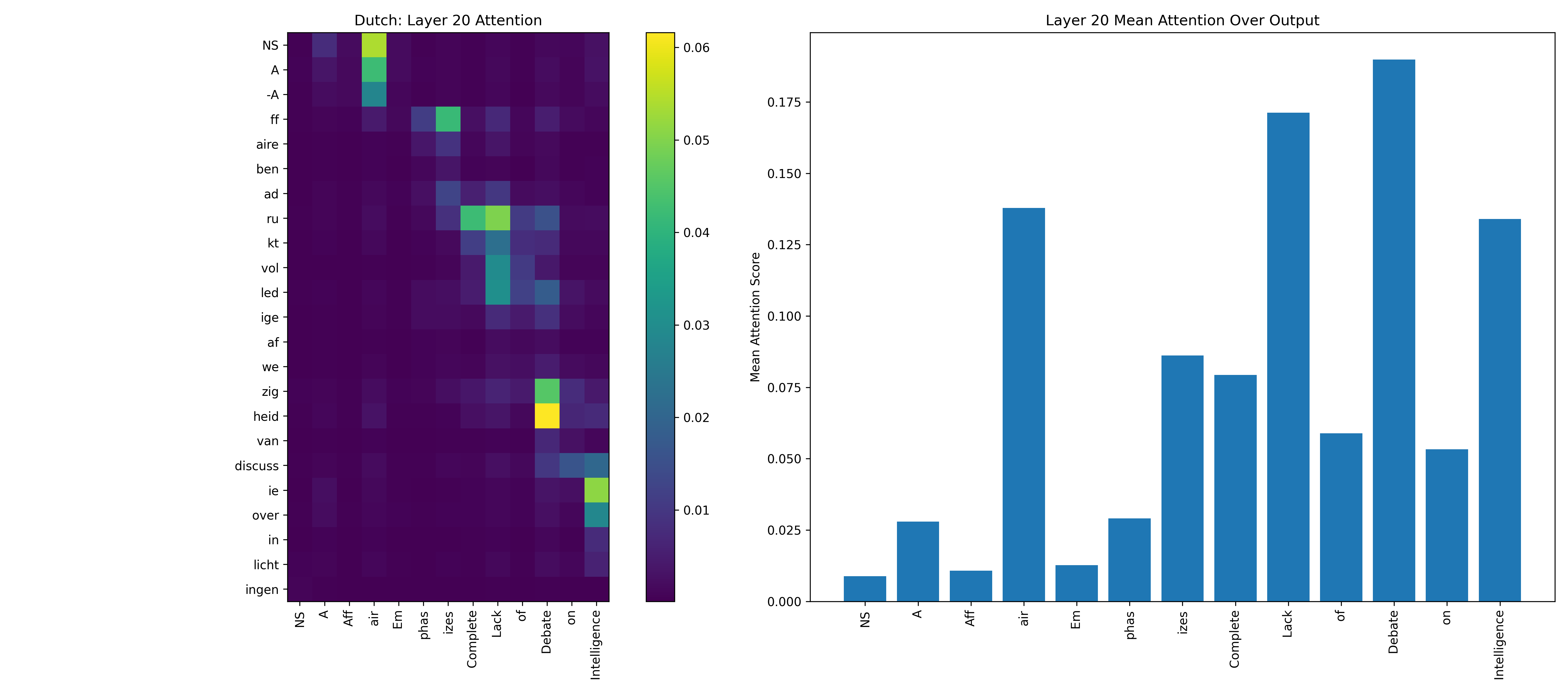}
			\caption{Dutch}
		\end{subfigure}

		\begin{subfigure}[b]{0.48\textwidth}
			\centering
			\includegraphics[width=\linewidth]{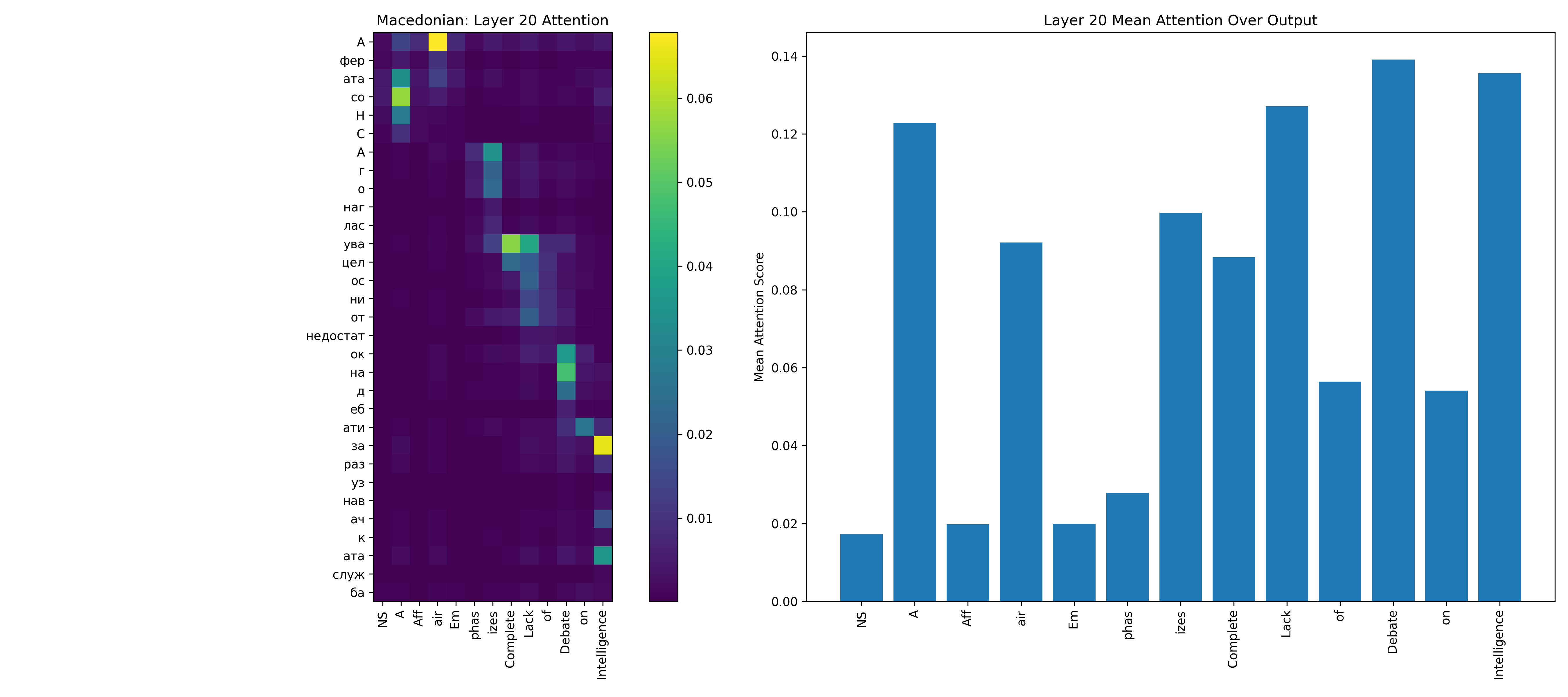}
			\caption{Macedonian}
		\end{subfigure}
		\hfill
		\begin{subfigure}[b]{0.48\textwidth}
			\centering
			\includegraphics[width=\linewidth]{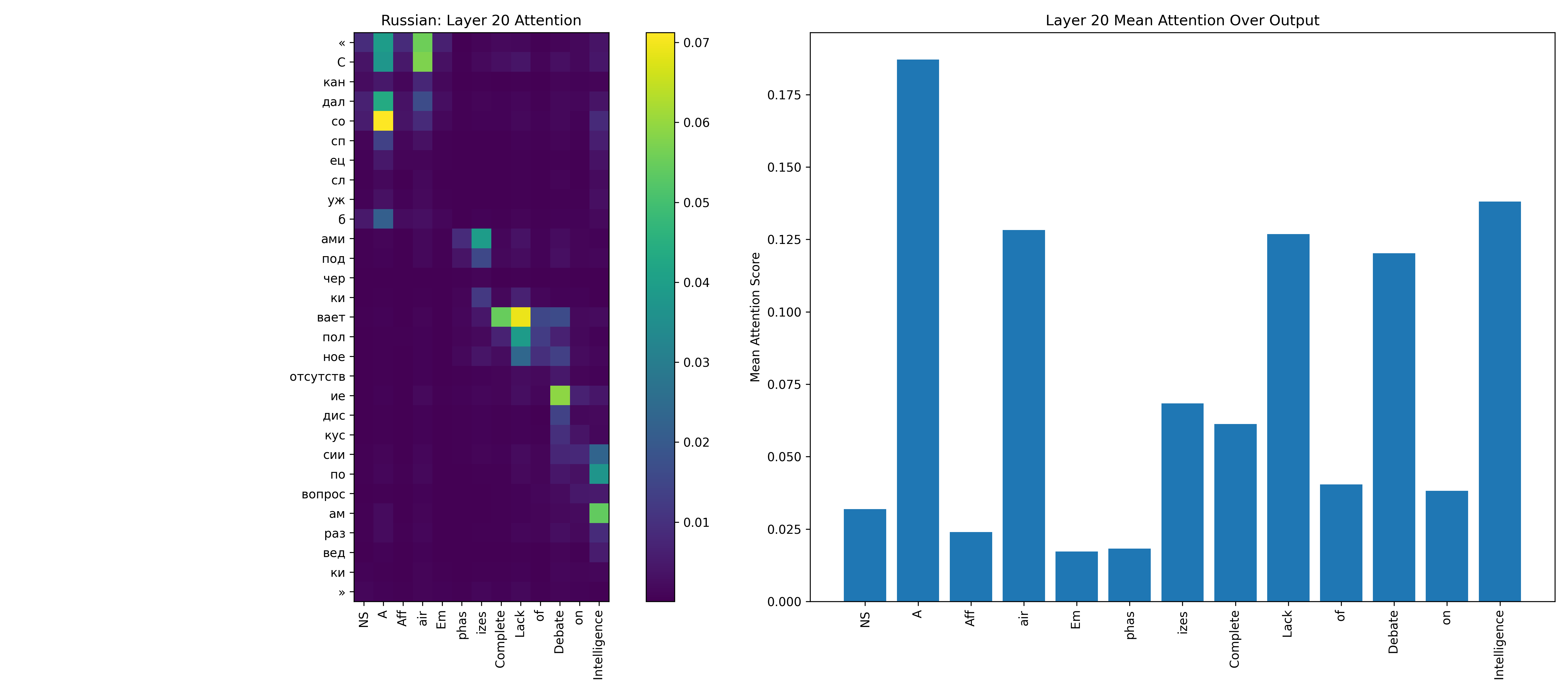}
			\caption{Russian}
		\end{subfigure}

		\begin{subfigure}[b]{0.48\textwidth}
			\centering
			\includegraphics[width=\linewidth]{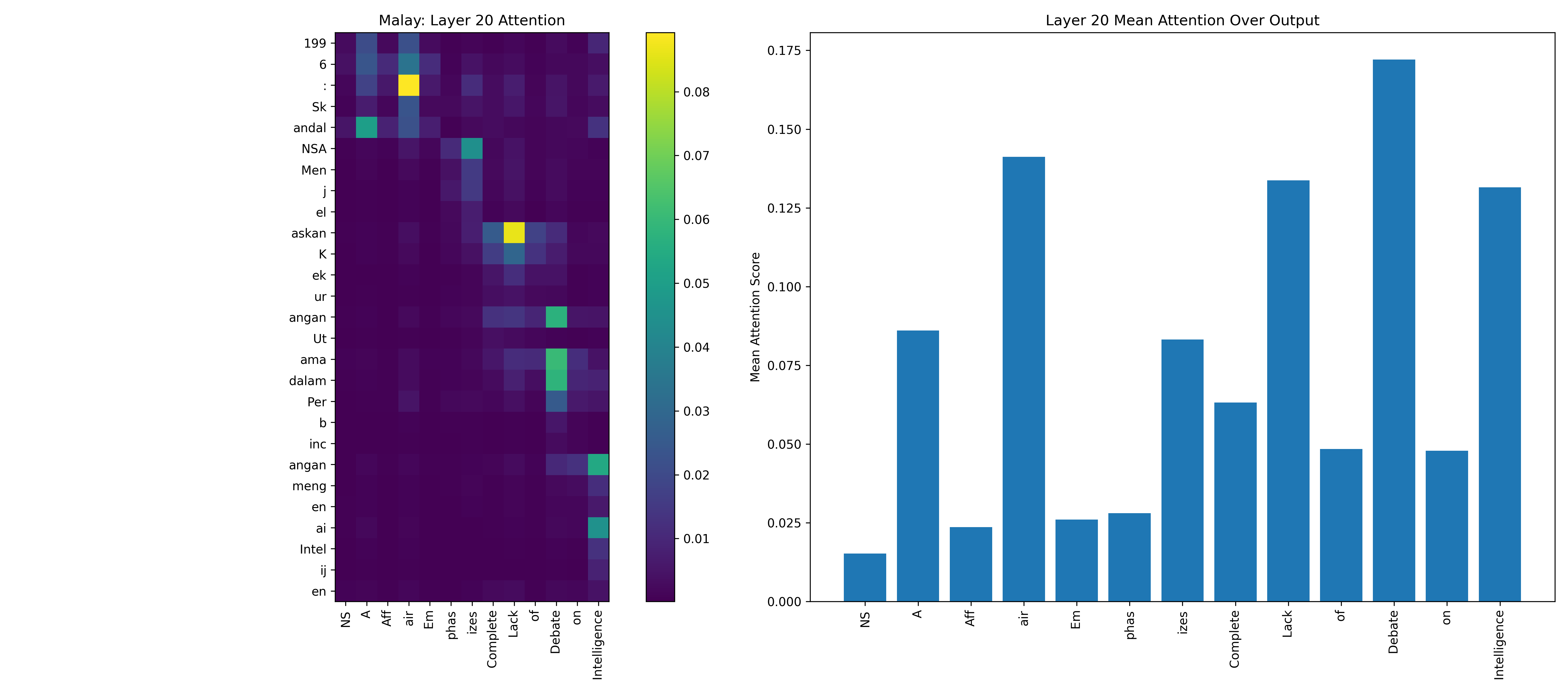}
			\caption{Malay}
		\end{subfigure}
		\hfill
		\begin{subfigure}[b]{0.48\textwidth}
			\centering
			\includegraphics[width=\linewidth]{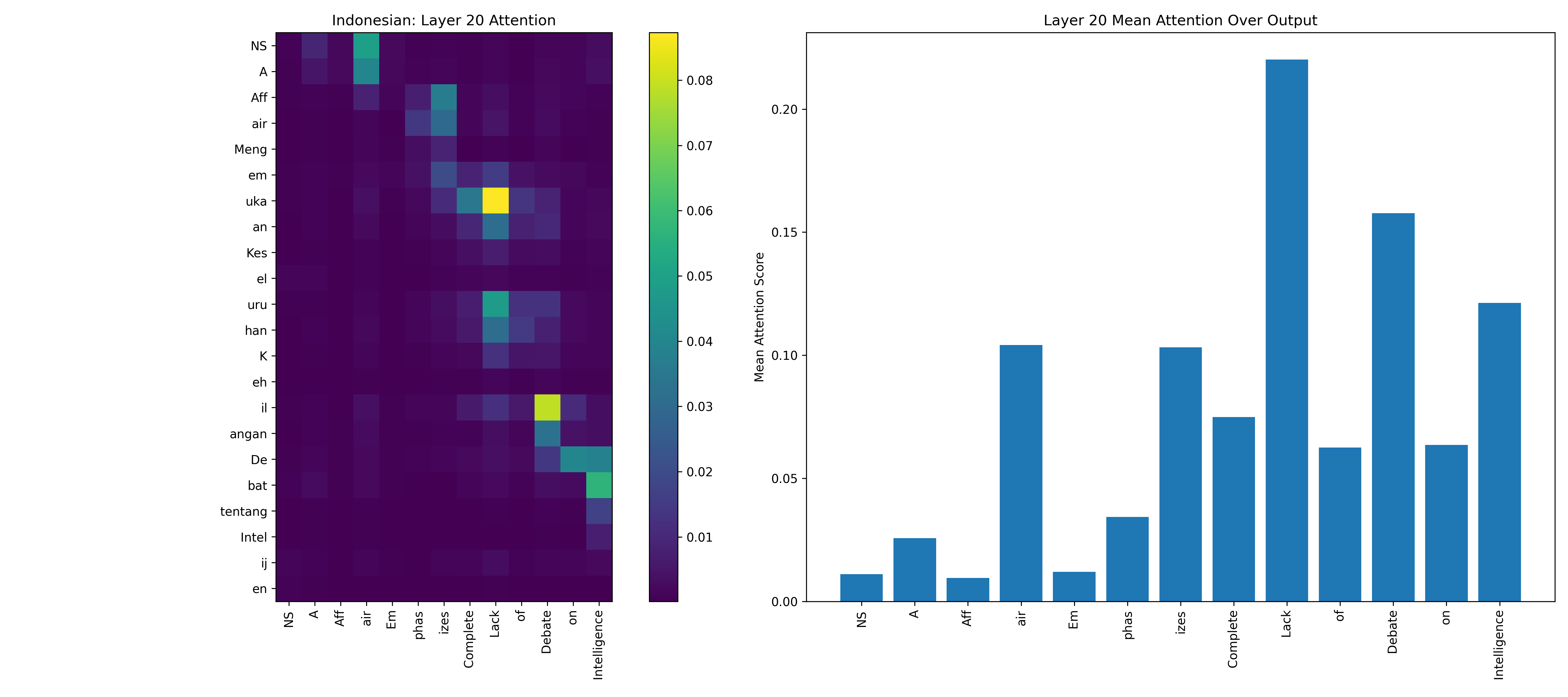}
			\caption{Indonesian}
		\end{subfigure}

		\begin{subfigure}[b]{0.48\textwidth}
			\centering
			\includegraphics[width=\linewidth]{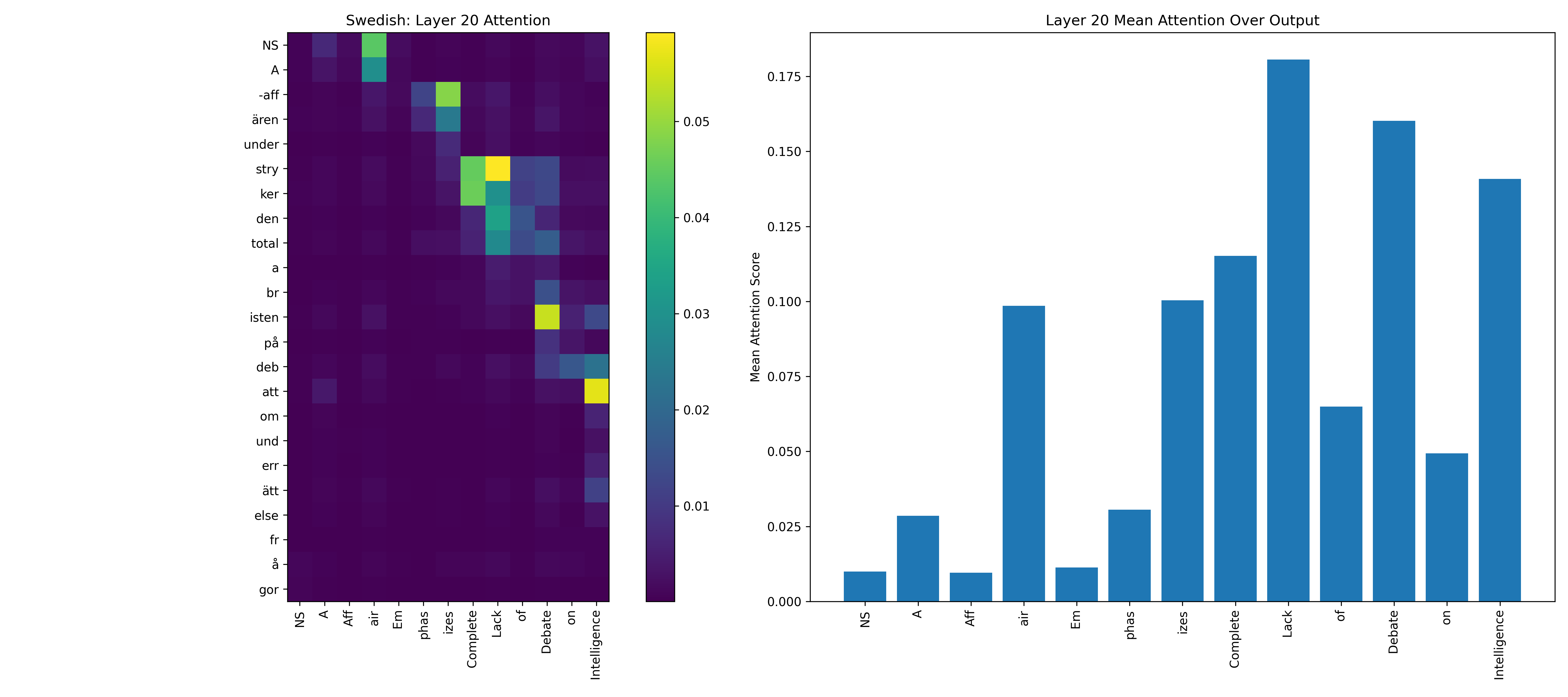}
			\caption{Swedish}
		\end{subfigure}
		\hfill
		\begin{subfigure}[b]{0.48\textwidth}
			\centering
			\includegraphics[width=\linewidth]{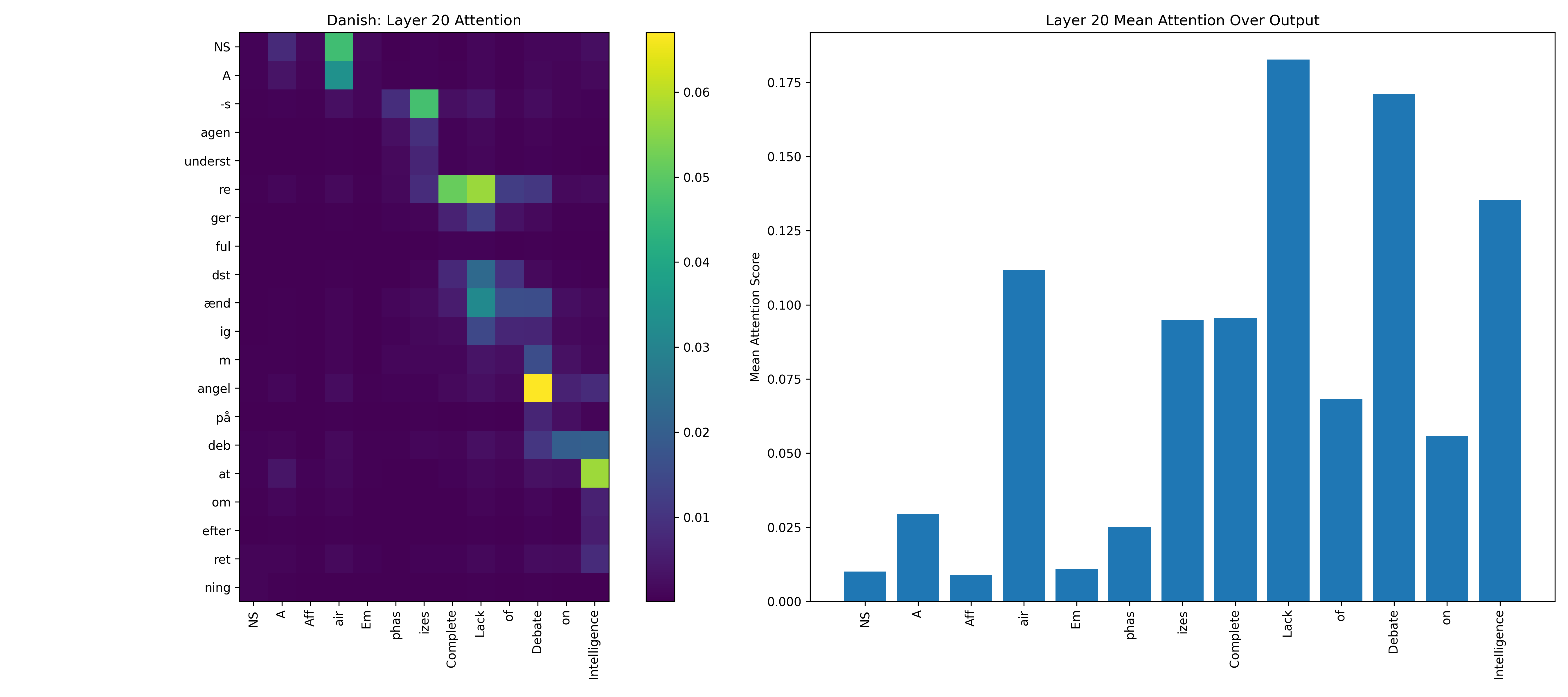}
			\caption{Danish}
		\end{subfigure}

		\caption{Attention maps for different languages in Llama-3-8B-Instruct}
		\label{fig:attention_llama}
	\end{figure}

	In this section, we illustrate attention behavior across selected language pairs by visualizing representative attention maps. Fig.~\ref{fig:attention_m2m} and~\ref{fig:attention_llama} show, for M2M-100 and Llama-3-8B-Instruct, examples from a specific layer and a specific attention head for multiple target languages. In each subfigure, the left panel displays the source to target attention matrix as a tokenwise heat map, and the right panel shows a one-dimensional attention profile obtained by averaging the matrix along the target dimension to yield a distribution.

	Each row presents two languages expected to be similar. Within a row, the heat maps and the aggregated profiles are closely aligned; across rows, clear differences appear, indicating language-specific patterns in attention. This motivates using the 2-Wasserstein distance to compare attention across translations of the same English sentence into different languages: although output lengths vary and the left heat maps have different sizes, averaging over the target dimension yields distributions on the same set of source tokens, which can be compared directly with $W_{2}$.

	\subsection{Translation Quality}\label{app:eval}

	We then use the following prompt template to evaluate translation quality via GPT-4o:

	\begin{lstlisting}[language=Python, frame=single, breaklines=true, basicstyle=\footnotesize\ttfamily, keywordstyle=\color{blue}\bfseries, stringstyle=\color{orange!85!black}, commentstyle=\color{green!50!black}, showspaces=false, showtabs=false, showstringspaces=false, columns=flexible, keepspaces=false, upquote=true, aboveskip=0pt, belowskip=0pt]
def eval_response_by_llm(original_text, translated_text, tgt_lang, model="gpt-4o"):
    """
    use the language model to get the evaluation
    """
    response = openai.chat.completions.create(
        model=model,
        messages=[
            {
                "role": "system",
                "content": (
                    "You are an expert translation evaluator. "
                    "Classify the quality of a translation based on the "
                    "following rules:\n\n"
                    "- Respond 'yes' if the translation is grammatically "
                    "readable and the meaning is correct, even if there are "
                    "small grammar or word choice issues.\n"
                    "- Respond 'almost' if the translation makes an attempt "
                    "but includes incorrect key words, semantic mistakes, or "
                    "seems like a direct word-for-word substitution.\n"
                    "- Respond 'no' only if the translation is completely "
                    "incoherent, grammatically broken beyond repair, or fails "
                    "to resemble a real sentence.\n\n"
                    "Only reply with one word: yes, almost, or no."
                ),
            },
            {
                "role": "user",
                "content": (
                    f"Original English text: {original_text}\n"
                    f"Translated text in {tgt_lang}: {translated_text}\n"
                    f"What is your evaluation? Respond with one word only: "
                    f"yes, almost, or no."
                ),
            },
        ],
        max_tokens=10,
    )

    answer = response.choices[0].message.content.strip().lower()
    return answer
	\end{lstlisting}

	Based on the model's judgment (yes, almost, or no), we assign a numerical score to each translation: 1 for yes (accurate and fluent translations), 0.5 for almost (partially correct or semantically flawed), 0 for no (incoherent or entirely incorrect outputs).

To construct a reliable evaluation subset, we first rank all sentences within each language by their model-assigned quality scores. For M2M, which is explicitly trained on the full set of 100 languages for translation, we select the top 2000 translation sentences. In contrast, Llama-3-8B-Instruct, a general LLM rather than a translation-specialized model, often fails to strictly follow translation instructions and may not translate certain languages adequately. To mitigate these issues, we apply a stricter filtering strategy and select only the top 500 highest-scoring sentences.

For cross-language comparison, we compute the average score for each language using these selected subsets. We use a threshold of 0.2 for M2M, which yields 81 languages with consistently high translation quality. Given Llama’s substantially lower and less stable translation performance, we adopt a more conservative threshold of 0.6, resulting in 55 languages retained for visualization and further evaluation.

\begin{figure}[!htbp]
		\centering

		\begin{subfigure}[b]{0.48\textwidth}
			\centering
			\includegraphics[width=\linewidth]{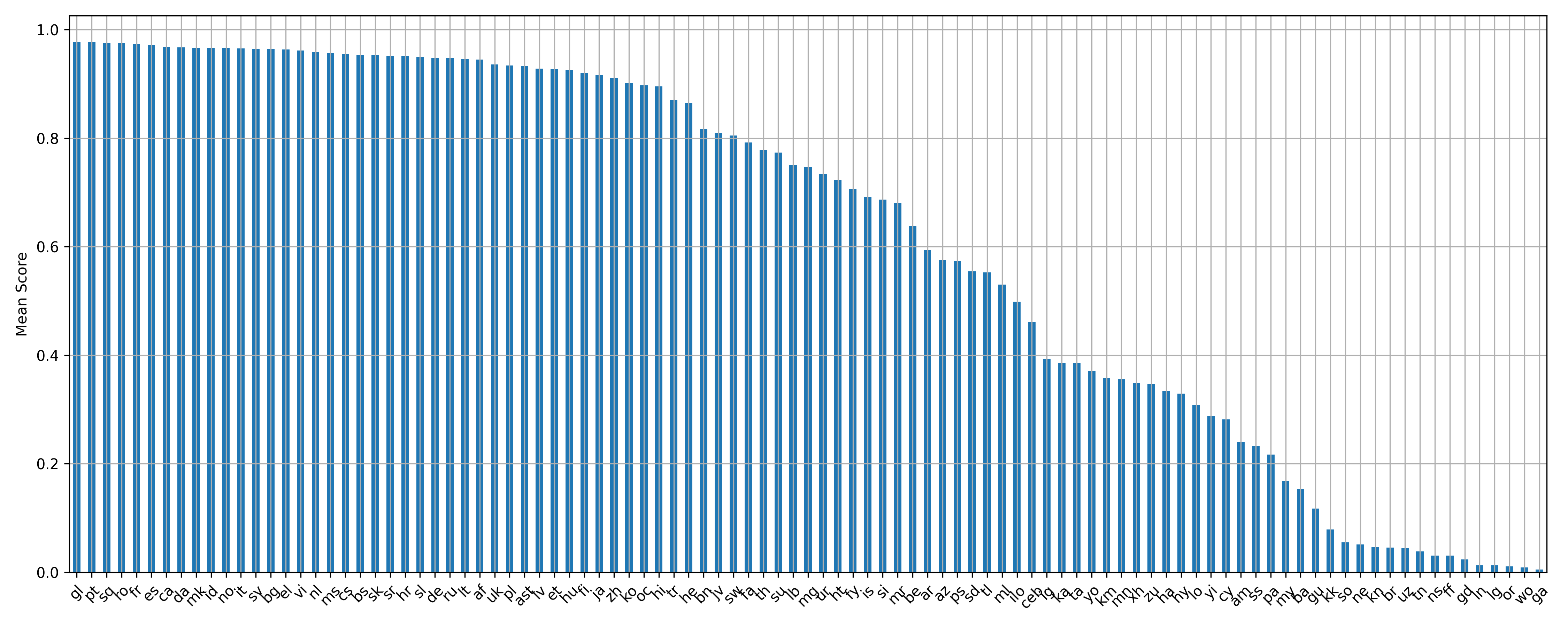}
			\caption{M2M-100}
		\end{subfigure}
		\hfill
		\begin{subfigure}[b]{0.48\textwidth}
			\centering
			\includegraphics[width=\linewidth]{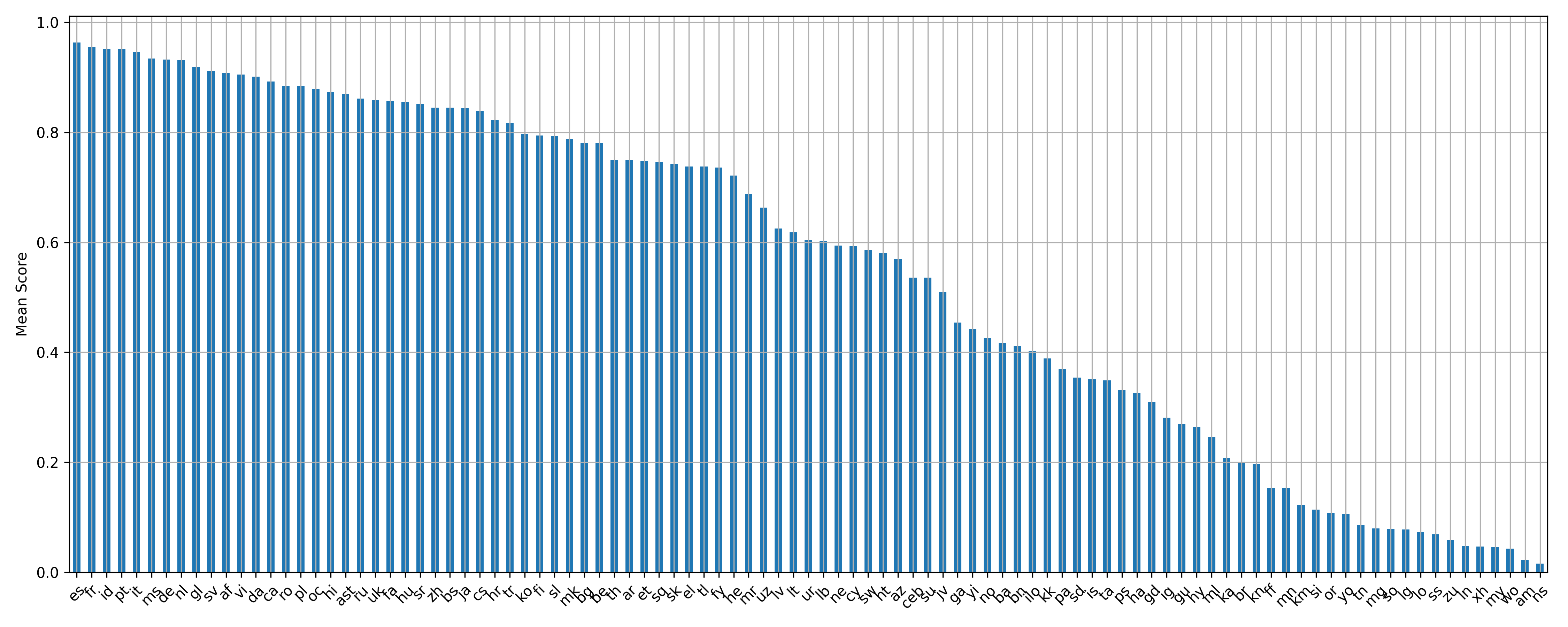}
			\caption{Llama-3-8B-Instruct}
		\end{subfigure}

		\caption{Translation quality scores for selected languages on M2M-100 and Llama-3-8B-Instruct}
		\label{fig:translation_scores}
	\end{figure}

	We plot all languages alongside their average evaluation scores in descending order in Fig.~\ref{fig:translation_scores}.

%% file: sections/appendix_control_stat.tex
\subsection{Language Groupings for Controlled Word Order Analysis}
\label{app:language_groupings}

To isolate the effect of word order typology from potential confounds arising from genetic relatedness or areal contact, we excluded language pairs that belong to the same language family or share substantial historical influence. Table~\ref{tab:language_groupings_full} provides the complete list of language groupings for each focal language.

For \textbf{Japanese} (SOV), we excluded Korean and Chinese. Korean shares extensive areal features and similar grammatical structure with Japanese despite being considered a language isolate by some scholars. Chinese has had strong historical contact and substantial lexical influence on Japanese through centuries of cultural exchange.

For \textbf{Xhosa} (SVO), we excluded seven languages: Zulu and Swati are closely related Nguni languages within the same branch of the Bantu family; Swahili, Yoruba, and Igbo belong to the broader Niger-Congo family; Afrikaans and Dutch were excluded due to prolonged areal contact in South Africa during the colonial and post-colonial periods.

For \textbf{Vietnamese} (SVO), we excluded five languages: Chinese has had massive lexical and structural influence from centuries of contact and shared cultural sphere; Khmer belongs to the same Austroasiatic family; Lao and Thai share extensive typological features as members of the Mainland Southeast Asian (MSEA) Sprachbund, including tonality and analytic morphology; French was excluded due to colonial influence during the French Indochina period.

\begin{longtable}{@{} >{\raggedright\arraybackslash}p{1.8cm} >{\raggedright\arraybackslash}p{2.2cm} >{\raggedright\arraybackslash}p{8cm} @{}}
\caption{Complete language groupings for controlled word order comparison. For each focal language, we list the excluded languages (due to genetic relatedness or areal contact), languages with the same dominant word order, and languages with different word orders. Numbers in parentheses indicate group sizes.} \label{tab:language_groupings_full} \\

\toprule
\textbf{Focal Language} & \textbf{Category} & \textbf{Languages} \\
\midrule
\endfirsthead

\multicolumn{3}{c}%
{{\bfseries \tablename\ \thetable{} -- continued from previous page}} \\
\toprule
\textbf{Focal Language} & \textbf{Category} & \textbf{Languages} \\
\midrule
\endhead

\midrule
\multicolumn{3}{r}{{Continued on next page}} \\
\bottomrule
\endfoot

\bottomrule
\endlastfoot

\multirow{6}{*}{\parbox{1.8cm}{\textbf{Japanese}\\(SOV)}} 
& Excluded (2) & Korean, Chinese \\[0.5ex]
& Same WO (16) & Amharic, Azerbaijani, Bengali, Farsi, Georgian, Hindi, Malayalam, Marathi, Mongolian, Panjabi, Pashto, Sindhi, Sinhala, Tamil, Turkish, Urdu \\[0.5ex]
& Different WO (62) & Afrikaans, Arabic, Asturian, Belarusian, Bosnian, Bulgarian, Catalan, Cebuano, Croatian, Czech, Danish, Dutch, Estonian, Finnish, French, Galician, German, Greek, Haitian Creole, Hausa, Hebrew, Hungarian, Icelandic, Igbo, Iloko, Indonesian, Italian, Javanese, Khmer, Lao, Latvian, Lithuanian, Luxembourgish, Macedonian, Malagasy, Malay, Norwegian, Occitan, Polish, Portuguese, Romanian, Russian, Serbian, Slovak, Slovenian, Spanish, Sundanese, Swahili, Swati, Swedish, Tagalog, Thai, Ukrainian, Vietnamese, Welsh, Western Frisian, Xhosa, Yiddish, Yoruba, Zulu, Albanian, Armenian \\
\midrule

\multirow{6}{*}{\parbox{1.8cm}{\textbf{Xhosa}\\(SVO)}} 
& Excluded (7) & Afrikaans, Dutch, Igbo, Swahili, Swati, Yoruba, Zulu \\[0.5ex]
& Same WO (48) & Asturian, Belarusian, Bosnian, Bulgarian, Catalan, Chinese, Croatian, Czech, Danish, Estonian, Finnish, French, Galician, German, Greek, Haitian Creole, Hausa, Hungarian, Icelandic, Indonesian, Italian, Javanese, Khmer, Lao, Latvian, Lithuanian, Luxembourgish, Macedonian, Malay, Norwegian, Occitan, Polish, Portuguese, Romanian, Russian, Serbian, Slovak, Slovenian, Spanish, Sundanese, Swedish, Thai, Ukrainian, Vietnamese, Western Frisian, Yiddish, Albanian, Armenian \\[0.5ex]
& Different WO (25) & Amharic, Arabic, Azerbaijani, Bengali, Cebuano, Farsi, Georgian, Hebrew, Hindi, Iloko, Japanese, Korean, Malagasy, Malayalam, Marathi, Mongolian, Panjabi, Pashto, Sindhi, Sinhala, Tagalog, Tamil, Turkish, Urdu, Welsh \\
\midrule

\multirow{6}{*}{\parbox{1.8cm}{\textbf{Vietnamese}\\(SVO)}} 
& Excluded (5) & Chinese, French, Khmer, Lao, Thai \\[0.5ex]
& Same WO (50) & Afrikaans, Asturian, Belarusian, Bosnian, Bulgarian, Catalan, Croatian, Czech, Danish, Dutch, Estonian, Finnish, Galician, German, Greek, Haitian Creole, Hausa, Hungarian, Icelandic, Igbo, Indonesian, Italian, Javanese, Latvian, Lithuanian, Luxembourgish, Macedonian, Malay, Norwegian, Occitan, Polish, Portuguese, Romanian, Russian, Serbian, Slovak, Slovenian, Spanish, Sundanese, Swahili, Swati, Swedish, Ukrainian, Western Frisian, Xhosa, Yiddish, Yoruba, Zulu, Albanian, Armenian \\[0.5ex]
& Different WO (25) & Amharic, Arabic, Azerbaijani, Bengali, Cebuano, Farsi, Georgian, Hebrew, Hindi, Iloko, Japanese, Korean, Malagasy, Malayalam, Marathi, Mongolian, Panjabi, Pashto, Sindhi, Sinhala, Tagalog, Tamil, Turkish, Urdu, Welsh \\

\end{longtable}

%% file: sections/appendix_m2m_100.tex
\section{Additional Results on the Full M2M-100 Language Set}\label{app:results_m2m_100}

In the main text, we restrict our analysis to 81 target languages for which the M2M-100 model produces consistently reliable translations under our quality threshold. This restriction is intended to ensure that the extracted attention patterns reflect stable cross-lingual alignment behavior rather than noise induced by severely degraded translations.

In this appendix, we extend the same ATD pipeline to the full set of 99 target languages supported by M2M-100 (excluding English, which is used as the fixed source language). This extended analysis allows us to examine how ATD behaves when including languages with lower translation quality, and to assess whether the large-scale structural patterns observed in the main text persist under weaker reliability conditions. All methodological choices, including attention extraction, optimal transport computation, and NJ reconstruction, are identical to those used in the main text.

\begin{figure}[H]
	\centering
	\includegraphics[width=\linewidth]{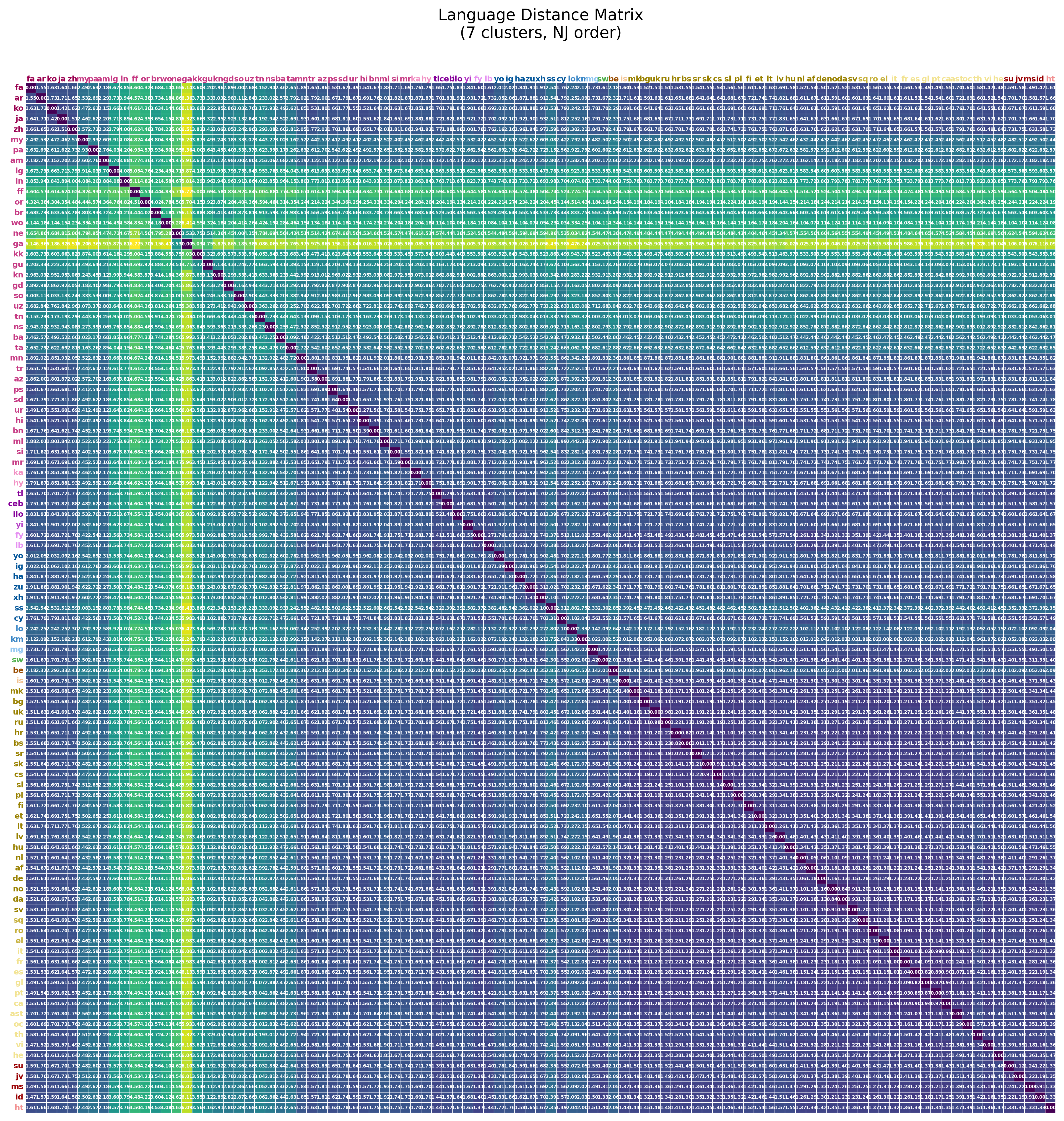}
	\caption{\textbf{ATD heat-map for M2M-100.}
		Pair-wise ATD values between the 99 target languages are shown as a
		heat-map.}
	\label{fig:m2m100_heatmap_atd}
\end{figure}

Fig.~\ref{fig:m2m100_heatmap_atd} presents the ATD distance matrix computed over the full 99-language set. Despite the inclusion of languages whose translation quality scores fall below the threshold used in the main text, the matrix continues to exhibit pronounced block structure at multiple scales.

Major macro-regions visible in the 81-language setting, such as Indo-European, East Asian, Austronesian, and Niger–Congo groupings, remain clearly identifiable. This suggests that ATD captures robust large-scale organization in model-internal attention patterns that is not entirely dependent on near-perfect translation quality.

At the same time, distances involving lower-quality languages tend to be inflated and more heterogeneous, resulting in less sharply defined sub-blocks. We interpret this effect as reflecting increased uncertainty in attention alignment rather than a breakdown of the underlying representational structure.

\begin{figure}[H]
	\centering

		\includegraphics[width=\textwidth]{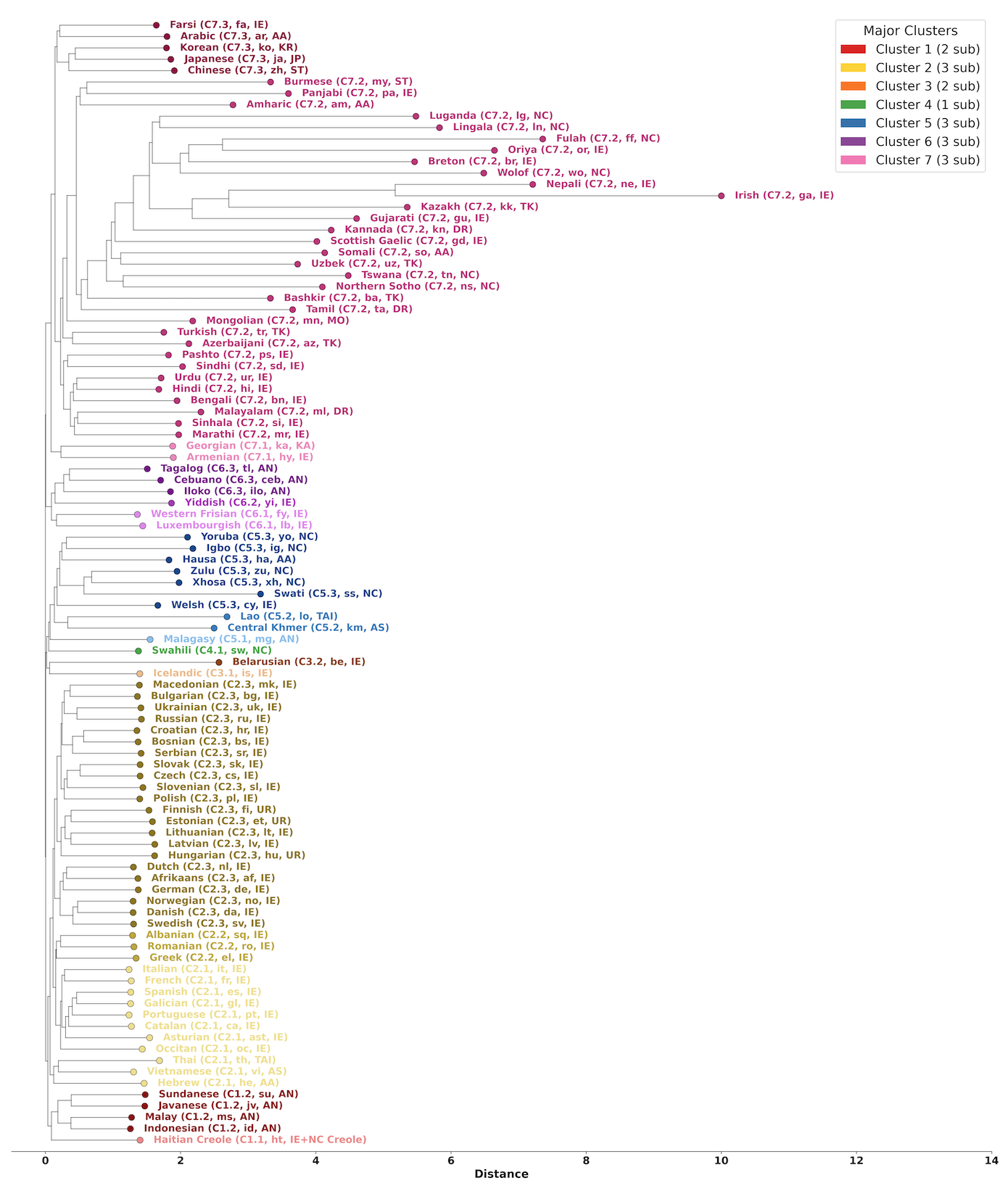}
		\caption{Neighbor-Joining tree inferred from ATD distances in the M2M-100 model over the full 99-language set.}
		\label{fig:m2m100_nj_tree_all}
\end{figure}

Fig.~\ref{fig:m2m100_nj_tree_all} shows the NJ tree reconstructed from the 99-language ATD matrix, with cluster assignments obtained using the same depth-based cutting procedure as in the main text.

The overall topology remains broadly consistent with the main-text results. Large genealogical families continue to form coherent regions of the tree, and languages added in this extended setting typically attach to the periphery of their expected macro-clusters rather than forming entirely spurious groupings.

However, several newly included languages appear as elongated or weakly attached branches. Such behavior is expected in a distance-based reconstruction when pairwise distances are noisier or less uniformly reliable. Importantly, these branches do not destabilize the core structure of the tree, indicating that the NJ reconstruction remains globally stable even under expanded and noisier input.

\begin{figure}[H]
		\centering
		\includegraphics[width=0.9\textwidth]{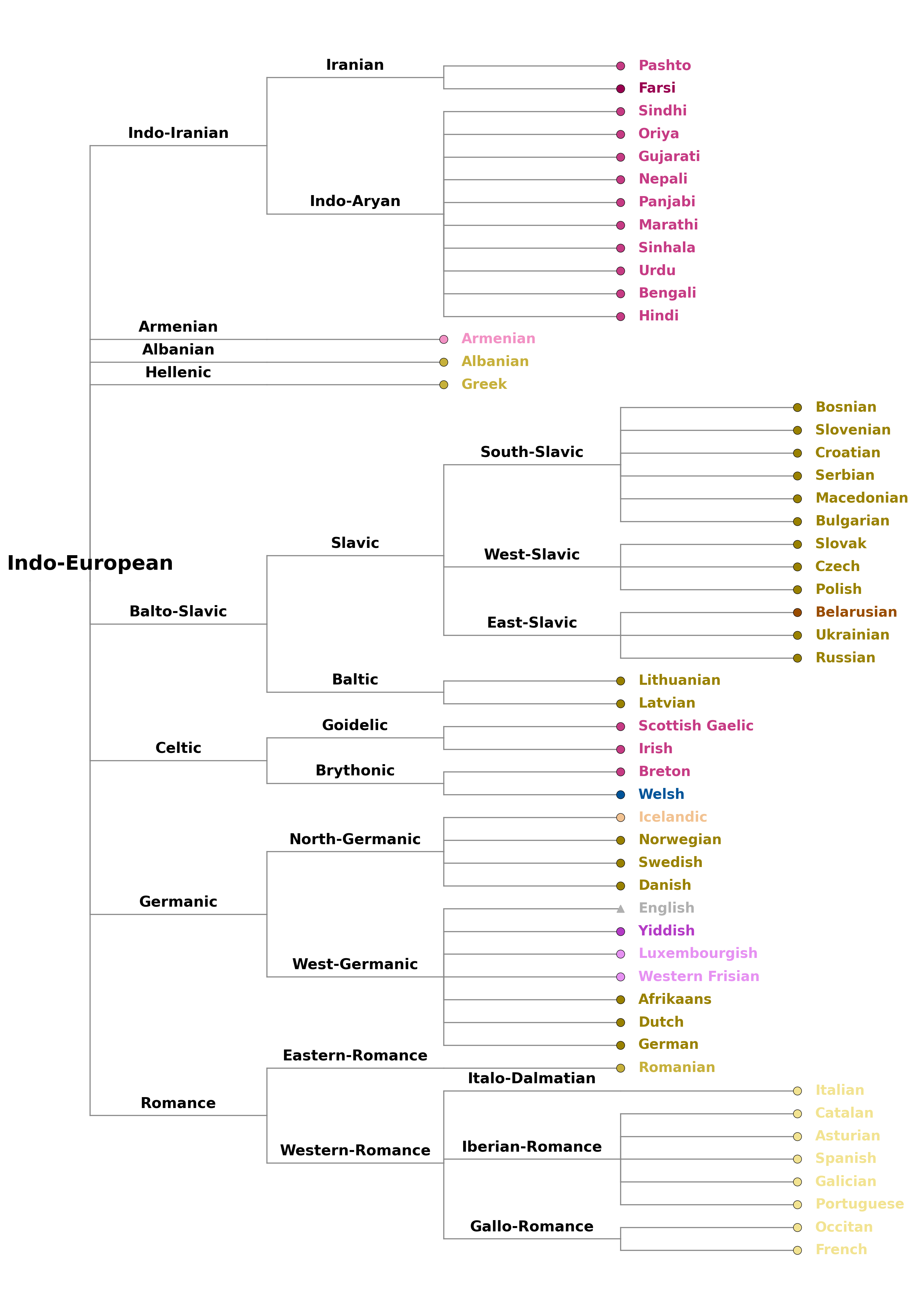}
		\caption{Established Indo-European language tree annotated with ATD-based cluster assignments over the full 99-language set.}
		\label{fig:m2m100_nj_tree_indo}
\end{figure}

\begin{figure}[H]
	\centering
	\includegraphics[width=\linewidth]{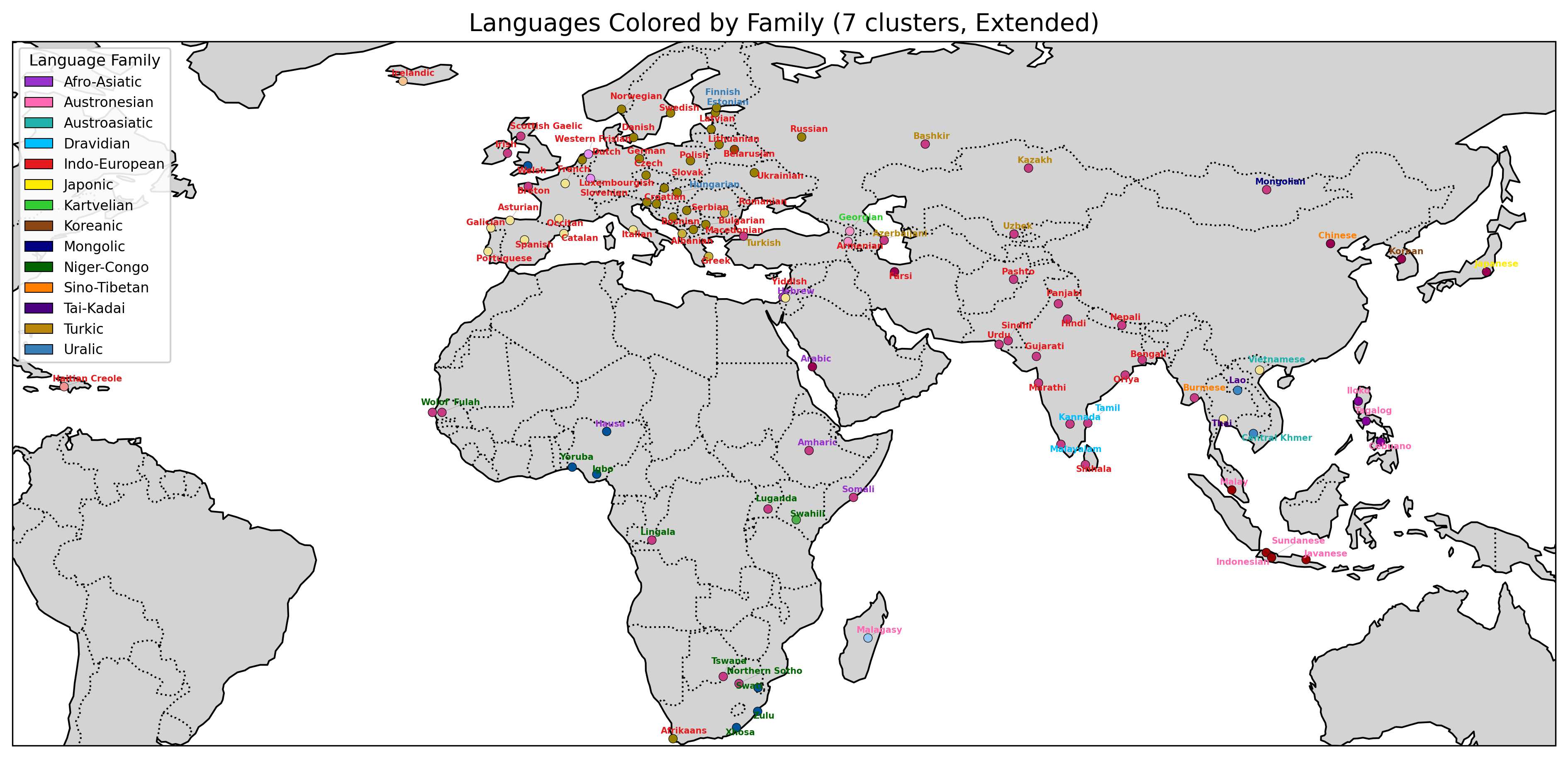}
	\caption{\textbf{Geographic distribution of ATD-derived language clusters.}
	Languages are shown at their approximate geographic locations, with markers colored according to ATD-based cluster assignments obtained from the M2M-100 model, using the same cluster definitions as in Fig.~\ref{fig:m2m100_nj_tree_all}. Text labels indicate established linguistic family classifications. This visualization provides an external reference for assessing the alignment between model-derived language structure and known geographic and genealogical patterns.}
	\label{fig:m2m100_geo_map}
\end{figure}

Fig.~\ref{fig:m2m100_geo_map} maps the ATD-derived cluster assignments for the 99-language set onto geographic space. As in the main text, substantial alignment between geographic proximity and model-derived clusters is observed at a coarse level, particularly within Europe, East Asia, and parts of Africa.

For languages with lower translation quality, geographic placement should be interpreted with additional caution. In such cases, ATD reflects the model's learned representation of these languages under limited or uneven training data, rather than a precise estimate of functional similarity under translation.

Overall, the 99-language analysis shows that the large-scale structure recovered by ATD remains robust when extending beyond high-reliability languages. Decreased translation quality mainly leads to increased distance variance and weaker attachment in the NJ tree, rather than arbitrary reassignment or structural collapse. While results for lower-quality languages should therefore be interpreted with caution, they remain informative as reflections of how pretrained multilingual models internally organize and relate languages under realistic and uneven coverage conditions. These findings complement the main-text analysis by clarifying both the stability and the interpretive limits of attention-based distance measures.

%% file: sections/appendix_llama_new.tex
\section{Additional Results for Llama-3-8B-Instruct}\label{app:results_llama}

In this appendix, we extend the ATD analysis to Llama-3-8B-Instruct, a decoder-only large language model that is not explicitly trained for multilingual translation. Unlike M2M-100, Llama-3 exhibits substantial variation in translation reliability across languages, reflecting both differences in pretraining coverage and the absence of dedicated translation supervision.

\begin{figure}[H]
	\centering
	\includegraphics[width=\linewidth]{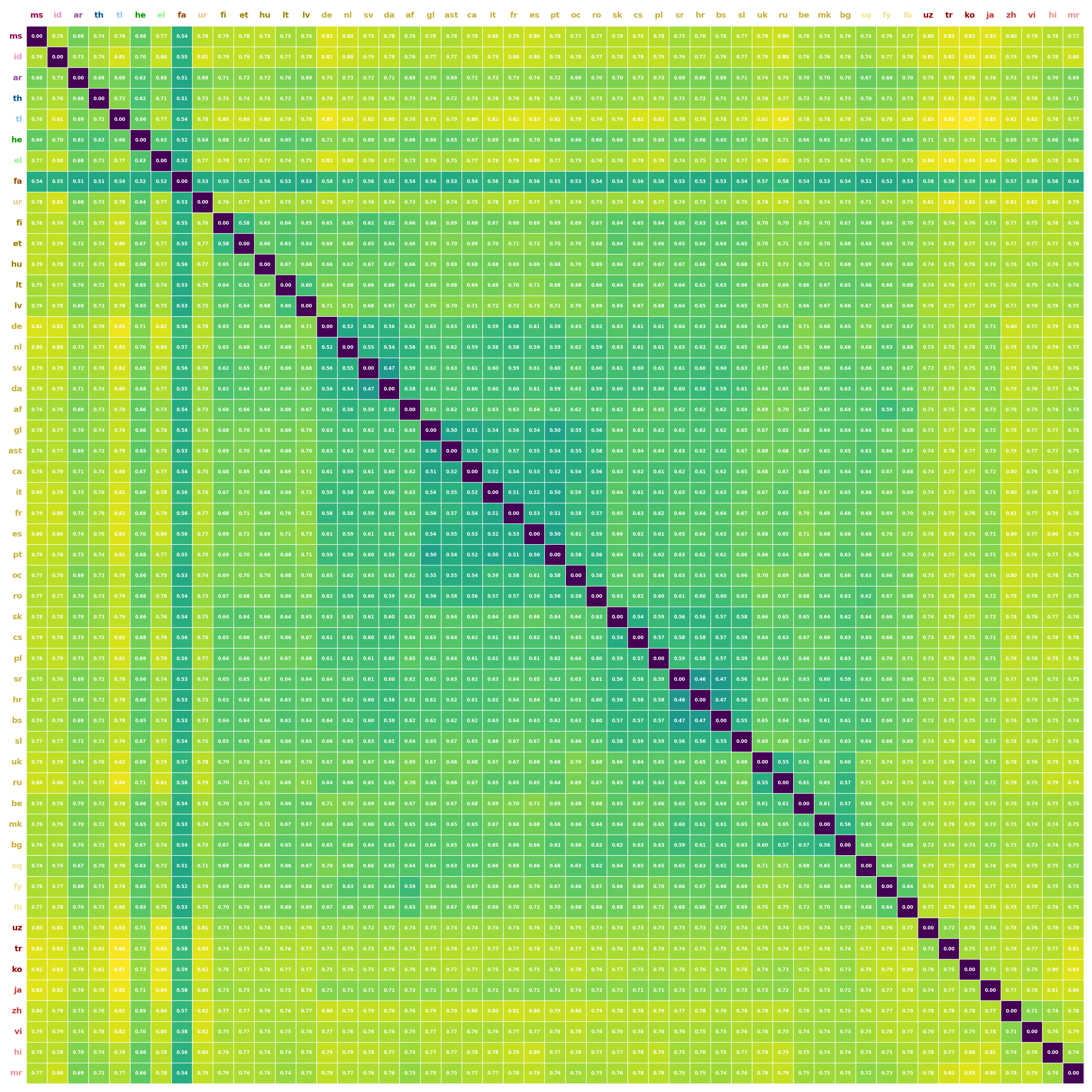}
	\caption{\textbf{ATD heat-map for Llama-3-8B-Instruct.}
		Pair-wise ATD values between the 51 target languages are shown as a
		heat-map.}
	\label{fig:llama_heatmap_atd}
\end{figure}

To ensure interpretability, we restrict our analysis to 51 target languages for which Llama-3 produces translations of minimally acceptable quality under our evaluation threshold. While this subset is smaller than that used for M2M-100, it still spans a diverse set of language families and typological profiles, allowing us to assess whether ATD recovers meaningful cross-linguistic structure beyond specialized translation models. All figures in this appendix follow the same visualization conventions as in the main text, including ATD distance matrices, NJ trees, Indo-European projections, and geographic maps.

Applying ATD to Llama-3 yields a dense, symmetric pairwise distance matrix over the selected 51 languages (Fig. \ref{fig:llama_heatmap_atd}). Compared to M2M-100, the overall contrast between within-cluster and between-cluster distances is visibly reduced, indicating a more diffuse representational geometry.

Nevertheless, the matrix still exhibits non-random block structure, with languages from similar genealogical or typological backgrounds tending to occupy contiguous regions. This suggests that, despite weaker alignment signals, Llama-3's attention patterns encode systematic cross-linguistic organization rather than arbitrary noise.

\begin{figure}[H]
		\centering
		\includegraphics[width=\textwidth]{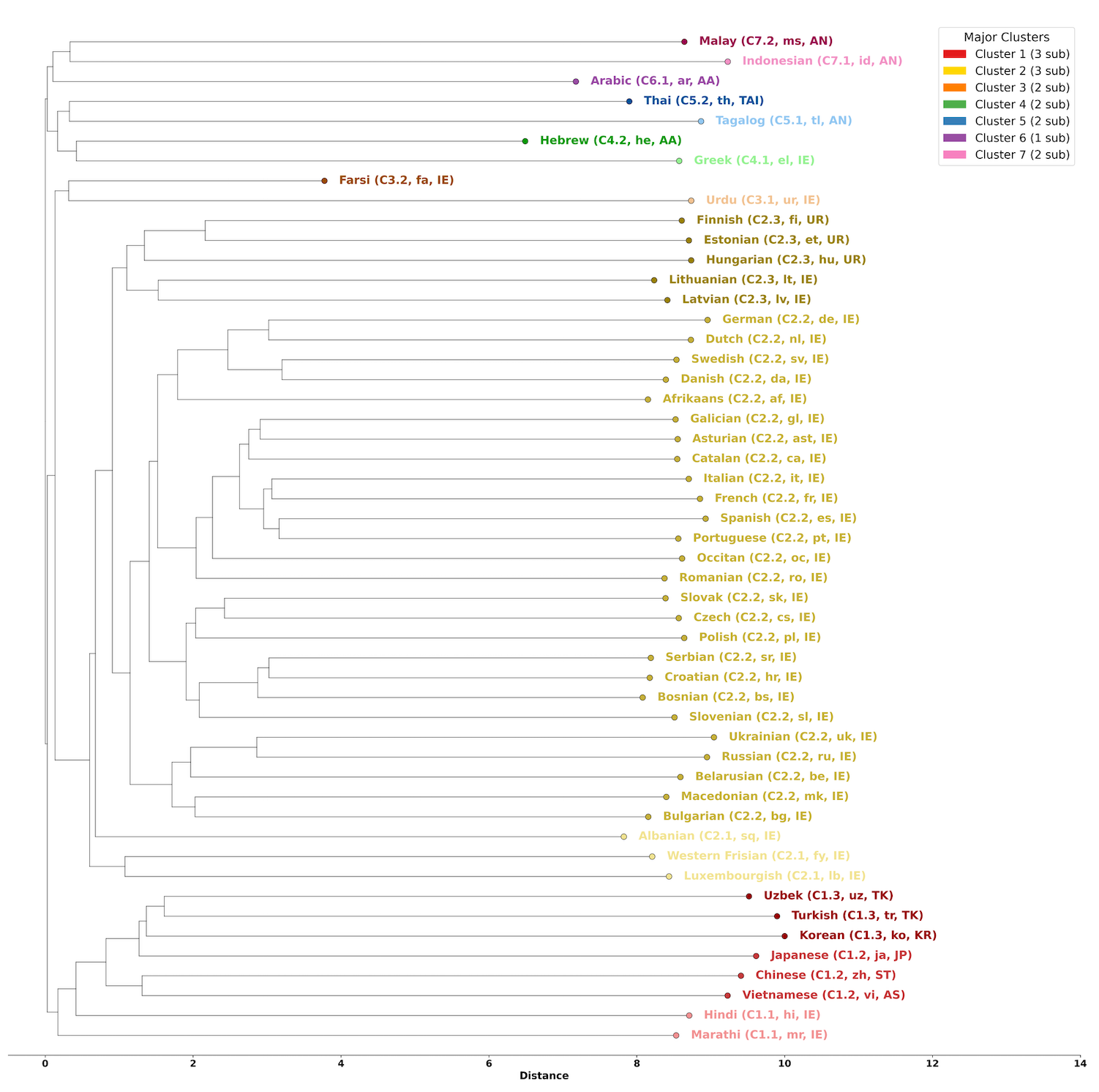}
		\caption{Neighbor-Joining tree inferred from ATD distances in the Llama-3-8B-Instruct model.}
		\label{fig:llama_nj_tree_all}
\end{figure}

We apply the same NJ procedure to the Llama-3 ATD matrix to obtain a hierarchical representation of language relationships (Fig. \ref{fig:llama_nj_tree_all}). As expected, the resulting NJ tree is less sharply resolved than that obtained from M2M-100, with longer branches and increased variance in attachment strength.

Despite this reduced resolution, several robust linguistic patterns persist. Major Indo-European subgroups remain partially coherent, and languages from distinct families tend to separate at higher levels of the tree. The persistence of these structures indicates that ATD captures graded similarity signals even in models not optimized for translation.

Quantitatively, the cophenetic correlation between ATD distances and NJ patristic distances remains high relative to chance, though lower than that observed for M2M-100, consistent with noisier alignment.

\begin{figure}[H]
		\centering
		\includegraphics[width=0.9\textwidth]{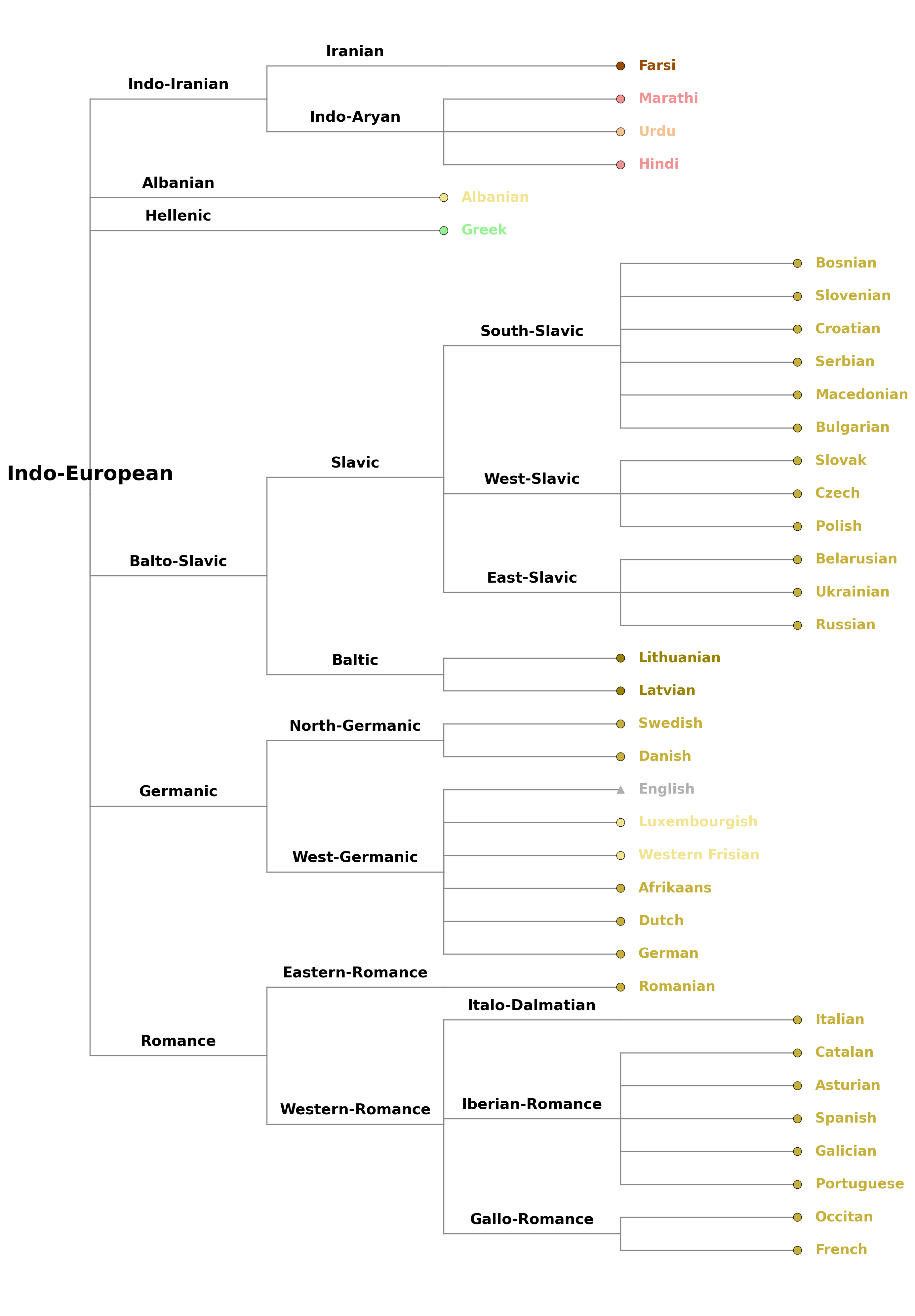}
		\caption{Established Indo-European language tree annotated with ATD-based cluster assignments.}
		\label{fig:llama_nj_tree_indo}
\end{figure}

Projecting ATD-based cluster assignments onto the Indo-European reference tree (Fig. \ref{fig:llama_nj_tree_indo}) reveals both agreement and divergence. Broad divisions among Romance, Germanic, Slavic, and Indo-Iranian languages are still visible, but internal boundaries are less sharply defined.

This behavior is consistent with the nature of Llama-3's training: without explicit translation supervision, its attention patterns reflect weaker and more indirect cross-linguistic alignment, shaped primarily by shared script, lexical overlap, and distributional exposure rather than systematic sentence-level correspondence.

\begin{figure}[H]
	\centering
	\includegraphics[width=\linewidth]{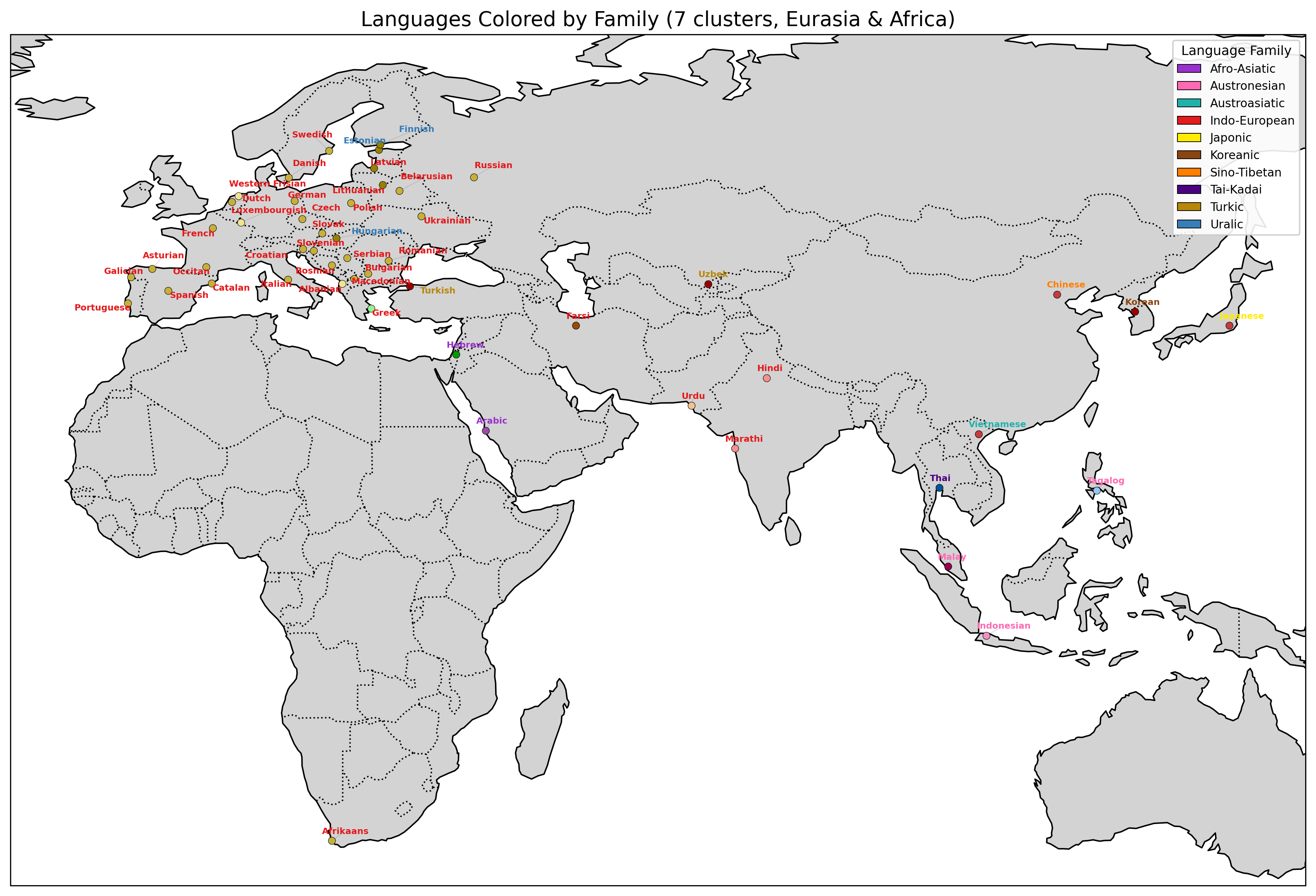}
	\caption{\textbf{Geographic distribution of ATD-derived language clusters.}
	Languages are shown at their approximate geographic locations, with markers colored according to ATD-based cluster assignments obtained from the Llama-3 model, using the same cluster definitions as in Fig.~\ref{fig:llama_nj_tree_all}. Text labels indicate established linguistic family classifications. This visualization provides an external reference for assessing the alignment between model-derived language structure and known geographic and genealogical patterns.}
	\label{fig:llama_geo_map}
\end{figure}

The geographic visualization of ATD-derived clusters for Llama-3 (Fig. \ref{fig:llama_geo_map}) shows a looser but still interpretable alignment between model-derived groupings and spatial distribution. Languages spoken in geographically proximal regions tend to appear closer in the ATD space, though with greater dispersion than in the M2M-100 results.

Overall, the ATD analysis on Llama-3 shows that attention-based language distances remain informative even in a model that is neither explicitly trained for translation nor uniformly exposed to all languages. Compared to M2M-100, the resulting distance matrix and NJ structure are visibly more diffuse, reflecting weaker and noisier cross-lingual alignment. Nevertheless, non-trivial linguistic organization persists: major language families remain partially coherent, Indo-European structure is still discernible, and geographic proximity continues to correlate with ATD-derived similarity. These findings suggest that ATD captures robust aspects of model-internal cross-linguistic representation that generalize beyond translation-centric architectures, while also highlighting the strong influence of training objectives and data coverage on the clarity of emergent linguistic structure.

%% file: sections/appendix_transfer.tex
\section{ATD Regularization Settings and Additional Results}
\label{app:transfer_appendix}

\subsection{Loss Function and Training Setup}

This appendix details the ATD loss formulation and fine-tuning setup.
We fine-tune the M2M-100 (1.2B) model~\cite{fan2021beyond} on the OPUS-100 corpus~\cite{zhang-etal-2020-improving}, using up to 50k training and 2k test sentence pairs.
The total objective combines the standard cross-entropy translation loss with an ATD regularization term:
\[
	\mathcal{L}_{\text{total}} = \mathcal{L}_{\text{CE}} + \lambda \, \mathcal{L}_{\text{ATD}}, \quad
	\mathcal{L}_{\text{ATD}} =
	\frac{1}{N \cdot L}\sum_{i=1}^{N} \sum_{l=1}^{L}
	W_2^{(\sigma)}(\tilde{A}_{\text{target}}^{i,l}, \tilde{A}_{\text{ref}}^{i,l}).
\]
The coefficient $\lambda$ serves as a loss balancing ratio designed to align the relative magnitudes of the two objectives. Rather than using a fixed constant, we adaptively scale $\lambda$ such that the effective contribution of the regularization term $\lambda \mathcal{L}_{\text{ATD}}$ maintains a target ratio relative to the translation loss $\mathcal{L}_{\text{CE}}$. We evaluate this setup on three target-reference language pairs: Pashto--Farsi (ps--fa), Marathi--Hindi (mr--hi), and Icelandic--Norwegian (is--no). For the ps--fa pair, we set this ratio to 1:10. For the mr--hi and is--no pairs, we use 1:1.

Here, we use the debiased Sinkhorn divergence implemented in \texttt{GeomLoss}~\cite{feydy2019interpolating} for the Wasserstein distance, which computes an entropy-regularized optimal transport in a kernelized form. The Gaussian blur parameter $\sigma=0.05$ controls the spatial smoothness of the transport cost and implicitly stabilizes the optimization, analogous to the role of the entropy term in standard Sinkhorn formulations.
$\tilde{A}_{\text{target}}$ and $\tilde{A}_{\text{ref}}$ are mean-aggregated cross-attention maps for the target and reference languages obtained from the same English source. 
Reference attention maps are precomputed once and cached, while target attentions are dynamically generated during fine-tuning.
The sum in $\mathcal{L}_{\text{ATD}}$ is taken over all sentences and attention layers.

Fine-tuning is performed for 20k steps on a single NVIDIA A100 (80GB) GPU using the AdamW optimizer with a learning rate of $3\times10^{-5}$, batch size 32, and a warm-up ratio of 0.1.
Evaluation is conducted every 1k steps.
To ensure robustness, all experiments are repeated with three random seeds, and the figures report the mean and standard deviation across runs. We summarize the training configuration and evaluation metrics in Table~\ref{tab:transfer_config}.

\vspace{1em}
\begin{table}[h]
	\centering
	\caption{Training configuration and evaluation metrics. Arrows (↑/↓) indicate whether higher or lower values denote better performance.}\label{tab:transfer_config}
	\small
	\begin{tabular}{ll}
		\toprule
		\textbf{Training configuration}      & \textbf{Value / Description}                          \\
		\midrule
		Base model                           & M2M-100 (1.2B) multilingual model                     \\
		Dataset                              & Pashto–English pairs (OPUS-100)                       \\
		Training / Test size                 & 50k (\texttt{ps, is}), 27,007 (\texttt{mr}) / 2k pairs                                        \\
		Learning rate                        & $3\times10^{-5}$                                      \\
		Batch size                           & 32                                                    \\
		Optimizer                            & AdamW ($\beta_1=0.9$, $\beta_2=0.999$)                \\
		Warm-up ratio                        & 0.1                                                   \\
		Loss balancing ratio ($\mathcal{L}_{\text{CE}} : \lambda \mathcal{L}_{\text{ATD}}$) & 10:1 (\texttt{ps}), 1:1 (\texttt{mr, is})                                                   \\
		GeomLoss blur $\sigma$               & 0.05             \\
		Beam Search                    & 5                                                  \\
		\midrule
		\textbf{Evaluation metrics}          & \textbf{Definition (↑/↓)}                             \\
		BLEU~\cite{papineni2002bleu}         & Measures $n$-gram precision (↑)  \\
		chrF~\cite{popovic2015chrf}          & Character-level F$_1$ score (↑) \\
		TER~\cite{snover2006study}           & Translation edit rate (↓)   \\
		COMET~\cite{rei2020comet}            & Learned metric reflecting semantic adequacy (↑)       \\
		\bottomrule
	\end{tabular}
\end{table}

\subsection{Additional Results on Transfer Training}
\label{app:transfer_ablation}

In the main text, we demonstrate that ATD regularization toward a linguistically related language significantly improves translation performance. Here, we further investigate the impact of reference language selection by comparing the ``related" reference with linguistically distant ones, specifically Spanish (\texttt{es}) and Arabic (\texttt{ar}).

\begin{figure}[h]
    \centering
    \includegraphics[width=\linewidth]{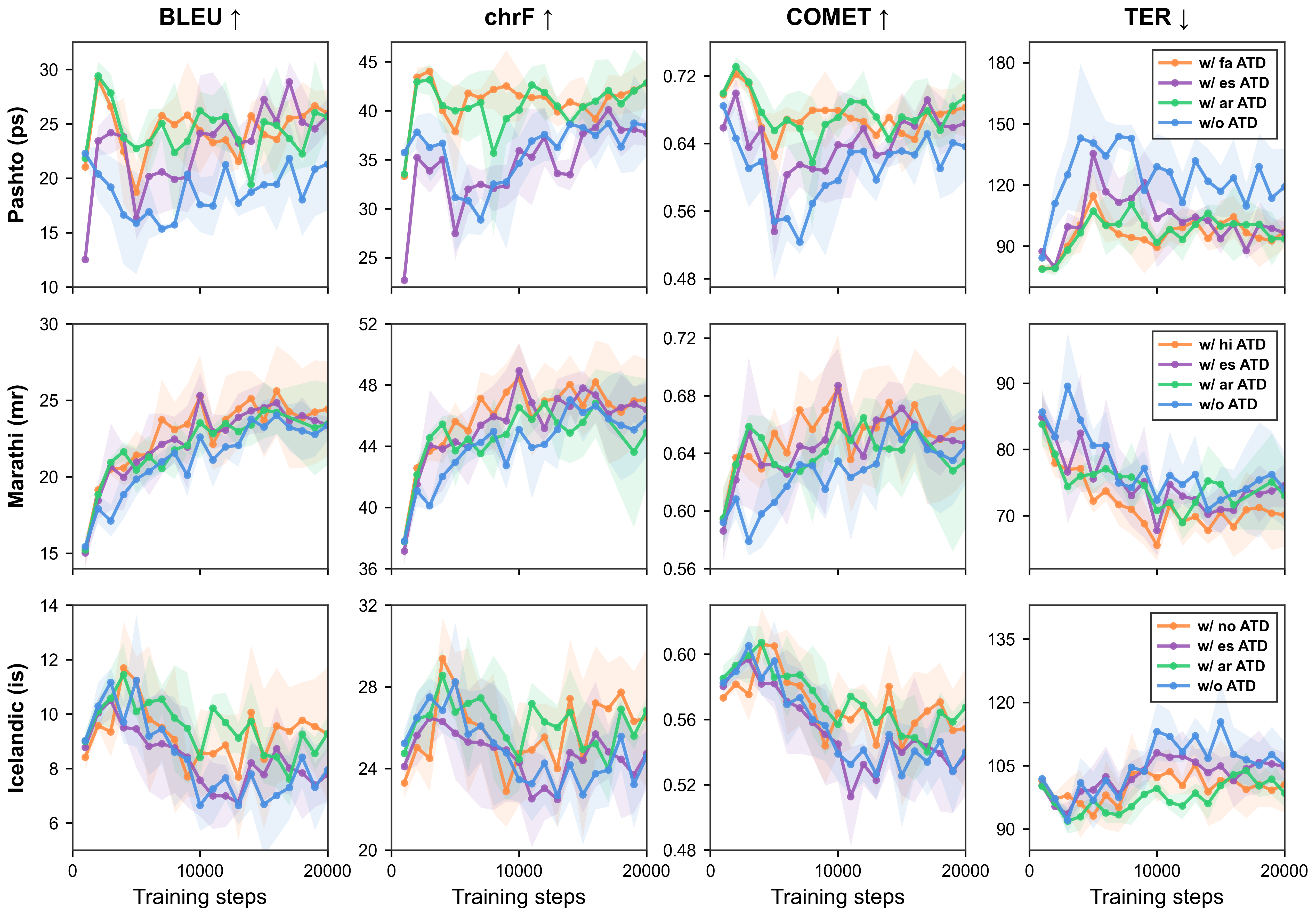}
    \caption{Comparison of ATD regularization using different reference languages (related, \texttt{es}, and \texttt{ar}). While all ATD variants generally outperform the baseline, the linguistically related reference provides the most consistent and principled improvement.}
    \label{fig:app_4lines}
\end{figure}

As shown in Fig.~\ref{fig:app_4lines}, adding ATD regularization generally leads to improvements over the baseline, regardless of the reference language. This can be attributed to the fact that transformer-based translation models share ``universal" attention patterns for core tasks, such as focusing on syntactic heads. However, the linguistically related reference (e.g., \texttt{fa} for \texttt{ps}) consistently serves as the most robust anchor, yielding the most stable performance across all metrics without significant degradation.

The subtle differences between reference languages are also influenced by the regularization weight $\lambda$. As an auxiliary loss, ATD must be balanced: if $\lambda$ is too small, the fine-grained cross-linguistic differences remain masked by the dominant cross-entropy loss ($\mathcal{L}_{\text{CE}}$); if $\lambda$ is too large, it may interfere with the translation objective itself. Within a reasonable weight range, the linguistically related language provides a more principled structural prior that aligns better with the target language's underlying typology, offering a safe and effective mechanism for low-resource transfer learning. However, within this constraint, the fine-grained variations in ATD loss across different references tend to be eclipsed, resulting in relatively similar performance regardless of the chosen reference language.

%% file: main_arxiv.bib
@article{arivazhagan2019massively,
  title={Massively multilingual neural machine translation in the wild: Findings and challenges},
  author={Arivazhagan, Naveen and Bapna, Ankur and Firat, Orhan and Lepikhin, Dmitry and Johnson, Melvin and Krikun, Maxim and Chen, Mia Xu and Cao, Yuan and Foster, George and Cherry, Colin and others},
  journal={arXiv preprint arXiv:1907.05019},
  year={2019}
}

@article{artetxe2019massively,
  title     = {Massively multilingual sentence embeddings for zero-shot cross-lingual transfer and beyond},
  author    = {Artetxe, Mikel and Schwenk, Holger},
  journal   = {Transactions of the association for computational linguistics},
  volume    = {7},
  pages     = {597--610},
  year      = {2019},
  publisher = {MIT Press One Rogers Street, Cambridge, MA 02142-1209, USA journals-info~…}
}

@inproceedings{bojar2014findings,
  title={Findings of the 2014 workshop on statistical machine translation},
  author={Bojar, Ond{\v{r}}ej and Buck, Christian and Federmann, Christian and Haddow, Barry and Koehn, Philipp and Leveling, Johannes and Monz, Christof and Pecina, Pavel and Post, Matt and Saint-Amand, Herve and others},
  booktitle={Proceedings of the ninth workshop on statistical machine translation},
  pages={12--58},
  year={2014}
}

@article{bouckaert2012mapping,
  title     = {Mapping the origins and expansion of the Indo-European language family},
  author    = {Bouckaert, Remco and Lemey, Philippe and Dunn, Michael and Greenhill, Simon J and Alekseyenko, Alexander V and Drummond, Alexei J and Gray, Russell D and Suchard, Marc A and Atkinson, Quentin D},
  journal   = {Science},
  volume    = {337},
  number    = {6097},
  pages     = {957--960},
  year      = {2012},
  publisher = {American Association for the Advancement of Science}
}

@book{campbell2013historical,
  title={Historical linguistics: An introduction},
  author={Campbell, Lyle},
  year={2013},
  publisher={Edinburgh University Press}
}

@article{conneau2017word,
  title   = {Word translation without parallel data},
  author  = {Conneau, Alexis and Lample, Guillaume and Ranzato, Marc'Aurelio and Denoyer, Ludovic and J{\'e}gou, Herv{\'e}},
  journal = {arXiv preprint arXiv:1710.04087},
  year    = {2017}
}

@article{cuturi2013sinkhorn,
  title   = {Sinkhorn distances: Lightspeed computation of optimal transport},
  author  = {Cuturi, Marco},
  journal = {Advances in neural information processing systems},
  volume  = {26},
  year    = {2013}
}

@misc{dryer2022wals,
  title     = {The World Atlas of Language Structures Online (v2020.3)},
  author    = {Dryer, Matthew S. and Haspelmath, Martin},
  year      = {2022},
  publisher = {Zenodo},
  doi       = {10.5281/zenodo.7385533},
  url       = {https://doi.org/10.5281/zenodo.7385533}
}

@article{dubey2024llama,
  title   = {The llama 3 herd of models},
  author  = {Dubey, Abhimanyu and Jauhri, Abhinav and Pandey, Abhinav and Kadian, Abhishek and Al-Dahle, Ahmad and Letman, Aiesha and Mathur, Akhil and Schelten, Alan and Yang, Amy and Fan, Angela and others},
  journal = {arXiv e-prints},
  pages   = {arXiv--2407},
  year    = {2024}
}

@article{fan2021beyond,
  title   = {Beyond english-centric multilingual machine translation},
  author  = {Fan, Angela and Bhosale, Shruti and Schwenk, Holger and Ma, Zhiyi and El-Kishky, Ahmed and Goyal, Siddharth and Baines, Mandeep and Celebi, Onur and Wenzek, Guillaume and Chaudhary, Vishrav and others},
  journal = {Journal of Machine Learning Research},
  volume  = {22},
  number  = {107},
  pages   = {1--48},
  year    = {2021}
}

@inproceedings{feng2022language,
  title={Language-agnostic BERT sentence embedding},
  author={Feng, Fangxiaoyu and Yang, Yinfei and Cer, Daniel and Arivazhagan, Naveen and Wang, Wei},
  booktitle={Proceedings of the 60th annual meeting of the association for computational linguistics (volume 1: Long papers)},
  pages={878--891},
  year={2022}
}

@inproceedings{feydy2019interpolating,
  title     = {Interpolating between Optimal Transport and MMD with Sinkhorn Divergences},
  author    = {Feydy, Jean and others},
  booktitle = {Proceedings of the 22nd International Conference on Artificial Intelligence and Statistics},
  year      = {2019}
}

@article{gray2003language,
  title     = {Language-tree divergence times support the Anatolian theory of Indo-European origin},
  author    = {Gray, Russell D and Atkinson, Quentin D},
  journal   = {Nature},
  volume    = {426},
  number    = {6965},
  pages     = {435--439},
  year      = {2003},
  publisher = {Nature Publishing Group UK London}
}

@inproceedings{gu2018universal,
  title={Universal neural machine translation for extremely low resource languages},
  author={Gu, Jiatao and Awadalla, Hany Hassan and Devlin, Jacob and Li, Victor OK},
  booktitle={Proceedings of the 2018 Conference of the North American Chapter of the Association for Computational Linguistics: Human Language Technologies, Volume 1 (Long Papers)},
  pages={344--354},
  year={2018}
}

@article{swj-glottocodes,
  author  = {Harald Hammarström and Robert Forkel},
  journal = {Semantic Web Journal},
  number  = {6},
  pages   = {917-924},
  title   = {Glottocodes: Identifiers Linking Families, Languages and Dialects to Comprehensive Reference Information},
  url     = {https://content.iospress.com/articles/semantic-web/sw212843},
  volume  = {13},
  year    = {2022}
}

@article{holman2008explorations,
  title={Explorations in automated language classification},
  author={Holman, Eric W and Wichmann, S{\o}ren and Brown, Cecil H and Velupillai, Viveka and M{\"u}ller, Andr{\'e} and Bakker, Dik and others},
  journal={Folia Linguistica},
  volume={42},
  number={2},
  pages={331--354},
  year={2008}
}

@article{johnson2017google,
  title={Google’s multilingual neural machine translation system: Enabling zero-shot translation},
  author={Johnson, Melvin and Schuster, Mike and Le, Quoc and Krikun, Maxim and Wu, Yonghui and Chen, Zhifeng and Thorat, Nikhil and Vi{\'e}gas, Fernanda and Wattenberg, Martin and Corrado, Greg and others},
  journal={Transactions of the Association for Computational Linguistics},
  volume={5},
  pages={339--351},
  year={2017}
}

@inproceedings{kudo2018sentencepiece,
  title     = {SentencePiece: A Simple and Language Independent Subword Tokenizer and Detokenizer for Neural Text Processing},
  author    = {Taku Kudo and John Richardson},
  booktitle = {Proceedings of the 2018 Conference on Empirical Methods in Natural Language Processing (EMNLP): System Demonstrations},
  year      = {2018},
  pages     = {66--71}
}

@inproceedings{lauscher2020zero,
  title={From zero to hero: On the limitations of zero-shot language transfer with multilingual Transformers},
  author={Lauscher, Anne and Ravishankar, Vinit and Vuli{\'c}, Ivan and Glava{\v{s}}, Goran},
  booktitle={Proceedings of the 2020 Conference on Empirical Methods in Natural Language Processing (EMNLP)},
  pages={4483--4499},
  year={2020}
}

@inproceedings{liu2016neural,
  title={Neural machine translation with supervised attention},
  author={Liu, Lemao and Utiyama, Masao and Finch, Andrew and Sumita, Eiichiro},
  booktitle={Proceedings of COLING 2016, the 26th International Conference on Computational Linguistics: Technical Papers},
  pages={3093--3102},
  year={2016}
}

@article{openai2023gpt4,
  title={GPT-4 Technical Report},
  author={OpenAI},
  journal={arXiv preprint arXiv:2303.08774},
  year={2023}
}

@inproceedings{papineni2002bleu,
  title     = {{BLEU}: a method for automatic evaluation of machine translation},
  author    = {Papineni, Kishore and Roukos, Salim and Ward, Todd and Zhu, Wei-Jing},
  booktitle = {Proceedings of the 40th annual meeting on Association for Computational Linguistics},
  pages     = {311--318},
  year      = {2002}
}

@book{peyre2019computational,
  title={Computational optimal transport: With applications to data science},
  author={Peyr{\'e}, Gabriel and Cuturi, Marco},
  year={2019},
  publisher={Now Foundations and Trends}
}

@inproceedings{pires2019multilingual,
  title={How multilingual is multilingual BERT?},
  author={Pires, Telmo and Schlinger, Eva and Garrette, Dan},
  booktitle={Proceedings of the 57th annual meeting of the association for computational linguistics},
  pages={4996--5001},
  year={2019}
}

@inproceedings{popovic2015chrf,
  title     = {chr{F}: character n-gram {F}-score for automatic {MT} evaluation},
  author    = {Popović, Maja},
  booktitle = {Proceedings of the Tenth Workshop on Statistical Machine Translation},
  pages     = {392--395},
  year      = {2015}
}

@inproceedings{rei2020comet,
  title     = {{COMET}: A neural framework for {MT} evaluation},
  author    = {Rei, Ricardo and Farinha, Ana C and Lavie, Alon and Specia, Lucia},
  booktitle = {Proceedings of the 2020 Conference on Empirical Methods in Natural Language Processing},
  year      = {2020}
}

@article{saitou1987neighbor,
  title={The neighbor-joining method: a new method for reconstructing phylogenetic trees},
  author={Saitou, Naruya and Nei, Masatoshi},
  journal={Molecular biology and evolution},
  volume={4},
  number={4},
  pages={406--425},
  year={1987}
}

@inproceedings{sennrich2016neural,
  title     = {Neural Machine Translation of Rare Words with Subword Units},
  author    = {Rico Sennrich and Barry Haddow and Alexandra Birch},
  booktitle = {Proceedings of the 54th Annual Meeting of the Association for Computational Linguistics (ACL)},
  year      = {2016},
  pages     = {1715--1725}
}

@article{serva2008indo,
  title     = {Indo-European languages tree by Levenshtein distance},
  author    = {Serva, Maurizio and Petroni, Filippo},
  journal   = {Europhysics letters},
  volume    = {81},
  number    = {6},
  pages     = {68005},
  year      = {2008},
  publisher = {IOP publishing}
}

@inproceedings{snover2006study,
  title={A study of translation edit rate with targeted human annotation},
  author={Snover, Matthew and Dorr, Bonnie and Schwartz, Richard and Micciulla, Linnea and Makhoul, John},
  booktitle={Proceedings of the 7th Conference of the Association for Machine Translation in the Americas: Technical Papers},
  pages={223--231},
  year={2006}
}

@article{vaswani2017attention,
  title   = {Attention is all you need},
  author  = {Vaswani, Ashish and Shazeer, Noam and Parmar, Niki and Uszkoreit, Jakob and Jones, Llion and Gomez, Aidan N and Kaiser, {\L}ukasz and Polosukhin, Illia},
  journal = {Advances in neural information processing systems},
  volume  = {30},
  year    = {2017}
}

@inproceedings{wu2020all,
  title={Are all languages created equal in multilingual BERT?},
  author={Wu, Shijie and Dredze, Mark},
  booktitle={Proceedings of the 5th Workshop on Representation Learning for NLP},
  pages={120--130},
  year={2020}
}

@inproceedings{zhang-etal-2020-improving,
    title = {Improving Massively Multilingual Neural Machine Translation and Zero-Shot Translation},
    author = {Zhang, Biao and Williams, Philip and Titov, Ivan and Sennrich, Rico},
    booktitle = {Proceedings of the 58th Annual Meeting of the Association for Computational Linguistics},
    series = {},  
    month = {July},
    year = {2020},
    address = {Online},
    publisher = {Association for Computational Linguistics},
    url = {https://aclanthology.org/2020.acl-main.148},
    pages = {1628--1639}
}

@inproceedings{zoph2016transfer,
  title={Transfer learning for low-resource neural machine translation},
  author={Zoph, Barret and Yuret, Deniz and May, Jonathan and Knight, Kevin},
  booktitle={Proceedings of the 2016 conference on empirical methods in natural language processing},
  pages={1568--1575},
  year={2016}
}
